\documentclass[10pt,letterpaper]{article}
\usepackage[utf8]{inputenc}
\usepackage[T1]{fontenc}
\usepackage[margin=0.85in]{geometry}
\usepackage{lmodern}
\usepackage{microtype}
\usepackage{authblk}
\usepackage{amsmath,amssymb}
\usepackage{graphicx}
\usepackage[table]{xcolor}
\usepackage{booktabs,multirow,array,tabularx,longtable}
\usepackage{caption}
\usepackage{subfigure}
\usepackage{adjustbox}
\usepackage{placeins}
\usepackage{enumitem}
\usepackage[numbers,sort&compress]{natbib}
\usepackage{xurl}
\usepackage[hidelinks]{hyperref}

\usepackage{url}
\usepackage{amsfonts}
\usepackage{nicefrac}
\usepackage{colortbl}
\usepackage{wrapfig}
\usepackage{lipsum}

\usepackage{newunicodechar}
\newunicodechar{：}{:}
\usepackage{CJKutf8}
\usepackage{algorithm}
\usepackage{algpseudocode}
\title{LongHisDoc: A Comprehensive Benchmark for Chinese Long Historical Document Understanding}
\author[1]{Zeyu Shan}
\author[1]{Yongxin Shi}
\author[1]{Shihang Wu}
\author[1]{Peirong Zhang}
\author[2]{Yuanyuan Fang}
\author[2]{Zhaopeng Wang}
\author[1,*]{Lianwen Jin}
\affil[1]{School of Electronic and Information Engineering, South China University of Technology, Guangzhou, 510641, China}
\affil[2]{
College of Literature and Journalism, Sichuan University, Chengdu, 610065, China
}
\affil[*]{Corresponding author: Lianwen Jin (eelwjin@scut.edu.cn)}
\date{}
\hypersetup{pdftitle={LongHisDoc: A Comprehensive Benchmark for Chinese Long Historical Document Understanding},pdfauthor={Zeyu Shan, Yongxin Shi, Shihang Wu, Peirong Zhang, Yuanyuan Fang, Zhaopeng Wang, Lianwen Jin}}
\begin{document}
\maketitle
\begin{abstract}
Chinese historical documents are invaluable cultural treasures, but their complexity poses significant comprehension challenges for non-specialists. Despite the success of Large Language Models (LLMs) and Large Vision-Language Models (LVLMs) in modern document tasks, their capabilities in \textbf{Long} (full-book) \textbf{His}torical \textbf{Doc}ument understanding remain largely unexplored. To bridge this gap, we introduce \textbf{LongHisDoc}, the \textbf{first} benchmark tailored for long historical document understanding. 
LongHisDoc comprises \textbf{1,012 expert-annotated} QA pairs derived from 101 PDF documents, with an average of \textbf{85.1 pages} and \textbf{42,776.4 tokens} per document. It spans 10 categories and features four QA types: single-page, cross-page, multi-hop, and unanswerable. Answering these questions requires integrating multimodal evidence from text, layout, and visual elements. Our evaluation reveals that the top-performing LVLM and LLM achieve only 62.9\% and 69.9\% accuracy, respectively, highlighting a substantial performance gap. Furthermore, we assess various modal retrievers and the end-to-end RAG performance to provide references for Retrieval-Augmented Generation (RAG) systems. We believe that LongHisDoc serves as a comprehensive benchmark to catalyze future research in full-book historical document understanding. 
The benchmark and related code are available at \url{https://huggingface.co/datasets/qweq12433454/LongHisDoc}.
\end{abstract}

\section{Introduction}

Over the course of five thousand years, China has preserved a vast collection of invaluable historical documents containing rich knowledge of history, culture, politics, and economics. The preservation and analysis of these materials are essential for cultural heritage preservation and digital humanities research. However, their archaic language, intricate layouts, and degraded visual quality pose substantial challenges to automatic document understanding. Recent advances in Large Language Models (LLMs) and Large Vision-Language Models (LVLMs) have created new opportunities for understanding these historical materials.

Despite these advances, existing historical document benchmarks primarily focus on low-level perception tasks, such as character detection and recognition, or high-level understanding of isolated pages. Meanwhile, general document understanding has progressively evolved toward multi-page and long-context reasoning, as demonstrated by MMLongBench-Doc~\cite{ma2024mmlongbench} and LongDocURL~\cite{deng2025longdocurl}. However, these benchmarks predominantly target modern documents and do not adequately capture the distinctive characteristics of Chinese historical documents. Consequently, the capabilities of current LLMs and LVLMs in understanding entire historical volumes remain largely unexplored.

\begin{figure*}[t]
  \centering
  \includegraphics[width=0.95\linewidth]{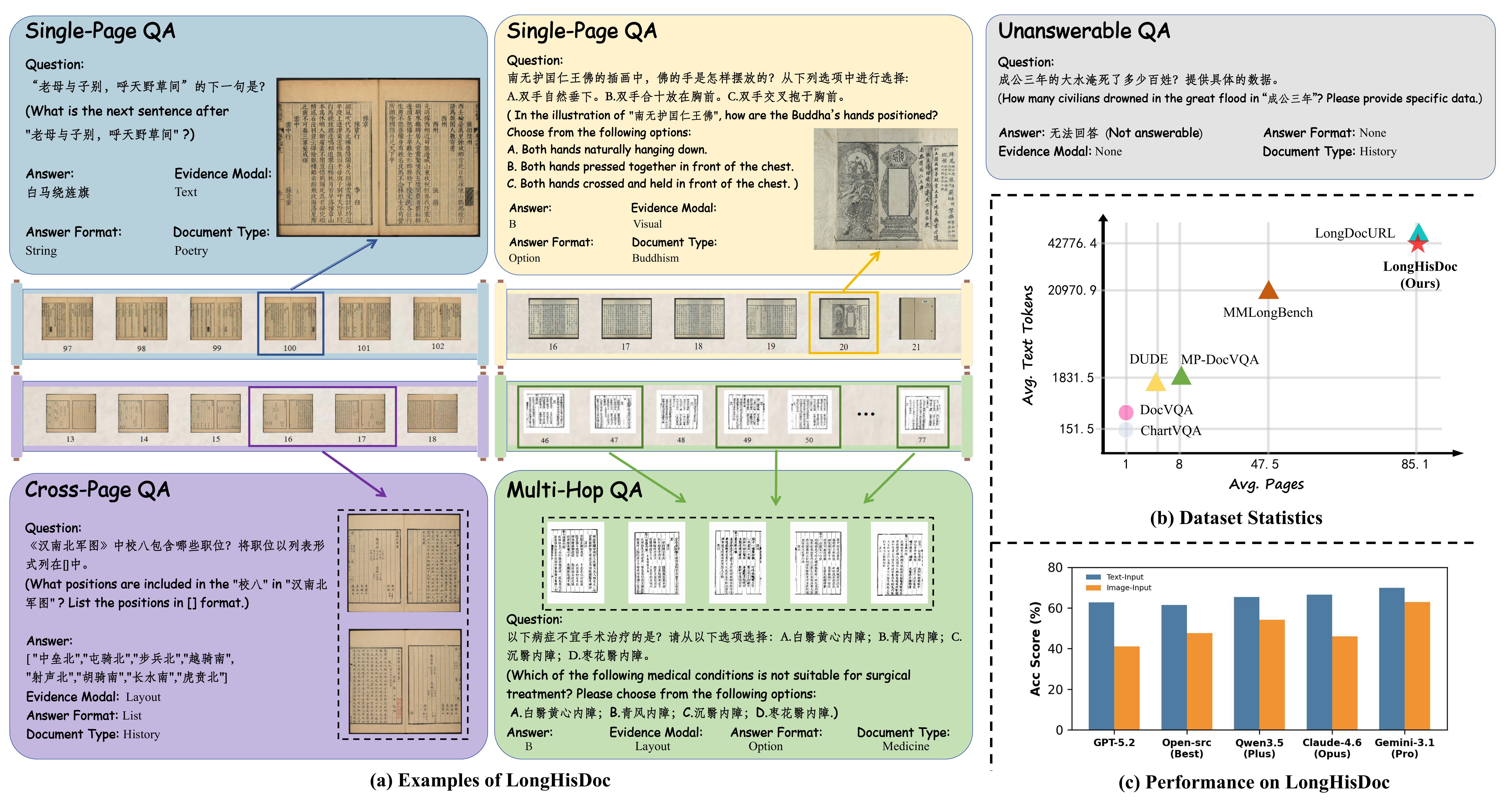}
  \caption{LongHisDoc assesses the capabilities of LLMs and LVLMs in understanding long-context historical documents. The experiments (bottom-right) reveal that most LVLMs face difficulties, showing significantly lower performance compared to LLMs fed with Optical Character Recognition (OCR)-parsed historical documents. Zoom in for better view.}
  \label{fig:LongHisDoc}
  \vspace{-3mm}
\end{figure*}

Unlike modern long documents, Chinese historical volumes introduce additional challenges arising from the interaction between long-context reasoning and historical document characteristics. Specifically, we identify four key challenges:
\textbf{(1) Global Structural Continuity and Local Layout Noise:} Historical volumes exhibit complex hierarchical structures spanning numerous pages, while interlinear annotations and unconventional reading orders complicate structural interpretation.
\textbf{(2) Long-range Semantic Cohesion in Classical Chinese:} The concise and context-dependent nature of Classical Chinese requires models to maintain semantic consistency and track entity relationships across distant pages.
\textbf{(3) Cumulative Degradation and Error Propagation:} Historical documents frequently suffer from ink fading, physical damage, and irregular printing, introducing visual degradation and OCR-induced noise that can affect downstream understanding.
\textbf{(4) Multi-hop Evidence Retrieval and Integration:} Relevant information is often scattered across non-consecutive pages, requiring models to locate and integrate distributed evidence for accurate reasoning.
These challenges highlight the necessity of a dedicated benchmark for systematically evaluating long historical document understanding.

To address this gap, we introduce \textbf{LongHisDoc}, the \textbf{first} benchmark specifically designed to evaluate the long-context historical document understanding capabilities of LLMs and LVLMs. LongHisDoc comprises 101 PDF-formatted historical documents spanning 10 categories, with an average length of 85.1 pages per document. To ensure annotation quality, 15 human experts participated in constructing and validating 1,012 question-answer pairs. These questions cover four types (\textit{single-page}, \textit{cross-page}, \textit{multi-hop}, and \textit{unanswerable}), with supporting evidence drawn from three sources (\textit{text}, \textit{layout}, and \textit{visual}). Figure~\ref{fig:LongHisDoc}(a) illustrates representative examples, while Figure~\ref{fig:LongHisDoc}(b) compares document lengths across different benchmarks.

We conduct comprehensive evaluations under two input settings: \textbf{text-input} and \textbf{image-input}. As shown in Figure~\ref{fig:LongHisDoc}(c), even the best-performing LVLM, Gemini-3.1-Pro~\cite{Gemini_3_1_pro}, achieves only 62.9\% accuracy in the image-input setting, while the same model achieves 69.9\% in the text-input setting using OCR results. Although OCR-based textual representations generally improve performance, they remain susceptible to OCR-induced noise. Furthermore, our evaluation of multimodal retrievers reveals that state-of-the-art visual-based methods achieve a Recall@5 of 80.1\%, outperforming text-based counterparts (78.2\%). These findings highlight the limitations of current models and the potential of visual-based retrieval for reducing reliance on OCR in historical document understanding.

Our main contributions are summarized as follows:
\begin{itemize}
    \item We introduce \textbf{LongHisDoc}, the first benchmark dedicated to long historical document understanding, comprising 101 historical documents across 10 categories and 1,012 human-annotated question-answer pairs covering four question types and three evidence sources.

    \item We systematically evaluate representative LLMs and LVLMs under different input modalities, together with text-based and visual-based retrievers and end-to-end Retrieval-Augmented Generation (RAG) methods, establishing comprehensive baselines for long historical document understanding.

    \item We provide fine-grained analyses across historical document categories, question types, evidence sources, and input lengths, revealing key challenges and offering insights for improving future LLMs, LVLMs, and RAG systems in historical document understanding.
\end{itemize}

\section{Related Work}

\subsection{Historical Document Analysis and Understanding}

Automatic historical document analysis has attracted increasing attention due to its importance in cultural heritage preservation and digital humanities. Early benchmarks, including MTHv2~\cite{ma2020joint}, M\textsuperscript{5}HisDoc~\cite{NEURIPS2023_f7b424d2}, and HisDoc-1B~\cite{shi2025large}, primarily focused on document parsing and low-level perception tasks, such as character detection and recognition. These benchmarks have facilitated the development of historical document digitization techniques but do not directly assess higher-level semantic understanding.

With the emergence of LLMs and LVLMs, recent benchmarks have extended historical document analysis toward semantic understanding and question answering. For instance, MCS-Bench~\cite{liu-etal-2025-mcs} and AncientDoc~\cite{ancientdoc} evaluate models' capabilities in interpreting historical document content. However, these benchmarks predominantly focus on isolated historical snippets or individual pages, leaving document-level comprehension and long-range reasoning largely unexplored.

In contrast, LongHisDoc extends historical document understanding from isolated pages to entire historical volumes. By incorporating cross-page, multi-hop, and unanswerable questions, our benchmark evaluates not only local content comprehension but also the ability to integrate distributed evidence and assess answerability across long historical documents.

\subsection{Multi-page and Long Document Understanding}

Document understanding has progressively evolved from single-page visual question answering toward multi-page and long-context reasoning. Early benchmarks such as DocVQA, ChartVQA, and InfoVQA primarily focused on understanding localized document content. Subsequently, MP-DocVQA, DUDE, and SlideVQA extended document question answering to multi-page scenarios, introducing additional challenges in page localization and cross-page information integration.

More recently, MMLongBench-Doc~\cite{ma2024mmlongbench} and LongDocURL~\cite{deng2025longdocurl} have advanced the evaluation of long-document understanding by incorporating substantially longer documents and questions requiring cross-page reasoning, multi-hop evidence integration, and robustness to unanswerable queries. These benchmarks provide valuable insights into the long-context capabilities of current LLMs and LVLMs. Nevertheless, their evaluation materials primarily consist of modern documents, whose language, visual quality, and document structures differ considerably from those of Chinese historical texts.

As summarized in Table~\ref{tab:table1}, existing historical document benchmarks lack comprehensive long-document reasoning tasks, whereas modern long-document benchmarks do not account for the distinctive characteristics of historical documents. LongHisDoc bridges this gap by combining full-volume historical documents with diverse question types and multimodal evidence sources, providing a dedicated evaluation framework for Chinese long historical document understanding.

\label{related work}

\begin{table*}[]
\caption{Comparison between LongHisDoc and previous document understanding datasets. \textbf{T/L/V}:text/layout/visual. \textbf{HisDoc Rel.}: historical document relevance.}
\label{tab:table1}
\resizebox{\textwidth}{!}{ %
\begin{tabular}{c|c|cc|ccc|clll|c|c}
\hline
  \multirow{2}{*}{\textbf{Benchmarks}}& 
  \multirow{2}{*}{\textbf{Venue \& Year}} & 
  \multicolumn{2}{c|}{\textbf{Document}} &
  \multicolumn{3}{c|}{\textbf{Question type}} &
  \multicolumn{4}{c|}{\textbf{Answer Evidence}} &
  \textbf{Doc Type} &
  \multicolumn{1}{c}{\textbf{HisDoc}} \\
                & & Avg. Pages & Avg. Tokens & Cross-Page & Multi-Hop  & Unans     & \multicolumn{4}{c|}{Source} & Num & \multicolumn{1}{c}{Rel.} \\ \hline
DocVQA \cite{mathew2021docvqa}     & WACV 2021    & 1.0     & 151.5    & \textcolor{gray!85!black}{\(\times\)}          & \textcolor{gray!85!black}{\(\times\)}          & \textcolor{gray!85!black}{\(\times\)}         & \multicolumn{4}{c|}{T/L/V}  & -   & \textcolor{gray!85!black}{\(\times\)}                    \\
ChartVQA \cite{masry2022chartqa}   & ACL 2022    & 1.0     & 236.9    & \textcolor{gray!85!black}{\(\times\)}          & \textcolor{gray!85!black}{\(\times\)}          & \textcolor{gray!85!black}{\(\times\)}         & \multicolumn{4}{c|}{V}      & -   & \textcolor{gray!85!black}{\(\times\)}                    \\
MCS-Bench \cite{liu-etal-2025-mcs}    & ACL 2025    & 1.0     & 314.2    & \textcolor{gray!85!black}{\(\times\)}          & \textcolor{gray!85!black}{\(\times\)}          & \textcolor{gray!85!black}{\(\times\)}         & \multicolumn{4}{c|}{T/L/V}    & -   & \textcolor{green}{\(\checkmark\)}  \\
AncientDoc \cite{ancientdoc}    & ArXiv 2025    & 1.0     & 283.2    & \textcolor{gray!85!black}{\(\times\)}          & \textcolor{gray!85!black}{\(\times\)}          & \textcolor{gray!85!black}{\(\times\)}         & \multicolumn{4}{c|}{T/L/V}    & 14   & \textcolor{green}{\(\checkmark\)}  \\
InfoVQA \cite{mathew2022infographicvqa}    & WACV 2022    & 1.2     & 288.0    & \textcolor{gray!85!black}{\(\times\)}          & \textcolor{gray!85!black}{\(\times\)}          & \textcolor{gray!85!black}{\(\times\)}         & \multicolumn{4}{c|}{L/V}    & -   & \textcolor{gray!85!black}{\(\times\)}                    \\
MP-DocVQA \cite{tito2023hierarchical}   & PR 2023   & 8.3     & 2026.6   & \textcolor{gray!85!black}{\(\times\)}          & \textcolor{gray!85!black}{\(\times\)}          & \textcolor{gray!85!black}{\(\times\)}         & \multicolumn{4}{c|}{T/L/V}  & -   & \textcolor{gray!85!black}{\(\times\)}                    \\
DUDE \cite{van2023dude}     & ICCV 2023      & 5.7     & 1831.5   & \textcolor{green}{\(\checkmark\)} (1.1\%)         & \textcolor{green}{\(\checkmark\)}(1.0\%)         & \textcolor{green}{\(\checkmark\)}(12.7\%)         & \multicolumn{4}{c|}{T/L/V}  & -   & \textcolor{gray!85!black}{\(\times\)}                    \\
SlideVQA \cite{tanaka2023slidevqa}    & AAAI 2023   & 20.0    & 2030.5   & \textcolor{green}{\(\checkmark\)}(4.7\%)          & \textcolor{green}{\(\checkmark\)}(9.2\%)          & \textcolor{gray!85!black}{\(\times\)}         & \multicolumn{4}{c|}{T/L/V}  & -   & \textcolor{gray!85!black}{\(\times\)}                    \\
MMLongBench-Doc \cite{ma2024mmlongbench} & NeurIPS 2024  & 47.5    & 20970.9  & \textcolor{green}{\(\checkmark\)}(11.6\%)          & \textcolor{green}{\(\checkmark\)}(21.3\%)          & \textcolor{green}{\(\checkmark\)}(22.5\%)         & \multicolumn{4}{c|}{T/L/V}  & 7   & \textcolor{gray!85!black}{\(\times\)}                    \\
LongDocURL \cite{deng2025longdocurl}   & ACL 2025  & 85.6    & 43622.6  & \textcolor{green}{\(\checkmark\)}(22.1\%)          & \textcolor{green}{\(\checkmark\)}(30.8\%)           & \textcolor{green}{\(\checkmark\)}(0.1\%)          & \multicolumn{4}{c|}{T/L/V}  & 8   & \textcolor{gray!85!black}{\(\times\)}                    \\ \hline 
LongHisDoc    &-  & 85.1    & 42776.4  & \textcolor{green}{\(\checkmark\)}(19.4\%) & \textcolor{green}{\(\checkmark\)}(17.0\%) & \textcolor{green}{\(\checkmark\)}(9.7\%) & \multicolumn{4}{c|}{T/L/V}  & 10  & \textcolor{green}{\(\checkmark\)}                    \\ \hline
\end{tabular}
}
\vspace{-3mm}
\end{table*}

\section{LongHisDoc}

\begin{figure*}[h]
  \centering
  \includegraphics[width=0.95\linewidth]{pic2/data_construction.pdf}
  \caption{The data construction pipeline of LongHisDoc consists of four stages: (1) \textbf{Documents Collection}; (2) \textbf{Q\&A Annotation}; (3) \textbf{Quality Control}; (4) \textbf{Data Standardization}.}
  \label{fig:Data_Construction}
\end{figure*}
\label{section3}

With the aim of constructing a comprehensive and reliable benchmark for evaluating capabilities in long historical document understanding, we develop a four-stage pipeline as shown in Figure \ref{fig:Data_Construction}, which includes \textit{documents collection}, \textit{QA annotation}, \textit{quality control}, and \textit{data standardization}. In the following sections, we will elucidate each of these stages in detail.

\renewcommand{\arraystretch}{1.05}

\begin{figure*}[t]  %
  \centering
  \begin{minipage}[t]{.55\textwidth}
    \vspace{0pt}\centering
    \includegraphics[width=.75\linewidth]{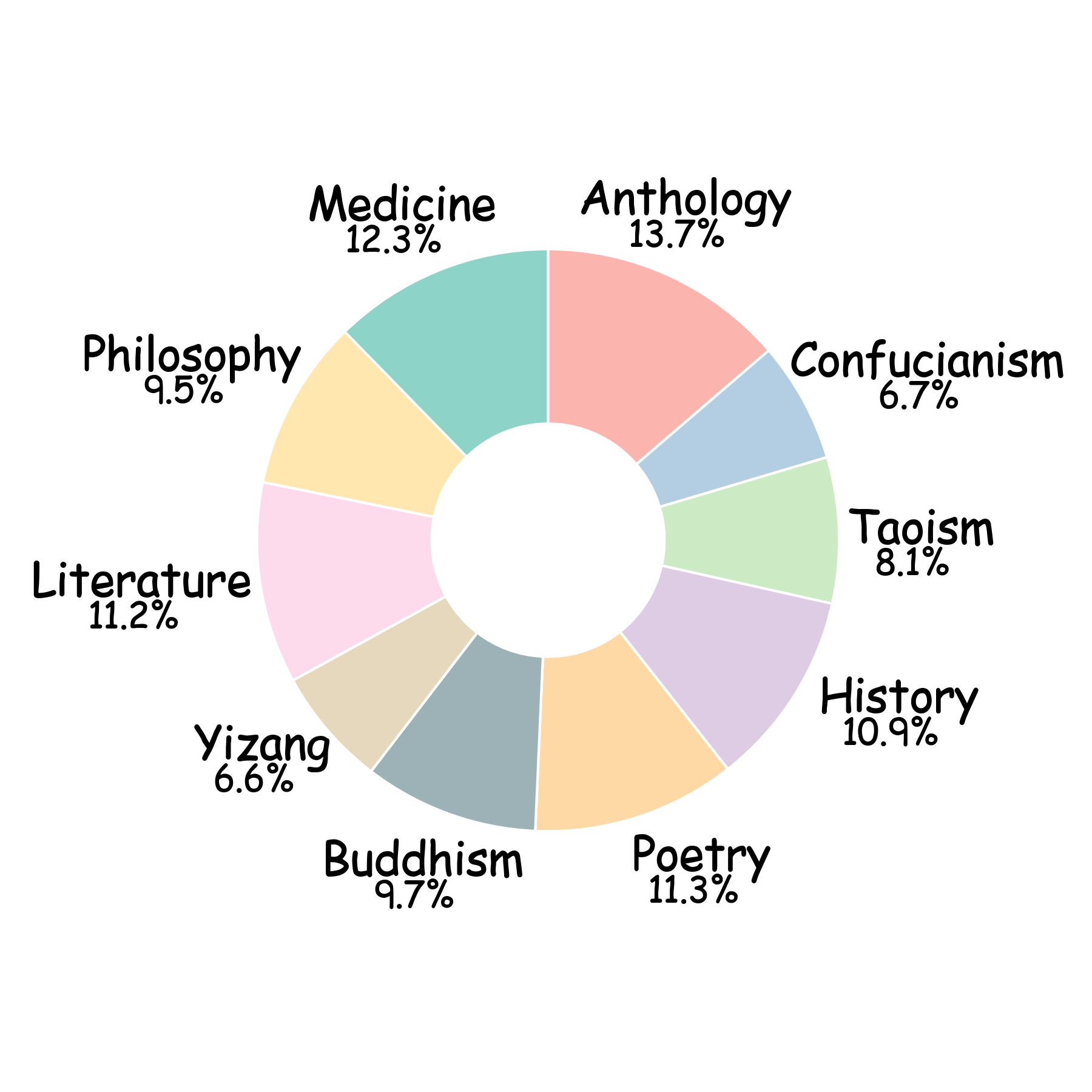}\\[4pt]
    \includegraphics[width=.85\linewidth]{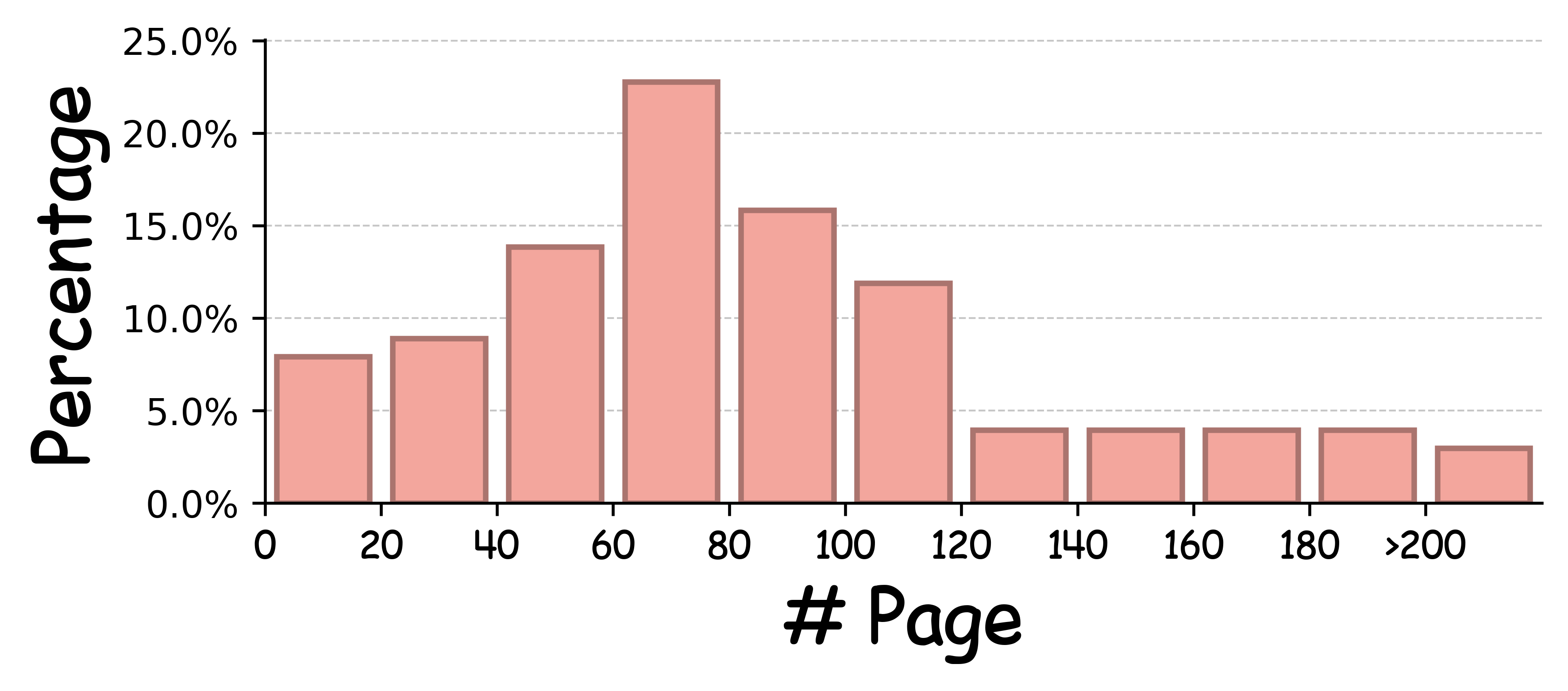}\\[4pt]
    \captionof{figure}{Detailed distribution of documents. \textbf{Top:} Document type. \textbf{Bottom:} Page number.}
    \label{fig:fig3}
  \end{minipage}
  \hfill
  \begin{minipage}[t]{.40\textwidth}
    \vspace{0pt}\centering
    \captionof{table}{Dataset Statistics}
    \label{tab:table2}
    \begin{tabular}{lc}
      \toprule
      Statistic & Number \\
      \midrule
      \textit{\textbf{Documents}} & \\
      \quad Total & 101 \\
      \quad Type & 10 \\
      \quad Average/Median pages & 85.1/76 \\
      \quad Average/Median tokens & 42776.4/33652 \\
      \midrule
      \textit{\textbf{Q\&A pairs}} & \\
      \quad Total & 1012 \\
      \quad Single-page QA & 546 (53.9\%) \\
      \quad Cross-page QA & 196 (19.4\%) \\
      \quad Multi-hop QA & 172 (17.0\%) \\
      \quad Unanswerable QA & 98 (9.7\%) \\
      \midrule
      \textit{\textbf{Evidence Source}} & \\
      \quad Text & 454 (44.8\%) \\
      \quad Layout & 363 (35.9\%) \\
      \quad Visual & 97 (9.6\%) \\
      \midrule
      \textit{\textbf{Answer Format}} & \\
      \quad String & 569 (56.2\%) \\
      \quad Integer & 120 (11.9\%) \\
      \quad List & 132 (13.0\%) \\
      \quad Bool & 57 (5.6\%) \\
      \quad Option & 36 (3.6\%) \\
      \quad None & 98 (9.7\%) \\
      \bottomrule
    \end{tabular}
  \end{minipage}
  \vspace{-3mm}
\end{figure*}

\subsection{Documents Collection}

To comprehensively assess the historical document understanding capabilities of Large Language Models (LLMs) and Large Visual-Language Models (LVLMs), our dataset needs to cover a broad spectrum of historical document categories. Therefore, we select HisDoc-1B \cite{shi2025large}, the largest and most comprehensive historical document dataset currently, as our data source. Although HisDoc-1B contains 40,281 historical books with a total of 3,163,330 images, it does not categorize these books. 
Following the classification scheme of the Daizhige corpus \cite{daizhige}, we categorize the books into the following 10 types: \textbf{History}, \textbf{Poetry}, \textbf{Buddhism}, \textbf{Confucianism}, \textbf{Taoism}, \textbf{Anthology}, \textbf{Yizang}, \textbf{Literature}, \textbf{Philosophy}, and \textbf{Medicine}. 
However, manually classifying the 40,281 documents in HisDoc-1B is impractical. To address this issue, we propose a \textbf{semi-automated pipeline based on keyword-matching}. First, we sample 50 corresponding book titles for each category from the existing dataset \cite{daizhige} and websites (\url{https://www.guoxuedashi.net/}, \url{https://www.shidianguji.com/}).
Next, we fed the definitions of the ten categories and the book titles into GPT-4o, Gemini-2.5-Pro, and Qwen-Max to extract keywords via in-context learning. After merging and deduplicating the results, we obtained 10–15 keywords per category. Using these keywords, we successfully match and categorize 8,181 historical documents from HisDoc-1B. To verify the reliability of the automated keyword-matching results, we conducted a manual sampling inspection. The manual review shows a classification accuracy of 96.3\%, indicating that the matching strategy based on authoritative metadata and category keywords is highly reliable for historical-document categorization. 
Subsequently, to ensure diversity, we selected no more than three representative historical documents for each keyword. 
Furthermore, to ensure the suitability for long-context evaluation, we strictly filtered out fragmented or partial document snippets, preserving only intact, multi-page volumes with high structural continuity. 
We ultimately selected 200 candidate documents.

\subsection{Q\&A Annotation}
Establishing a reliable benchmark for long historical documents requires considerations beyond simple question-answering.
Specifically, the benchmark must provide explicit reasoning paths to verify the interpretability of model outputs. Meanwhile, it should incorporate "unanswerable" scenarios to assess model reliability and prevent hallucinations. Crucially, the tasks must transcend low-level perceptive functions, such as text detection or recognition; instead, annotators are required to propose questions with practical significance that demand deep semantic understanding and historical interpretation. 

In order to operationalize these requirements, our QA pairs annotation follows these rules: \textbf{(1)} For each historical document, annotators are asked to provide at least 5 single-page, 2 cross-page, and 2 multi-hop QA pairs, along with 1 unanswerable question.
{Specifically, \textbf{Single-page QA} refers to question-answer pairs whose answers can be directly derived from the content of a single page in a historical document without requiring complex reasoning. \textbf{Cross-page QA} refers to question-answer pairs that require combining continuous content across pages, i.e., two consecutive page images where the relevant information is interrupted by a page boundary. \textbf{Multi-hop QA} refers to question-answer pairs that can only be answered by synthesizing information distributed across multiple pages in different locations of the historical document, where the information appears in \(\ge 2\) non-consecutive or separate page images. Meanwhile, \textbf{Unanswerable questions} refer to questions that cannot be answered based on the entire historical document. During annotation, annotators were required to ensure that the relevant evidence for their questions did not appear anywhere in the historical document, while also ensuring that the questions were not completely unrelated to the topic of the document.
\textbf{(2)} Besides QA pairs, annotators must indicate the evidence page number (aligned with the PDF page number). \textbf{(3)} Questions should have professional and practical significance. \textbf{(4)} To ensure data diversity, evidence page locations should not be overly concentrated, and question patterns should not be too singular.}  
 
In addition, we employ 15 expert-level annotators. All of them are at least graduate students in the field of historical document studies. Before the formal annotation, we first train them to annotate two historical documents respectively and provide personalized feedback on the issues in their annotations until they meet the aforementioned standards.
During the formal annotation, annotators are allowed to prioritize documents related to their research interests. 
{Meanwhile, based on feedback from expert annotators, we further screened the candidate documents during the QA annotation process. We excluded documents whose content fell outside the scope of this study, was overly narrow in thematic focus, or lacked sufficient scholarly representativeness. Annotating a single historical document required approximately 60--90 minutes.} 

\subsection{Quality Control}

For the purpose of ensuring consistency in quality, we conduct quality control from three dimensions: \textbf{Relevance}, \textbf{Correctness}, and \textbf{Novelty}.

\textbf{Relevance} refers to whether the information on the evidence page can sufficiently support the correct answer to the question. And for unanswerable questions, it is essential to ensure that sufficient evidence does not exist in the historical document. 
In this process, we introduce LVLMs to reduce the workload. Concretely, for answerable questions, we input the question, the relevant evidence page images, and their OCR results into GPT-4o. If the model response does not match the standard answer, annotators are required to re-check the relevance of these samples. 
{
Therefore, for annotated unanswerable samples, we input both the full-text information and the annotated question into GPT-4o, allowing the model to make an initial judgment. Based on the evidence provided by the model, we preliminarily identify samples that contain supporting evidence but were incorrectly annotated as unanswerable. For samples that are judged to be correctly annotated, we still conduct manual verification to ensure that the ancient documents do not contain answers to these questions.
The overall sample pass rate in this process is 95.1\%. For samples that did not meet the requirements, we ask the original annotators to revise them.
}

\textbf{Correctness} involves checking for misunderstandings of the historical documents, ambiguous wording, or violations of annotation guidelines.  We require annotators to conduct cross-checks. 
{Specifically, we assign a secondary annotator (other than the original creator) to act as a verifier, who answers the question relying exclusively on the provided evidence pages. Only when the verifier's answer matches the ground truth is the QA pair deemed to have passed cross-validation. The pass rate at this stage was 82.3\%. For samples with inconsistent judgments, we asked the two expert annotators to discuss them and determine a reasonable answer.}

\textbf{Novelty} refers to the diversity and practical significance of annotation results. If at least two reviewers consider a QA pair to lack innovation or be meaningless, the pair will be deleted. {Ultimately, 28.3\% of the samples are removed because they did not meet the novelty criterion.}

\subsection{Data Standardization}
\label{Data Formatting}
\hypertarget{subsec:Data_Formatting}{} %
To facilitate multifaceted evaluation and ensure the reliability of scoring, we perform data standardization on the collected annotations. Specifically, to facilitate a multifaceted evaluation, the evidence required to answer questions in historical documents is categorized into three types based on their sources: (1) \textbf{Text}: Pure text, specifically referring to the main content of historical documents; (2) \textbf{Layout}: The evidence involves text elements with special layout significance, such as annotations, authors, translators, headers, footers, and headings of various levels in historical documents; (3) \textbf{Visual}: Pure visual elements in historical documents, such as illustrations, seals, font styles, and condition of damage. 
{We use a \textbf{single-label} evidence-source annotation to indicate the primary evidence type required for answering each question. When multiple modalities are involved, annotators determine the label according to explicit hierarchical rules based on the core reasoning dependency of the question.}

Different answer formats necessitate specific evaluation metrics. For example, String types require Average Normalized Levenshtein Similarity (ANLS) to measure the discrepancy between model responses and ground truth ($y^*$). This metric utilizes the Levenshtein distance, denoted as $\text{NL}(\hat{y}, y^*)$, to provide a robust character-level similarity score as defined in equation (1). In contrast, Integer types demand Exact Match (EM) for numerical precision, as expressed in equation (2). Consequently, we categorize the annotated answers into six formats to ensure the rationality of subsequent scoring: \textbf{String}, \textbf{Integer}, \textbf{Option}, \textbf{List}, \textbf{Boolean}, and \textbf{None}.

\begin{equation}
\hspace{2cm} 
\text{ANLS}(\hat{y}, y^*) = 
\begin{cases} 
1 - \dfrac{\text{NL}(\hat{y}, y^*)}{\max(|\hat{y}|, |y^*|)} & \text{if } 1 - \dfrac{\text{NL}(\hat{y}, y^*)}{\max(|\hat{y}|, |y^*|)} \geq \tau \\[10pt]
0 & \text{otherwise}
\end{cases}
\label{eq:anls}
\end{equation}

\begin{equation}
\hspace{2cm} 
\text{EM}(\hat{y}, y^*) = 
\begin{cases}
1 & \text{if } \hat{y} = y^* \\
0 & \text{otherwise}
\end{cases}
\label{eq:em}
\end{equation}

The data standardization work is primarily conducted by the authors, as this stage involves knowledge related to the field of intelligent document analysis. After that, we also invite annotators to conduct a final data check to ensure that our modifications did not alter the original intent of the expert annotators.  

\subsection{Dataset Overview}

The key statistics of \textbf{LongHisDoc} are provided in Table~\ref{tab:table2}. This benchmark consists of 1,012 QA pairs derived from 101 long historical documents spanning 10 categories, with an average of 85.1 pages per document (variance: 2450.86, max: 249 pages). The distribution of documents is depicted in Figure~\ref{fig:fig3}.
The QA pairs are categorized as follows: 546 single-page QA pairs (evidence located on a single page), 196 cross-page QA pairs (evidence spread across two consecutive pages), 172 multi-hop QA pairs (evidence distributed across two or more non-contiguous pages, with an average page span of 21.10 pages and a maximum span of 104 pages), and 98 unanswerable QA pairs (no sufficient supporting evidence available in the document).
These four types of QA pairs evaluate the long-context historical document understanding capabilities of LLMs and LVLMs from complementary angles: localization, cross-page comprehension, multi-hop reasoning, and robustness to hallucination.
For answerable questions, the supporting evidence is sourced from the three types of evidence as described in \hyperlink{subsec:Data_Formatting}{the Data Standardization section} (Text, Layout, Visual), and the answers to all questions are categorized into six types for fine-grained evaluation (see Table~\ref{tab:table2}). Overall, \textbf{LongHisDoc} offers a comprehensive, diverse, and domain-specific benchmark that lays the foundation for the systematic evaluation of long-context historical document understanding capabilities in LLMs and LVLMs. 

\subsection{Data Records}
\textcolor{black}{
The LongHisDoc dataset consists of historical document page images, corresponding optical character recognition (OCR) results, and annotated question--answer (Q\&A) pairs. These components are distributed across two primary directories and one annotation file, as described below.
}
\textcolor{black}{
\begin{itemize}
    \item \texttt{LongHisDoc\_IMG/} contains the historical document page images. Each document is assigned a unique document identifier (Doc ID), and each page is stored as an individual JPEG image named \texttt{page\_<id>.jpg}, where \texttt{<id>} denotes the page number within the document, starting from 1. The directory occupies approximately 1.36GB.
    \item \texttt{OCR\_res/} contains the page-level OCR results, organized into one JSON file per document. Each file is named \texttt{<doc\_id>.json}, where \texttt{<doc\_id>} identifies the corresponding source document. Within each JSON file, the keys are string representations of zero-based page indices, and the associated values contain the OCR results for those pages. Thus, the key \texttt{"0"} corresponds to \texttt{page\_1.jpg}; more generally, the key representing index $k$ corresponds to the image numbered $k+1$. The directory occupies approximately 239MB.
    \item \texttt{LongHisDoc.json} contains the annotated Q\&A pairs and their associated metadata in a single JSON file. The \texttt{doc\_id} field links each Q\&A record to its source document and the corresponding OCR file.
\end{itemize}
}
\textcolor{black}{
Each Q\&A record includes the following fields:
\begin{itemize}
    \item \texttt{doc\_id}: the identifier of the source document.
    \item \texttt{item\_id}: the identifier of the Q\&A item.
    \item \texttt{question} and \texttt{answer}: the question text and its reference answer, respectively.
    \item \texttt{evidence\_pages}: the page numbers containing evidence that supports the reference answer.
    \item \texttt{input\_pages}: the page numbers of the 30 pages selected as input according to the page-selection procedure described in this study.
    \item \texttt{type}: the question category.
    \item \texttt{answer\_type}: the type of the reference answer.
    \item \texttt{evidence\_modal}: the modality of the evidence supporting the answer.
    \item \texttt{book\_name}: the title or bibliographic description of the source document.
    \item \texttt{guji\_type\_zh} and \texttt{guji\_type\_en}: the category of the source document in Chinese and English, respectively.
    \item \texttt{author\_dynasty}: the dynasty associated with the author of the source document.
    \item \texttt{publication\_dynasty}: the dynasty in which the source document was published.
\end{itemize}
}

\textcolor{black}{
The page numbers in \texttt{evidence\_pages} and \texttt{input\_pages} are one-based and correspond directly to the page numbers in the image filenames. To retrieve the corresponding OCR results, each page number must therefore be reduced by one and converted to a string to obtain the JSON key.
}

\textcolor{black}{
The dataset is distributed through the Hugging Face repository under the Creative Commons Attribution--NonCommercial--NoDerivatives 4.0 International (CC BY-NC-ND 4.0) license.
}

{\tiny
\begin{table*}[!t]
\caption{Generalized accuracy scores (\%) on LongHisDoc. There are four types of evidence pages: single-page (SIN), cross-page (CRO), multi-hop (MUL), and unanswerable (UNA). There are six types of answer formats: String (STR), Integer (INT), Option (OPT), List (LIST), Boolean (BOOL), and None (NONE). There are three types of evidence sources: text (TXT), layout (LAY), and visual (VIS). (Think) refers to using the model's thinking mode. The highest scores among models in each section are highlighted in \colorbox{green!20}{\textbf{green}}.}
\label{tab:table3}
\vspace{2pt}
\resizebox{\textwidth}{!}{ %
\begin{tabular}{lccccccccccccccc}
\hline
\multicolumn{1}{c}{\multirow{2}{*}{\textbf{Models}}} &
  \multicolumn{1}{c|}{\multirow{2}{*}{\textbf{Size}}} &
  \multicolumn{4}{c|}{\textbf{Evidence Page}} &
  \multicolumn{6}{c|}{\textbf{Answer Format}} &
  \multicolumn{3}{c|}{\textbf{Evidence Source}} &
  \multirow{2}{*}{\textbf{AVG}} \\ \cline{3-15}
\multicolumn{1}{c}{} &
  \multicolumn{1}{c|}{} &
  SIN &
  CRO &
  MUL &
  \multicolumn{1}{c|}{UNA} &
  STR &
  INT &
  OPT &
  LIST &
  BOOL &
  \multicolumn{1}{c|}{NONE} &
  TXT &
  LAY &
  \multicolumn{1}{c|}{VIS} &
    \\ \hline
\multicolumn{16}{c}{\textit{OCR Results + Large Language Models (LLMs)}} \\ \hline
\multicolumn{16}{l}{\textit{\textcolor{gray!90}{Open-Source Models}}} \\
LLaVA-V1.6-Mistral \cite{li2024llava} &
  \multicolumn{1}{c|}{7B} &
  17.0 &
  12.2 &
  14.4 &
  \multicolumn{1}{c|}{2.0} &
  12.1 &
  11.9 &
  36.1 &
  13.0 &
  49.1 &
  \multicolumn{1}{c|}{2.0} &
  17.5 &
  12.6 &
  \multicolumn{1}{c|}{16.7} &
  14.2 \\
Llama-3.2-Vision-Instruct \cite{meta2024llama3.2} &
  \multicolumn{1}{c|}{11B} &
  47.7 &
  35.0 &
  28.9 &
  \multicolumn{1}{c|}{14.3} &
  39.0 &
  41.5 &
  58.3 &
  36.4 &
  66.7 &
  \multicolumn{1}{c|}{14.3} &
  49.7 &
  35.7 &
  \multicolumn{1}{c|}{24.1} &
  38.8 \\
InternLM3-Instruct \cite{cai2024internlm2} &
  \multicolumn{1}{c|}{8B} &
  49.7 &
  38.6 &
  29.8 &
  \multicolumn{1}{c|}{20.4} &
  41.4 &
  39.8 &
  63.9 &
  41.4 &
  64.9 &
  \multicolumn{1}{c|}{20.4} &
  50.0 &
  40.1 &
  \multicolumn{1}{c|}{26.1} &
  41.3 \\
Qwen3 \cite{yang2025qwen3} &
  \multicolumn{1}{c|}{8B} &
  52.3 &
  41.8 &
  37.3 &
  \multicolumn{1}{c|}{36.7} &
  44.5 &
  40.3 &
  77.8 &
  43.3 &
  79.0 &
  \multicolumn{1}{c|}{36.7} &
  53.7 &
  44.1 &
  \multicolumn{1}{c|}{28.6} &
  46.2 \\
Qwen3 (Think) \cite{yang2025qwen3} &
  \multicolumn{1}{c|}{8B} &
  52.6 &
  47.5 &
  39.5 &
  \multicolumn{1}{c|}{39.8} &
  46.0 &
  42.9 &
  75.0 &
  44.8 &
  86.0 &
  \multicolumn{1}{c|}{39.8} &
  56.6 &
  46.0 &
  \multicolumn{1}{c|}{24.9} &
  48.2 \\
InternVL3 \cite{zhu2025internvl3} &
  \multicolumn{1}{c|}{8B} &
  60.7 &
  47.9 &
  28.7 &
  \multicolumn{1}{c|}{29.6} &
  52.2 &
  39.8 &
  52.8 &
  30.6 &
  73.7 &
  \multicolumn{1}{c|}{29.6} &
  58.2 &
  43.8 &
  \multicolumn{1}{c|}{24.4} &
  49.8 \\
Qwen3 \cite{yang2025qwen3} &
  \multicolumn{1}{c|}{14B} &
  59.7 &
  49.6 &
  36.6 &
  \multicolumn{1}{c|}{30.6} &
  52.2 &
  44.5 &
  72.2 &
  47.4 &
  82.5 &
  \multicolumn{1}{c|}{30.6} &
  61.1 &
  48.8 &
  \multicolumn{1}{c|}{32.3} &
  51.0 \\
Qwen3 (Think) \cite{yang2025qwen3} &
  \multicolumn{1}{c|}{14B} &
  56.3 &
  48.3 &
  41.0 &
  \multicolumn{1}{c|}{52.0} &
  48.6 &
  47.1 &
  72.2 &
  51.1 &
  80.7 &
  \multicolumn{1}{c|}{52.0} &
  56.8 &
  52.3 &
  \multicolumn{1}{c|}{25.6} &
  51.8 \\
Qwen3-VL-Instruct \cite{qwen3-vl} &
  \multicolumn{1}{c|}{8B} &
  62.1 &
  51.4 &
  36.1 &
  \multicolumn{1}{c|}{30.6} &
  53.6 &
  50.8 &
  69.4 &
  48.3 &
  82.5 &
  \multicolumn{1}{c|}{30.6} &
  62.8 &
  50.4 &
  \multicolumn{1}{c|}{34.6} &
  52.6 \\
Qwen3 \cite{yang2025qwen3} &
  \multicolumn{1}{c|}{32B} &
  61.7 &
  52.0 &
  39.9 &
  \multicolumn{1}{c|}{28.6} &
  53.7 &
  50.4 &
  80.6 &
  49.4 &
  82.5 &
  \multicolumn{1}{c|}{28.6} &
  63.1 &
  51.5 &
  \multicolumn{1}{c|}{35.5} &
  52.9 \\
Qwen3 (Think) \cite{yang2025qwen3} &
  \multicolumn{1}{c|}{32B} &
  60.4 &
  55.7 &
  41.4 &
  \multicolumn{1}{c|}{38.8} &
  53.2 &
  47.9 &
  \cellcolor{green!10}\textbf{86.1} &
  54.4 &
  82.5 &
  \multicolumn{1}{c|}{38.8} &
  62.1 &
  54.2 &
  \multicolumn{1}{c|}{32.6} &
  54.2 \\
Qwen3-VL-Instruct \cite{qwen3-vl} &
  \multicolumn{1}{c|}{32B} &
  66.1 &
  58.2 &
  43.9 &
  \multicolumn{1}{c|}{42.9} &
  60.4 &
  49.2 &
  69.4 &
  55.7 &
  86.0 &
  \multicolumn{1}{c|}{42.9} &
  67.4 &
  60.2 &
  \multicolumn{1}{c|}{26.7} &
  58.5 \\
DeepSeek-R1 \cite{guo2025deepseek-r1} &
  \multicolumn{1}{c|}{671B} &
  64.0 &
  60.4 &
  47.0 &
  \multicolumn{1}{c|}{47.6} &
  57.2 &
  52.9 &
  72.2 &
  63.0 &
  87.7 &
  \multicolumn{1}{c|}{47.6} &
  65.0 &
  62.3 &
  \multicolumn{1}{c|}{27.8} &
  58.9 \\
DeepSeek-V3.2 \cite{liu2024deepseek} &
  \multicolumn{1}{c|}{671B} &
  68.7 &
  64.7 &
  46.3 &
  \multicolumn{1}{c|}{40.8} &
  62.1 &
  56.7 &
  77.8 &
  65.1 &
  80.7 &
  \multicolumn{1}{c|}{40.8} &
  69.5 &
  64.7 &
  \multicolumn{1}{c|}{32.2} &
  61.4 \\
\multicolumn{16}{l}{\textit{\textcolor{gray!90}{Proprietary Models}}} \\
GPT-4o \cite{gpt-4o} &
  \multicolumn{1}{c|}{-} &
  60.5 &
  56.5 &
  41.5 &
  \multicolumn{1}{c|}{33.7} &
  54.0 &
  50.0 &
  75.0 &
  53.6 &
  82.5 &
  \multicolumn{1}{c|}{33.7} &
  64.8 &
  52.5 &
  \multicolumn{1}{c|}{28.4} &
  53.9 \\
Qwen3-Max\cite{qwen3max} &
  \multicolumn{1}{c|}{-} &
  63.9 &
  62.4 &
  49.4 &
  \multicolumn{1}{c|}{46.9} &
  58.3 &
  55 &
  72.2 &
  62.4 &
  87.7 &
  \multicolumn{1}{c|}{46.9} &
  67.1 &
  61.9 &
  \multicolumn{1}{c|}{27.8} &
  59.5 \\
GPT-5.2-Chat\cite{GPT5} &
  \multicolumn{1}{c|}{-} &
  69.1 &
  66.9 &
  53.9 &
  \multicolumn{1}{c|}{34.7} &
  62.3 &
  65.0 &
  77.8 &
  67.2 &
  \cellcolor{green!10}\textbf{91.2} &
  \multicolumn{1}{c|}{34.7} &
  72.0 &
  65.7 &
  \multicolumn{1}{c|}{37.3} &
  62.8 \\
Gemini-2.5-Pro \cite{comanici2025gemini} &
  \multicolumn{1}{c|}{-} &
  66.9 &
  61.3 &
  62.4 &
  \multicolumn{1}{c|}{55.1} &
  60.6 &
  62.2 &
  77.8 &
  71.6 &
  89.5 &
  \multicolumn{1}{c|}{55.1} &
  70.2 &
  67.7 &
  \multicolumn{1}{c|}{28.9} &
  63.9 \\
Qwen-3.5-Plus\cite{qwen35blog} &
  \multicolumn{1}{c|}{-} &
  69.3 &
  65.0 &
  54.7 &
  \multicolumn{1}{c|}{\cellcolor{green!10}\textbf{63.3}} &
  64.4 &
  60 &
  77.8 &
  64 &
  86.0 &
  \multicolumn{1}{c|}{\cellcolor{green!10}\textbf{63.3}} &
  72.5 &
  68.4 &
  \multicolumn{1}{c|}{23.2} &
  65.4 \\
Claude-4.6-Opus\cite{claude_opus_4_6} &
  \multicolumn{1}{c|}{-} &
  71.0 &
  68.6 &
  55.8 &
  \multicolumn{1}{c|}{57.1} &
  65.0 &
  64.2 &
  80.6 &
  71.3 &
  84.2 &
  \multicolumn{1}{c|}{57.1} &
  71.4 &
  71.4 &
  \multicolumn{1}{c|}{35.7} &
  66.6 \\
Gemini-3.1-Pro\cite{Gemini_3_1_pro} &
  \multicolumn{1}{c|}{-} &
  \cellcolor{green!10}\textbf{73.7} &
  \cellcolor{green!10}\textbf{73.0} &
  \cellcolor{green!10}\textbf{63.0} &
  \multicolumn{1}{c|}{54.1} &
  \cellcolor{green!10}\textbf{69.3} &
  \cellcolor{green!10}\textbf{69.2} &
  80.6 &
  \cellcolor{green!10}\textbf{72.6} &
  \cellcolor{green!10}\textbf{91.2} &
  \multicolumn{1}{c|}{54.1} &
  \cellcolor{green!10}\textbf{76.3} &
  \cellcolor{green!10}\textbf{73.9} &
  \multicolumn{1}{c|}{\cellcolor{green!10}\textbf{40.5}} &
  \cellcolor{green!10}\textbf{69.9} \\ \hline
\multicolumn{16}{c}{\textit{Page Screenshots + Large Vision-Language Models (LVLMs)}} \\ \hline
\multicolumn{16}{l}{\textit{\textcolor{gray!90}{OpenSource Models}}} \\
mPlug-DocOwl2 \cite{hu2024mplug} &
  \multicolumn{1}{c|}{7B} &
  6.1 &
  7.3 &
  7.7 &
  \multicolumn{1}{c|}{2.0} &
  2.5 &
  5.9 &
  36.1 &
  1.7 &
  42.1 &
  \multicolumn{1}{c|}{2.0} &
  7.1 &
  4.5 &
  \multicolumn{1}{c|}{12.4} &
  6.2 \\
Llama-3.2-Vision-Instruct \cite{meta2024llama3.2} &
  \multicolumn{1}{c|}{11B} &
  8.3 &
  9.6 &
  16.0 &
  \multicolumn{1}{c|}{1.0} &
  4.1 &
  8.5 &
  41.7 &
  6.3 &
  61.4 &
  \multicolumn{1}{c|}{1.0} &
  9.9 &
  6.6 &
  \multicolumn{1}{c|}{23.1} &
  9.1 \\
LLaVA-v1.6-Mistral \cite{li2024llava} &
  \multicolumn{1}{c|}{7B} &
  8.8 &
  7.8 &
  12.9 &
  \multicolumn{1}{c|}{11.2} &
  5.1 &
  7.6 &
  36.1 &
  1.2 &
  57.9 &
  \multicolumn{1}{c|}{11.2} &
  9.9 &
  7.6 &
  \multicolumn{1}{c|}{13.5} &
  9.6 \\
MiniCPM-V-2.6 \cite{yao2024minicpm} &
  \multicolumn{1}{c|}{8B} &
  13.1 &
  13.8 &
  19.1 &
  \multicolumn{1}{c|}{9.2} &
  8.4 &
  10.2 &
  52.8 &
  8.5 &
  71.9 &
  \multicolumn{1}{c|}{9.2} &
  15.0 &
  12.3 &
  \multicolumn{1}{c|}{19.0} &
  13.9 \\
MiniCPM-o-2.6 \cite{yao2024minicpm} &
  \multicolumn{1}{c|}{8B} &
  15.9 &
  8.2 &
  24.1 &
  \multicolumn{1}{c|}{16.3} &
  9.6 &
  14.4 &
  52.8 &
  12.8 &
  64.9 &
  \multicolumn{1}{c|}{16.3} &
  14.4 &
  13.9 &
  \multicolumn{1}{c|}{29.7} &
  15.9 \\
GLM-4V \cite{glm2024chatglm} &
  \multicolumn{1}{c|}{9B} &
  13.9 &
  15.0 &
  17.8 &
  \multicolumn{1}{c|}{25.5} &
  9.3 &
  13.6 &
  44.4 &
  9.7 &
  66.7 &
  \multicolumn{1}{c|}{25.5} &
  15.7 &
  11.4 &
  \multicolumn{1}{c|}{23.7} &
  15.9 \\
InternVL3 \cite{zhu2025internvl3} &
  \multicolumn{1}{c|}{8B} &
  21.6 &
  15.2 &
  21.0 &
  \multicolumn{1}{c|}{10.2} &
  14.0 &
  23.7 &
  55.6 &
  16.1 &
  61.4 &
  \multicolumn{1}{c|}{10.2} &
  19.8 &
  16.8 &
  \multicolumn{1}{c|}{34.3} &
  19.2 \\
InternVL3 \cite{zhu2025internvl3} &
  \multicolumn{1}{c|}{38B} &
  26.3 &
  19.2 &
  28.7 &
  \multicolumn{1}{c|}{2.0} &
  19.4 &
  25.4 &
  50.0 &
  23.4 &
  71.9 &
  \multicolumn{1}{c|}{2.0} &
  22.9 &
  25.0 &
  \multicolumn{1}{c|}{37.3} &
  23.0 \\
Qwen3-VL-Instruct \cite{qwen3-vl} &
  \multicolumn{1}{c|}{8B} &
  45.2 &
  26.3 &
  34.0 &
  \multicolumn{1}{c|}{29.6} &
  34.4 &
  43.3 &
  58.3 &
  38.6 &
  64.9 &
  \multicolumn{1}{c|}{29.6} &
  36.2 &
  42.2 &
  \multicolumn{1}{c|}{40.6} &
  38.1 \\
\quad \textit{+ All Images} &
  \multicolumn{1}{c|}{} &
  36.6 &
  25.9 &
  24.0 &
  \multicolumn{1}{c|}{23.5} &
  29.8 &
  24.2 &
  55.6 &
  31.2 &
  56.1 &
  \multicolumn{1}{c|}{23.5} &
  31.8 &
  31.7 &
  \multicolumn{1}{c|}{33.4} &
  31.1\\

Qwen3-VL-Instruct \cite{qwen3-vl} &
  \multicolumn{1}{c|}{32B} &
  53.9 &
  36.2 &
  44.3 &
  \multicolumn{1}{c|}{40.8} &
  42.5 &
  55.0 &
  61.1 &
  50.2 &
  79.0 &
  \multicolumn{1}{c|}{40.8} &
  47.4 &
  49.9 &
  \multicolumn{1}{c|}{46.6} &
  47.6 \\
\quad \textit{+ All Images} &
  \multicolumn{1}{c|}{} &
  39.2 &
  25.6 &
  23.2 &
  \multicolumn{1}{c|}{25.5} &
  29.3 &
  28.3 &
  52.8 &
  34.9 &
  54.4 &
  \multicolumn{1}{c|}{25.5} &
  35.7 &
  29.4 &
  \multicolumn{1}{c|}{32.1} &
  32.8\\
\textit{\textcolor{gray!90}{Proprietary Models}} &
   \\
GPT-4o \cite{gpt-4o} &
  \multicolumn{1}{c|}{-} &
  31.2 &
  22.7 &
  29.7 &
  \multicolumn{1}{c|}{12.2} &
  20.9 &
  34.8 &
  66.7 &
  32.4 &
  68.4 &
  \multicolumn{1}{c|}{12.2} &
  25.9 &
  27.6 &
  \multicolumn{1}{c|}{49.3} &
  27.4 \\
\quad \textit{+ All Images} &
  \multicolumn{1}{c|}{} &
  11.4 &
  11.5 &
  6.5 &
  \multicolumn{1}{c|}{7.1} &
  10.2 &
  8.4 &
  50 &
  5.82 &
  3.51 &
  \multicolumn{1}{c|}{7.1} &
  11.1 &
  7.9 &
  \multicolumn{1}{c|}{17.5} &
  10.2\\
  
GPT-5.2-Chat\cite{GPT5} &
  \multicolumn{1}{c|}{-} &
  49.5 &
  36.2 &
  33.2 &
  \multicolumn{1}{c|}{17.4} &
  39.1 &
  42.5 &
  63.9 &
  47.0 &
  70.2 &
  \multicolumn{1}{c|}{17.4} &
  45.7 &
  40.2 &
  \multicolumn{1}{c|}{46.3} &
  41.0 \\
\quad \textit{+ All Images} &
  \multicolumn{1}{c|}{} &
  22.8 &
  18.1 &
  13.8 &
  \multicolumn{1}{c|}{20.4} &
  17.9 &
  11.0 &
  36.1 &
  9.8 &
  43.9 &
  \multicolumn{1}{c|}{20.4} &
  22.1 &
  16.5 &
  \multicolumn{1}{c|}{24.4} &
  20.2\\
Gemini-2.5-Pro \cite{comanici2025gemini} &
  \multicolumn{1}{c|}{-} &
  48.0 &
  35.9 &
  44.8 &
  \multicolumn{1}{c|}{54.1} &
  37.5 &
  52.9 &
  63.9 &
  50.6 &
  75.4 &
  \multicolumn{1}{c|}{54.1} &
  45.4 &
  42.0 &
  \multicolumn{1}{c|}{52.6} &
  45.7 \\
\quad \textit{+ All Images} &
  \multicolumn{1}{c|}{} &
  45.5 &
  32.6 &
  41.7 &
  \multicolumn{1}{c|}{41.8} &
  35.7 &
  42.0 &
  66.7 &
  48.9 &
  73.7 &
  \multicolumn{1}{c|}{41.8} &
  41.1 &
  40.6 &
  \multicolumn{1}{c|}{51.4} &
  42.0\\

Claude-4.6-Opus\cite{claude_opus_4_6} &
  \multicolumn{1}{c|}{-} &
  50.4 &
  38.2 &
  43.5 &
  \multicolumn{1}{c|}{41.8} &
  40.0 &
  49.2 &
  69.4 &
  53.7 &
  73.7 &
  \multicolumn{1}{c|}{41.8} &
  42.9 &
  49.9 &
  \multicolumn{1}{c|}{50.4} &
  46.0 \\
\quad \textit{+ All Images} &
  \multicolumn{1}{c|}{} &
  24.9 &
  18.1 &
  22.3 &
  \multicolumn{1}{c|}{18.4} &
  17.6 &
  22.5 &
  58.3 &
  18.2 &
  66.7 &
  \multicolumn{1}{c|}{18.4} &
  22.4 &
  21.5 &
  \multicolumn{1}{c|}{31.0} &
  22.5\\
Qwen-3.5-Plus\cite{qwen35blog} &
  \multicolumn{1}{c|}{-} &
  62.0 &
  43.2 &
  44.8 &
  \multicolumn{1}{c|}{48.0} &
  51.6 &
  54.2 &
  58.3 &
  58.1 &
  77.2 &
  \multicolumn{1}{c|}{48.0} &
  55.2 &
  55.0 &
  \multicolumn{1}{c|}{51.0} &
  54.1 \\
\quad \textit{+ All Images} &
  \multicolumn{1}{c|}{} &
  34.7 &
  31.3 &
  56.3 &
  \multicolumn{1}{c|}{28.4} &
  33.7 &
  38.6 &
  26.3 &
  52.6 &
  53.3 &
  \multicolumn{1}{c|}{28.4} &
  30.0 &
  34.8 &
  \multicolumn{1}{c|}{54.8} &
  37.1\\

Gemini-3.1-Pro\cite{Gemini_3_1_pro} &
  \multicolumn{1}{c|}{-} &
  \cellcolor{green!10}\textbf{66.9} &
  \cellcolor{green!10}\textbf{61.5} &
  \cellcolor{green!10}\textbf{49.3} &
  \multicolumn{1}{c|}{\cellcolor{green!10}\textbf{67.4}} &
  \cellcolor{green!10}\textbf{60.8} &
  \cellcolor{green!10}\textbf{56.7} &
  \cellcolor{green!10}\textbf{75.0} &
  \cellcolor{green!10}\textbf{61.6} &
  \cellcolor{green!10}\textbf{84.2} &
  \multicolumn{1}{c|}{\cellcolor{green!10}\textbf{67.4}} &
  \cellcolor{green!10}\textbf{66.0} &
  \cellcolor{green!10}\textbf{60.2} &
  \multicolumn{1}{c|}{\cellcolor{green!10}\textbf{53.6}} &
  \cellcolor{green!10}\textbf{62.9} \\
\quad \textit{+ All Images} &
  \multicolumn{1}{c|}{} &
  51.4 &
  38.4 &
  35.5 &
  \multicolumn{1}{c|}{31.6} &
  44.1 &
  39.2 &
  55.6 &
  45.6 &
  68.4 &
  \multicolumn{1}{c|}{31.6} &
  48.5 &
  44.0 &
  \multicolumn{1}{c|}{38.6} &
  44.3 \\ \hline
\end{tabular}
}
\vspace{-3mm}
\end{table*}
}

\begin{figure*}[]
  \centering
  \includegraphics[scale=0.9]{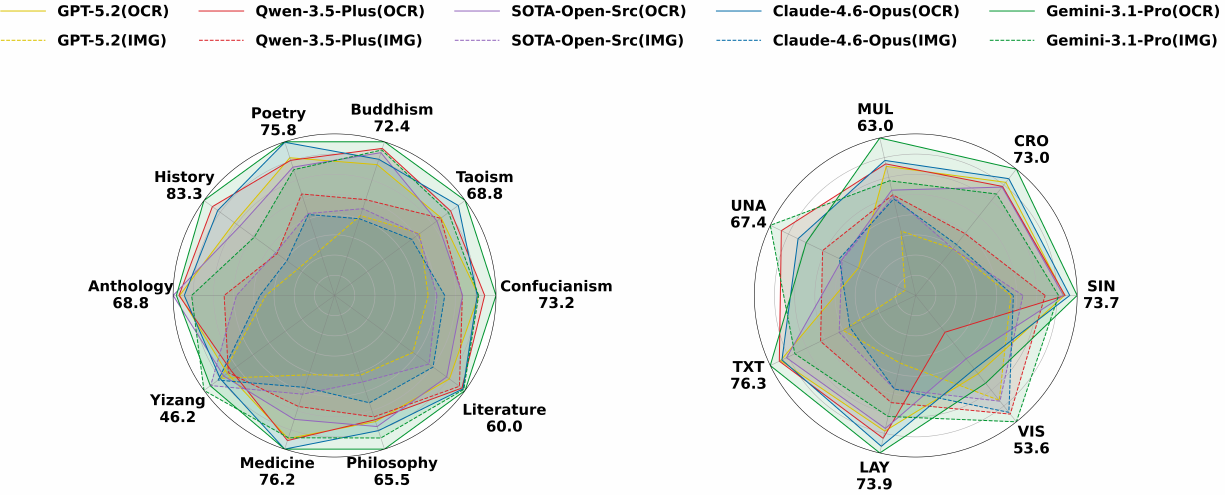}
  \caption{Multi-faceted analysis on various document types, evidence pages and evidence sources. Zoom in for better view.}
  \label{fig:Fine-grained}
  \vspace{-3mm}
\end{figure*}

\section{Evaluation}
With a view to verifying the utility of the LongHisDoc dataset in evaluating the ability of Large Language Models (LLMs) and Large Vision-Language Models (LVLMs) to understand long Chinese historical documents, we conducted comprehensive technical validation experiments on representative LLMs and LVLMs, while also testing various modal retrievers. Furthermore, we performed end-to-end RAG evaluation by pairing each retrieval method with state-of-the-art generation models. These experiments fully demonstrate the performance of existing models in understanding long historical documents, providing a valuable reference for the development of LLMs, LVLMs, and RAG systems.

\subsection{Evaluation Protocol}
Following MMLongBench-Doc \cite{ma2024mmlongbench} and LongDocURL \cite{deng2025longdocurl}, we adopt a three-stage scoring scheme: \textit{response generation}, \textit{answer extraction}, and \textit{score calculation}.
Specifically, during the response generation phase, we require the model to answer user questions rigorously and concisely based on the content of historical documents. In the answer extraction phase, we standardize the model results into simplified Chinese and then use GPT-4o to extract the answers from the free-form model responses into the specified format. 
For the six answer formats in our benchmark, we design dedicated prompts to guide the extractor in extracting answers that match the expected format. 
In the score calculation phase, we apply appropriate rule-based scoring methods for each answer type to compute generalized accuracy scores. 

\subsection{Experimental Setup}
\hypertarget{subsec:Experimental_Setup}{} %
\label{Experimental Setup}
\textbf{Input Modality.} 
We categorize our experiments into two types: \textbf{text-input} and \textbf{image-input}. For the text-input setting, we input the OCR results (generated by a YOLOv7\cite{yolov7}-based detector and a Vision Transformer\cite{vit} (ViT)-based recognizer) sourced from HisDoc-1B into LLMs. 
For the image-input setting, we input document pages into LVLMs. 

\noindent\textbf{Input Paradigm.}
The documents in LongHisDoc have a maximum of 249 pages, and most models are unable to process the full context due to GPU memory or interface limitations. Therefore, referring to the approach used in previous benchmarks~\cite{ma2024mmlongbench,deng2025longdocurl}, we adopt a \textbf{cut-off} paradigm for evaluating current models. Specifically, for overly long text inputs, we truncate the content to the maximum input length supported by the model. For image inputs, we cut 30 pages surrounding the answer evidence, keeping the selected page segments as contiguous as possible. 
{To evaluate models under a more realistic long-document setting, we have additionally conducted full-document input evaluation. Specifically, we provide the complete historical document as input to advanced open-source models and to some accessible proprietary models that support sufficiently long visual contexts.}

\noindent\textbf{Other Configurations.}
We evaluate the proprietary models via API resources, whereas the open-source models (except for the DeepSeek model, which is accessed via API due to its large number of parameters) are assessed on NVIDIA A6000 GPUs. To minimize variance, we set the temperature parameter to 0.0 during the generation of free-form responses and answer extraction.

\subsection{Main Results}
\begin{figure*}[ht]
\centering
\subfigure{%
    \resizebox*{8.6cm}{!}{\includegraphics{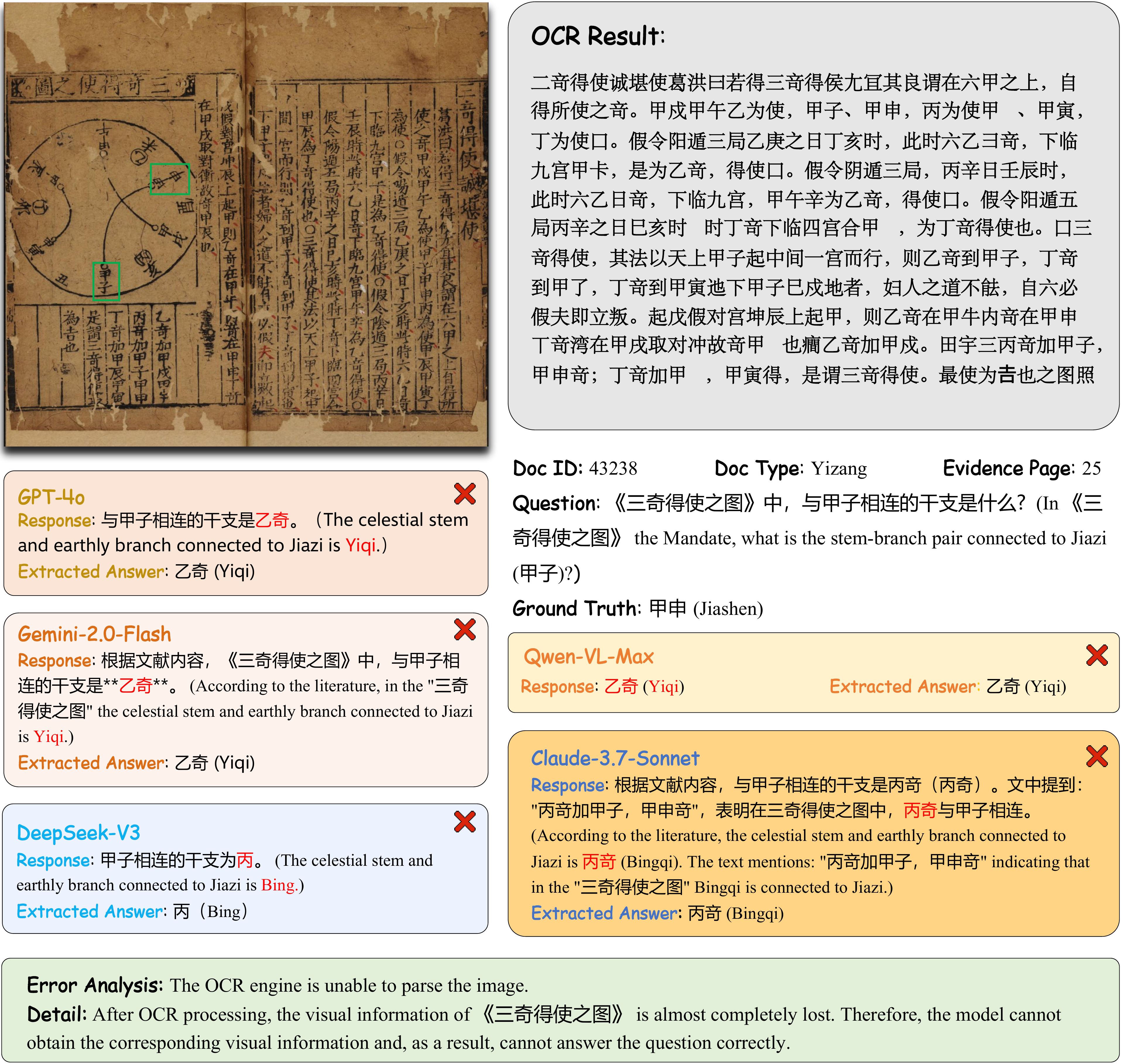}}} 
    \label{fig:panel-1-a}
\hspace{0.3cm}
\subfigure{%
    \resizebox*{7.3cm}{!}{\includegraphics{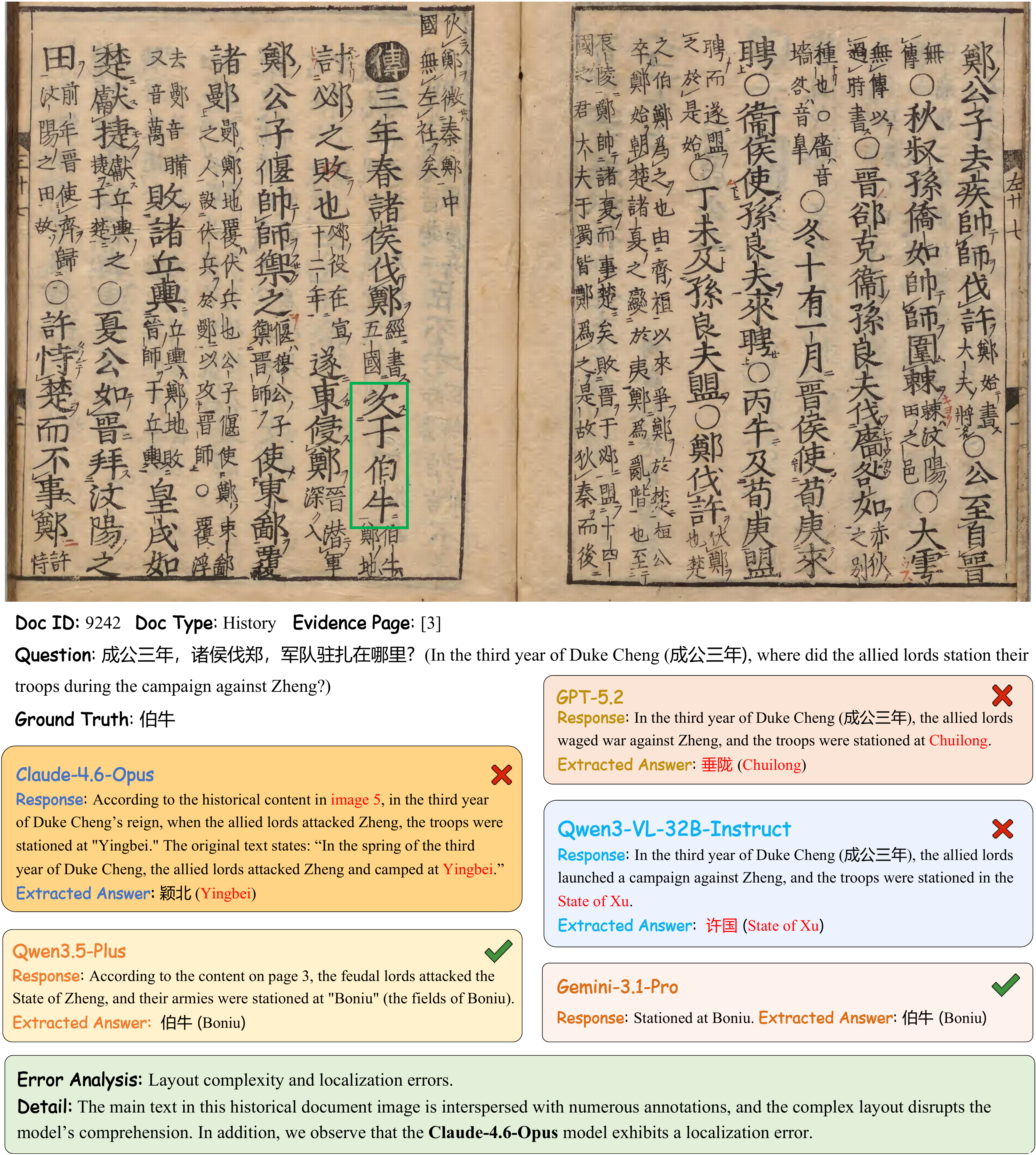}}} \label{fig:panel-1-b}

\caption{ Visualization of test results. \textbf{Left}: Incorrect response attributed to OCR parsing failure; \textbf{Right}: Complex layouts of historical documents with interlinear annotations pose challenges to model understanding. Zoom in for better view.}
\label{fig:visualize_results}
\vspace{-3mm}
\end {figure*}

As shown in Table \ref{tab:table3}, we calculate the generalized accuracy score to evaluate model capabilities. Based on the evaluation results, we draw the following conclusions: 
{\textbf{(1) Long-context historical document understanding remains a challenging and unresolved task for LVLMs.} In the image-input setting, the best-performing LVLM, Gemini-3.1-Pro, achieves an accuracy score of only 62.9\%. When taking all images of the entire document as input, its overall accuracy further decreases to 44.3\%.} Among the evaluated open-source LVLMs, only Qwen3-VL-Instruct (32B and 8B) achieve scores above 30\% in the cut-off setting. We attribute this to the nature of historical documents, which often feature diverse fonts, irregular layouts, and low-quality images. These factors make it challenging for current LVLMs to directly process historical document images.
\textbf{(2) Introducing high-quality OCR results generally improves model performance. } In the text-input setting, the best-performing model is Gemini-3.1-Pro, which achieves the highest performance with an accuracy score of 69.9\%. Compared to the image-input setting, high-quality OCR results significantly enhance LVLM performance, despite the fact that some OCR-induced noise still exists. (The impact of OCR-induced noise is analyzed in detail in Figure~\ref{fig:visualize_results}.) 
Notably, GPT-5.2, Claude-4.6-Opus, Gemini-3.1-Pro, and Qwen-3.5-Plus show performance improvements of 21.8\%, 20.6\%, 7.0\%, and 11.3\%, respectively.
\textbf{(3) The use of thinking mechanisms is effective for historical document understanding.} In our evaluation results, all sizes of the Qwen3 model demonstrate overall performance improvements when the thinking mode is enabled, particularly on question types that require inference and information integration, such as multi-hop and unanswerable questions.

\begin{figure*}[htbp]
\centering
\subfigure{%
    \resizebox*{7.6cm}{!}{\includegraphics{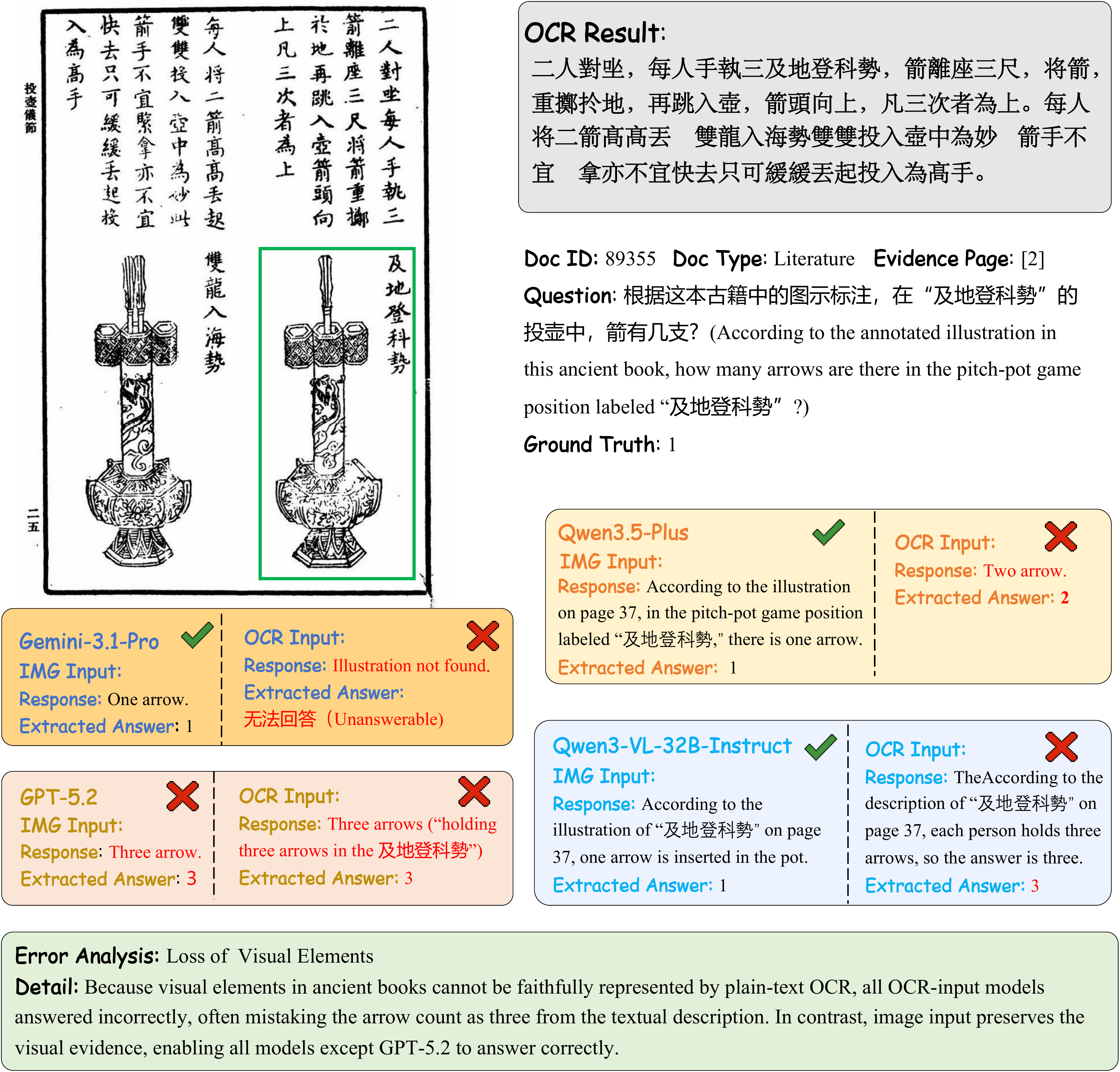}}} 
    \label{fig:panel-2-a}
\hspace{0.3cm}
\subfigure{%
    \resizebox*{8.7cm}{!}{\includegraphics{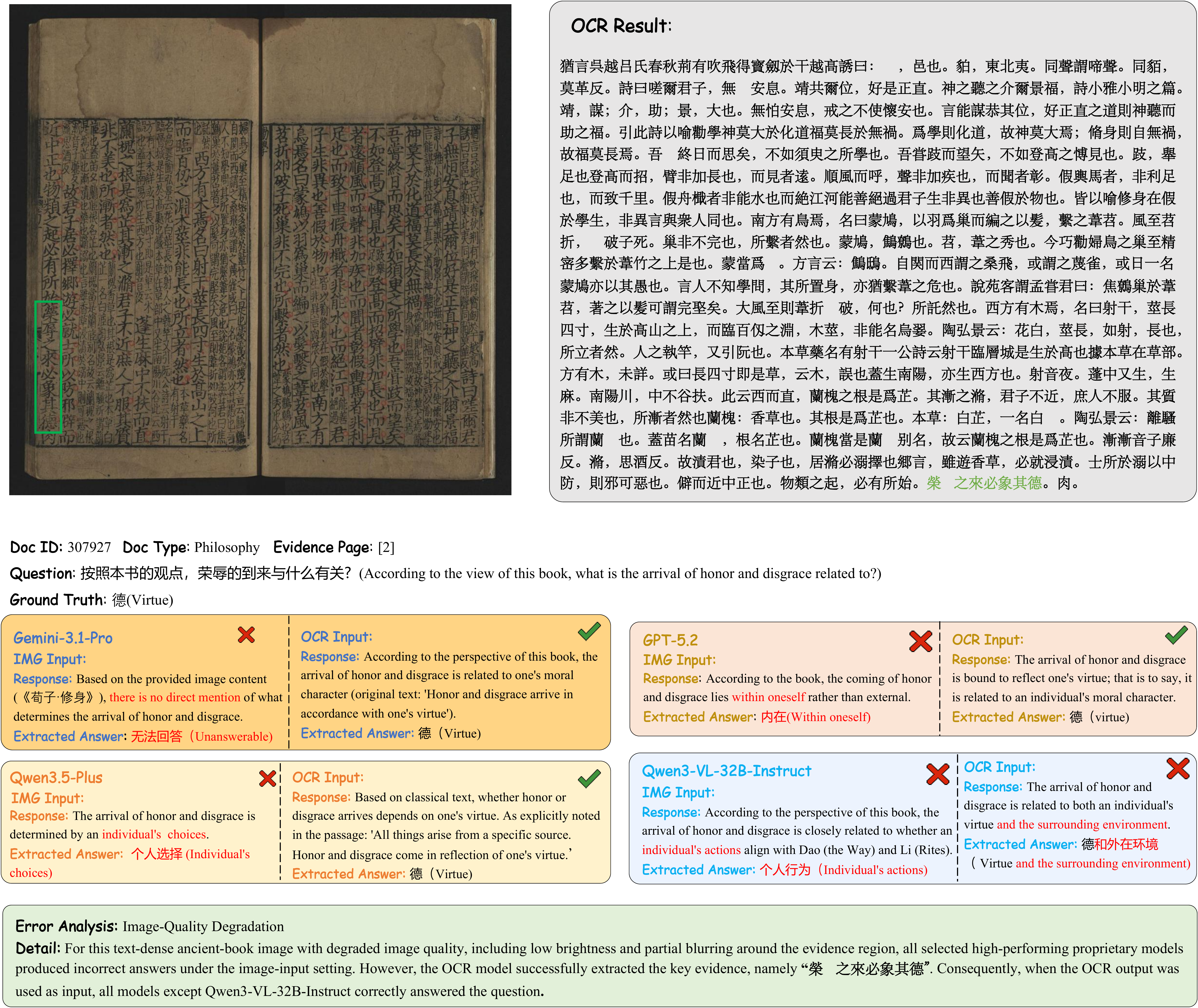}}} \label{fig:panel-2-b}

\caption{{Visualization comparison between Text-Input and Image-Input . \textbf{Left}: Text-input models fail due to the loss of visual elements in OCR outputs. \textbf{Right}: Image-input models fail due to image-quality degradation.}}
\label{fig:visualize_results_ocr_img}
\vspace{-3mm}
\end {figure*}

\subsection{Multi-faceted Analysis of LongHisDoc}
To comprehensively benchmark the capabilities of LLMs and LVLMs on LongHisDoc, we conduct a multi-faceted analysis across three dimensions: historical document types, evidence sources, and QA pair categories.

\noindent\textbf{Historical Document Type.}
As shown in Figure~\ref{fig:Fine-grained}, LVLMs and LLMs exhibit distinct performance across various document types. Our findings are as follows:  
(1) Under the text-input setting, all tested models consistently perform worst on historical documents in the \textbf{Yizang} category. We attribute this to the presence of numerous visual elements in \textbf{Yizang} documents, such as hexagrams and Bagua diagrams. These elements cannot be parsed by OCR engines, resulting in highly noisy OCR results that degrade model performance. The specific visualization samples are presented in Figure~\ref{fig:visualize_results} left. 
(2) Under the image-input setting, the \textbf{History} category exhibits the most significant performance gap compared to the text-input setting. Based on error example analysis, we attribute this to the relatively complex layout of such documents. Specifically, \textbf{History} documents often contain dense interlinear annotations embedded within the main text. Without the aid of OCR, it remains challenging for LVLMs to accurately interpret such intricate layout structures in historical documents. The specific visualization samples are presented in Figure~\ref{fig:visualize_results} right. 

\noindent\textbf{Evidence Source.}
We categorize the QA pairs based on the source of evidence and present fine-grained results in Table~\ref{tab:table3} and Figure~\ref{fig:Fine-grained}. The OCR engine's powerful character recognition and layout analysis capabilities significantly enhance the model's performance on questions supported by text and layout modality evidence. However, for questions involving visual elements, performance under the text-input setting is generally lower than that under the image-input setting. This highlights the limitations of the OCR-based method in processing visual components such as figures and seals in historical documents.

\noindent\textbf{QA Pairs Type.}
As shown in Table~\ref{tab:table3} and Figure~\ref{fig:Fine-grained}, under the text-input setting, most models perform best on single-page QA pairs, followed by cross-page and then multi-hop pairs. 
Surprisingly, under the image-input setting, most evaluated LVLMs achieve higher scores on multi-hop QA than cross-page QA. This may be attributed to the unique reading order of Chinese historical documents, specifically the traditional right-to-left vertical layout. In this format, a text sequence is typically truncated at the bottom-left of the former page and resumes at the top-right of the subsequent page. Unlike modern documents, this specific layout creates a unique spatial gap that causes LVLMs to perform poorly when processing information truncated across pages. While multi-hop QA involves evidence distributed across non-contiguous pages, the content on each page remains complete.
For unanswerable questions, accuracy under the text-input setting is higher than that under the image-input setting, indicating that current LVLMs are more prone to hallucination when interpreting document screenshots.

\noindent\textbf{{Image-Input and OCR-Input.}}
{On LongHisDoc, LVLMs with text input generally achieve substantially better overall performance than those with image input, even though our image-input setting already adopts a cut-off paradigm to reduce the difficulty of processing long visual contexts. However, through case-level error analysis, we find that the two input paradigms exhibit distinct advantages and limitations in long-context ancient document understanding. Specifically, we categorize the errors under text-input and image-input settings as follows:
\begin{itemize}
    \item \textbf{Errors of OCR-based LLMs primarily arise from three sources:}
    \begin{enumerate}
        \item \textit{OCR Recognition Errors}, including the misrecognition or hallucination of rare and obscure historical characters.
        \item \textit{Reading Order Errors}, caused by incorrect text sequencing in complex vertical or multi-column layouts.
        \item \textit{Loss of Visual Elements}, where non-textual elements, such as illustrations, seals, and structural charts, cannot be faithfully represented in plain-text OCR outputs.
    \end{enumerate}
    \item \textbf{Failure cases in the VLM setting primarily arise from three sources:}
    \begin{enumerate}
        \item \textit{Cross-Page and Multi-Hop Reasoning Failures}, where the model struggles to aggregate evidence distributed across multiple high-resolution pages.
        \item \textit{Layout and Reading-Order Confusion}, caused by complex visual structures in historical documents.
        \item \textit{Image-Quality Degradation}, including low resolution, faded ink, physical damage, and other artifacts in historical materials.
    \end{enumerate}
\end{itemize}
As shown in Figure~\ref{fig:visualize_results_ocr_img}, we present two representative cases: text-input models fail due to the loss of visual elements in OCR outputs, while image-input models fail due to image-quality degradation. 
}

\begin{figure}[h]
  \centering
  \includegraphics[scale=0.55]{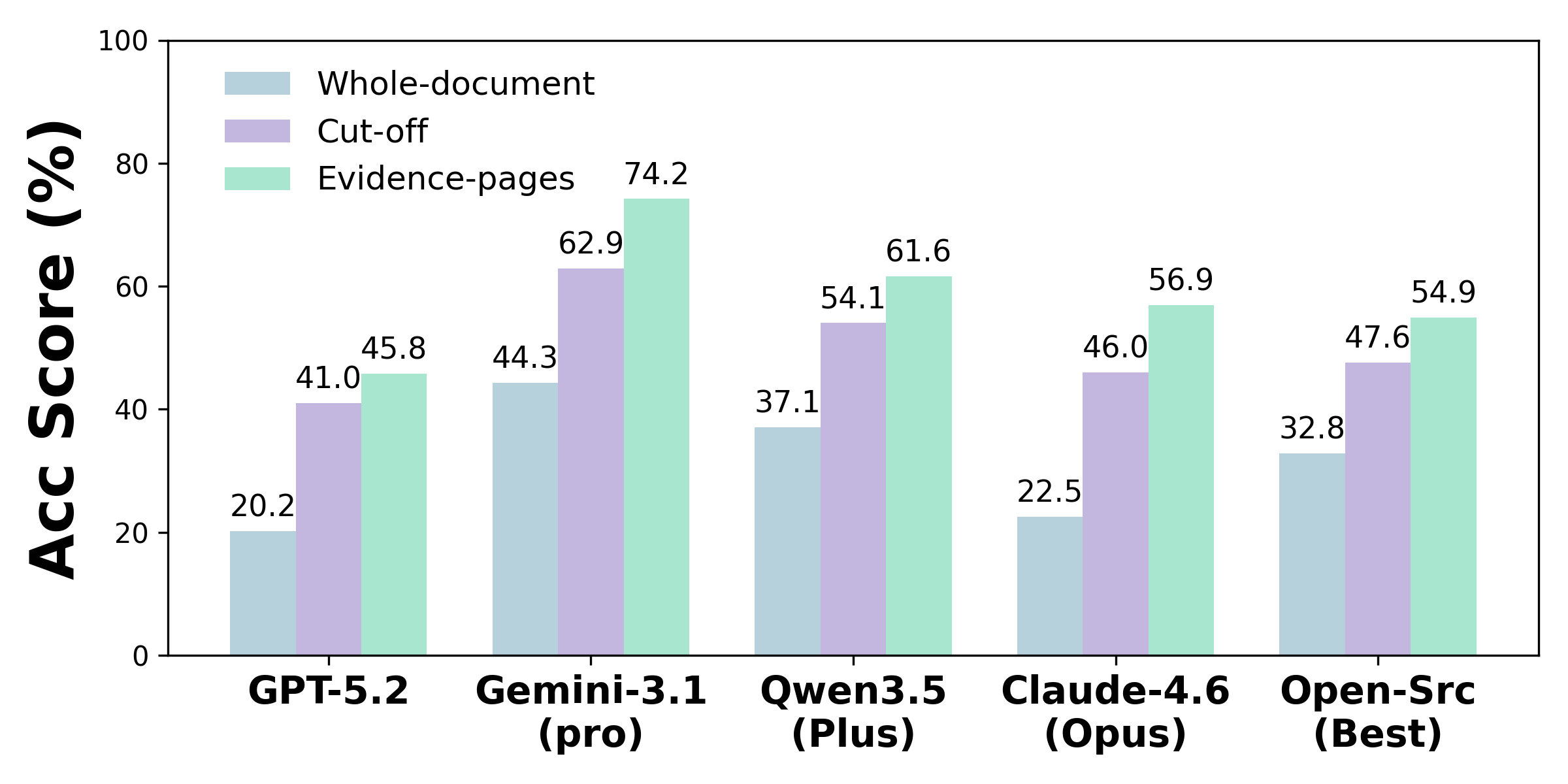}
  \caption{Comparison among different input lengths on LongHisDoc benchmark.}
  \label{fig:fig5}
  \vspace{-4mm}
\end{figure}

\subsection{{Ablation Study}}
\subsubsection{{Input Length Ablation Study}}

With the intention of demonstrating the capability of LVLMs in understanding entire historical documents and to investigate the specific impact of different input length settings on model performance, we select a group of high-performing proprietary models along with the best-performing open-source model, and conduct an ablation study under three settings: \textbf{whole-document}, \textbf{cut-off} (as described in \hyperlink{subsec:Experimental_Setup}{the Experimental Setup section}), and \textbf{evidence-pages} (only input evidence pages). The experimental results are shown in Figure~\ref{fig:fig5}.
On the one hand, current LVLMs still exhibit limited capabilities in understanding one or several pages of historical documents. Even with only the evidence pages fed into the model, the best-performing Gemini-3.1-Pro achieves just 74.2\% accuracy. On the other hand, as input length increases, the performance of all LVLMs consistently declines. This degradation arises not only from redundant information but also from the fact that some LVLMs fail to handle long inputs (e.g., historical documents in LongHisDoc that exceed 200 pages), resulting in response refusal.

\begin{table}[h]
\caption{{Performance of generation models on LongHisDoc using outputs from different OCR models.}}
\label{tab:OCR_ABL}
\resizebox{\textwidth}{!}{
\begin{tabular}{ll|l|cccc|cccccc|ccc|c}
\hline
\multirow{2}{*}{\textbf{OCR Models}} & \multirow{2}{*}{\textbf{size}} & \multirow{2}{*}{\textbf{Generation Models}} & \multicolumn{4}{c|}{\textbf{Evidence Page}} & \multicolumn{6}{c|}{\textbf{Answer Format}} & \multicolumn{3}{l|}{\textbf{Evidence Source}} & \multicolumn{1}{l}{\multirow{2}{*}{\textbf{AVG}}} \\ \cline{4-16}
 &  &  & \multicolumn{1}{l}{SIN} & \multicolumn{1}{l}{CRO} & \multicolumn{1}{l}{MUL} & \multicolumn{1}{l|}{UNA} & Str & Int & Enum & LIST & Bool & None & TXT & LAY & VIS & \multicolumn{1}{l}{} \\ \hline
\multirow{4}{*}{DeepSeek-OCR-2 \cite{wei2026deepseek}} & \multirow{4}{*}{3B} & Qwen3-VL-8B & 38.8 & 29.6 & 31.3 & 35.7 & 30.0 & 37.5 & 58.3 & 36.4 & 68.4 & 35.7 & 40.8 & 30.1 & 30.4 & 35.5 \\
 &  & Qwen3-32B & 38.8 & 33.2 & 35.1 & 38.8 & 32.2 & 33.3 & 63.9 & 35.6 & 77.2 & 38.8 & 43.9 & 33.4 & 17.2 & 37.1 \\
 &  & Qwen3-VL-32B & 41.8 & 32.6 & 29.3 & 44.9 & 35.5 & 32.5 & 63.9 & 31.5 & 64.9 & 44.9 & 46.4 & 31.2 & 19.3 & 38.2 \\
 &  & Gemini-3.1-Pro & 53.3 & 44.8 & 42.5 & 59.2 & 45.7 & 42.5 & 69.4 & 52.7 & 80.7 & 59.2 & 55.2 & 46.6 & 33.2 & 50.4 \\ \hline
\multirow{4}{*}{dots.ocr \cite{li2025dots}} & \multirow{4}{*}{1.7B} & Qwen3-VL-8B & 58.5 & 53.5 & 38.3 & 29.6 & 53.2 & 45.0 & 66.7 & 50.1 & 75.4 & 29.6 & 59.4 & 51.8 & 33.8 & 51.3 \\
 &  & Qwen3-32B & 62.6 & 54.8 & 52.0 & 36.7 & 56.1 & 54.2 & 75.0 & 60.3 & 84.2 & 36.7 & 65.8 & 59.3 & 25.4 & 56.8 \\
 &  & Qwen3-VL-32B & 62.6 & 59.2 & 50.6 & 41.8 & 56.4 & 55.8 & 75.0 & 61.2 & 86.0 & 41.8 & 65.3 & 60.7 & 28.5 & 57.9 \\
 &  & Gemini-3.1-Pro & 67.8 & 68.4 & 58.0 & 51.0 & 62.1 & 68.3 & 75.0 & 70.2 & 86.0 & 51.0 & 70.3 & 69.3 & 34.4 & 64.6 \\ \hline
\multirow{4}{*}{Hunyuan-OCR \cite{team2025hunyuanocr}} & \multirow{4}{*}{1B} & Qwen3-VL-8B & 49.1 & 45.6 & 30.4 & 54.1 & 47.8 & 25.8 & 61.1 & 30.0 & 79.0 & 54.1 & 51.2 & 41.0 & 29.0 & 45.7 \\
 &  & Qwen3-32B & 44.8 & 44.6 & 33.3 & \cellcolor{green!10}\textbf{64.3} & 42.8 & 33.3 & 58.3 & 34.9 & 68.4 & \cellcolor{green!10}\textbf{64.3} & 48.9 & 40.5 & 21.0 & 44.7 \\
 &  & Qwen3-VL-32B & 58.2 & 57.9 & 37.7 & 50.0 & 55.2 & 51.7 & 66.7 & 36.1 & 84.2 & 50.0 & 61.0 & 53.3 & 26.2 & 53.9 \\
 &  & Gemini-3.1-Pro & 69.8 & 67.4 & 59.2 & 54.1 & 63.8 & 62.5 & 83.3 & 71.9 & 91.2 & 54.1 & 70.9 & 69.4 & \cellcolor{green!10}\textbf{42.4} & 66.0 \\ \hline
\multirow{4}{*}{GLM-OCR \cite{duan2026glm}} & \multirow{4}{*}{0.9B} & Qwen3-VL-8B & 54.5 & 46.6 & 37.7 & 30.6 & 46.7 & 46.7 & 63.9 & 50.0 & 75.4 & 30.6 & 56.8 & 46.0 & 29.7 & 47.8 \\
 &  & Qwen3-32B & 60.9 & 52.5 & 50.2 & 36.7 & 54.2 & 55.8 & 75.0 & 54.7 & 82.5 & 36.7 & 63.9 & 57.1 & 25.4 & 55.1 \\
 &  & Qwen3-VL-32B & 56.2 & 54.3 & 42.6 & 37.8 & 50.6 & 44.2 & 80.6 & 56.5 & 73.7 & 37.8 & 59.5 & 51.8 & 29.0 & 51.7 \\
 &  & Gemini-3.1-Pro & 67.2 & 65.5 & 59.4 & 55.1 & 60.5 & 64.2 & \cellcolor{green!10}\textbf{86.1} & 69.8 & \cellcolor{green!10}\textbf{93.0} & 55.1 & 69.9 & 66.2 & 41.2 & 64.4 \\ \hline
\multirow{4}{*}{Yolov7+ViT(ours)} & \multirow{4}{*}{} & Qwen3-VL-8B & 62.1 & 51.4 & 36.1 & 30.6 & 53.6 & 50.8 & 69.4 & 48.3 & 82.5 & 30.6 & 62.8 & 50.4 & 34.6 & 52.6 \\
 &  & Qwen3-32B & 60.4 & 55.7 & 41.4 & 38.8 & 53.2 & 47.9 & \cellcolor{green!10}\textbf{86.1} & 54.4 & 82.5 & 38.8 & 62.1 & 54.2 & 32.6 & 54.2 \\
 &  & Qwen3-VL-32B & 66.1 & 58.2 & 43.9 & 42.9 & 60.4 & 49.2 & 69.4 & 55.7 & 86.0 & 42.9 & 67.4 & 60.2 & 26.7 & 58.5 \\
 &  & Gemini-3.1-Pro & \cellcolor{green!10}\textbf{73.7} & \cellcolor{green!10}\textbf{73.0} & \cellcolor{green!10}\textbf{63.0} & 54.1 & \cellcolor{green!10}\textbf{69.3} & \cellcolor{green!10}\textbf{69.2} & 80.6 & \cellcolor{green!10}\textbf{72.6} & 91.2 & 54.1 & \cellcolor{green!10}\textbf{76.3} & \cellcolor{green!10}\textbf{73.9} & 40.5 & \cellcolor{green!10}\textbf{69.9} \\ \hline
\end{tabular}
}
\end{table}

\subsubsection{{OCR Quality Analysis and Ablation Study}}
{
To investigate how outputs generated by different types of OCR models affect the final performance of models under text-input settings, we conducted an OCR Result Ablation Study. Specifically, we applied four mainstreamLVLM-based OCR models, including DeepSeek-OCR-2\cite{wei2026deepseek}, dots.ocr\cite{li2025dots}, Hunyuan-OCR\cite{team2025hunyuanocr}, and GLM-OCR\cite{duan2026glm}, to perform OCR on each historical-document image in the LongHisDoc dataset. The OCR outputs produced by different models were then fed into four strong-performing models on LongHisDoc to conduct the ablation analysis.
The results are reported in Table~\ref{tab:OCR_ABL}. Among the evaluated LVLM-based OCR models, dots.ocr achieves the best average downstream QA performance, with a score of 57.7\%. Nevertheless, this result remains below that obtained using our historical-document OCR pipeline based on YOLOv7+ViT, which achieves 58.8\%. 
This indicates that current domain-specific OCR models designed for ancient texts still have advantages in long historical document scenarios.
}

{
Our qualitative analysis further reveals several challenges faced by current LVLM-based OCR models in this setting. For example, DeepSeek-OCR-2 occasionally produces text in a left-to-right reading order, whereas the historical documents in our dataset require a different reading order. In addition, all evaluated LVLM-based OCR models exhibit occasional hallucinated content and repetitive generation. Although these issues may occur on only a small number of pages, they can disrupt the textual continuity and structural integrity of long documents, thereby adversely affecting downstream QA performance.
}

{
To further investigate the potential performance gains achievable through improved OCR quality, we conducted an additional evaluation using human-verified OCR annotations. Specifically, human annotators carefully verified and corrected the automated OCR annotations of 10 randomly sampled ancient books from LongHisDoc, covering a total of 826 images. The resulting human-verified annotations were treated as Ground Truth (GT). We first evaluated the quality of the original automated OCR annotations against this GT, as summarized in Table~\ref{tab:table_val}. The evaluation metrics include character location (F1-score at 0.7 IoU), character recognition (top-1 accuracy), character arrangement (ARDs), and text punctuation (F1-score). We then evaluated four representative models on this subset using either the original automated OCR annotations or the corresponding human-verified GT as input. As shown in Table~\ref{tab:table_ocr_manual}, all four models consistently achieve better performance with the human-verified GT, with gains ranging from 0.31 to 1.27 percentage points. These results indicate that improving OCR accuracy and structural fidelity can positively contribute to long historical document understanding. However, the relatively modest improvements also suggest that OCR quality is not the sole or dominant bottleneck under the current setting, and that other challenges, such as long-context reasoning and document-level semantic understanding, also substantially affect downstream performance.
}

{
In addition to the human-verified GT evaluation, we further investigated the use of texts from existing historical text corpora as an external baseline. However, due to the large number and diversity of Chinese historical books, some of the books included in LongHisDoc do not have corresponding manually curated or human-verified textual versions available in existing online historical-document databases. Among the books for which such external texts were available, we carefully compared their editions with those in LongHisDoc and, despite extensive searching, identified only 12 books whose editions were broadly consistent with those in our dataset. Their digitized page organization and layout, however, were not necessarily identical. We used the corresponding texts from these external corpora as alternative inputs, denoted as External-TC, and evaluated the same QA models. The results are reported in Table~\ref{tab:table_ocr_tc}.
Although the same book editions were matched, we observed discrepancies in page ordering, page layout, and intra-page reading sequence between the externally collected texts and the document images in LongHisDoc. Consequently, replacing the original automated OCR annotations with External-TC does not consistently improve downstream QA performance, with performance differences ranging from -1.62 to +1.47 percentage points across the evaluated models. Together with the human-verified GT experiment, these results show that directly correcting OCR errors while preserving the original document structure and reading order leads to consistent, albeit modest, improvements, whereas texts obtained from external corpora do not necessarily provide similar gains due to discrepancies in document organization and structural correspondence. This further suggests that accurate textual content alone does not fully determine downstream performance; preserving the correspondence between textual content and the original document structure is also important for long historical document understanding.
}

\begin{table}[h]
\centering
\caption{\label{tab:data_val}{Quality evaluation of automated OCR annotations against human-verified ground truth.}}
\label{tab:table_val}
\begin{tabular}{c c c c}
\hline
Character Location(\%)~$\uparrow$ & Character Annotation(\%)~$\uparrow$ & Character Arrangement~$\downarrow$ & Text Punctuation(\%)~$\uparrow$ \\
\hline
98.57 & 98.43 & 1.92 & 89.34\\
\hline
\end{tabular}
\end{table}

\begin{table}[htbp]
\centering
\begin{minipage}[t]{0.5\textwidth}
\centering
\caption{{Performance with human-verified ground truth (Human-GT).}}
\begin{tabular}{l|c c}
\hline
Model & OCR (\%) & Human-GT (\%) \\
\hline
GPT-5.2         & 63.62 & 63.93\textcolor{green!70!black}{(+0.31)} \\
Gemini-3.1-Pro  & 70.34 & 71.29\textcolor{green!70!black}{(+0.95)} \\
Claude-4.6-Opus & 67.38 & 68.51\textcolor{green!70!black}{(+1.13)} \\
Qwen-3.5-Plus    & 66.21 & 67.48\textcolor{green!70!black}{(+1.27)} \\
\hline
\end{tabular}
\label{tab:table_ocr_manual}
\end{minipage}%
\hfill
\begin{minipage}[t]{0.5\textwidth}
\centering
\caption{{Performance using texts matched from external text corpora (External-TC).}}
\begin{tabular}{l|c c}
\hline
Model & OCR (\%) & External-TC (\%) \\
\hline
GPT-5.2         & 57.39 & 56.41\textcolor{red!70!black}{(-0.98)} \\
Gemini-3.1-Pro  & 64.74 & 63.12\textcolor{red!70!black}{(-1.62)} \\
Claude-4.6-Opus & 63.48 & 62.91\textcolor{red!70!black}{(-0.57)} \\
Qwen-3.5-Plus    & 55.66 & 57.13\textcolor{green!70!black}{(+1.47)} \\
\hline
\end{tabular}
\label{tab:table_ocr_tc}
\end{minipage}
\end{table}

{\tiny
\begin{table*}[h]
\caption{Retriever performance on LongHisDoc using top-k Recall metrics.}
\label{tab:table4}
\resizebox{\textwidth}{!}{ %
\begin{tabular}{lccccccccccc}
\hline
\multicolumn{1}{c}{\multirow{2}{*}{\textbf{Method}}} &
  \multicolumn{1}{c|}{\multirow{2}{*}{\textbf{Size}}} &
  \multicolumn{2}{c|}{\textbf{SIN}} &
  \multicolumn{2}{c|}{\textbf{CRO}} &
  \multicolumn{2}{c|}{\textbf{MUL}} &
  \textbf{TXT} &
  \textbf{LAY} &
  \multicolumn{1}{c|}{\textbf{VIS}} &
  \textbf{AVG} \\ \cline{3-12} 
\multicolumn{1}{c}{} &
  \multicolumn{1}{c|}{} &
  top 1 &
  \multicolumn{1}{c|}{top 5} &
  top 2 &
  \multicolumn{1}{c|}{top 5} &
  top 5 &
  \multicolumn{1}{c|}{top 25} &
  \multicolumn{3}{c|}{top 5} &
  \multicolumn{1}{c}{top 5} \\ \hline
\multicolumn{12}{c}{\textit{\textbf{Text-based Method}}} \\ \hline
SBERT \cite{reimers2019sentence}&
  \multicolumn{1}{c|}{0.02B} &
  11.0 &
  \multicolumn{1}{c|}{32.1} &
  9.4 &
  \multicolumn{1}{c|}{20.4} &
  19.1 &
  \multicolumn{1}{c|}{55.6} &
  26.3 &
  25.1 &
  \multicolumn{1}{c|}{38.5} &
  27.1 \\
BGE-large-en-v1.5 \cite{bge_embedding}&
  \multicolumn{1}{c|}{0.3B} &
  19.6 &
  \multicolumn{1}{c|}{44.7} &
  19.9 &
  \multicolumn{1}{c|}{34.4} &
  23.0 &
  \multicolumn{1}{c|}{55.9} &
  40.3 &
  35.2 &
  \multicolumn{1}{c|}{41.3} &
  38.4 \\
BM25 \cite{10.1561/1500000019}&
  \multicolumn{1}{c|}{-} &
  39.7 &
  \multicolumn{1}{c|}{62.1} &
  36.2 &
  \multicolumn{1}{c|}{52.8} &
  26.0 &
  \multicolumn{1}{c|}{54.9} &
  58.7 &
  50.7 &
  \multicolumn{1}{c|}{37.8} &
  53.3 \\
BGE-small-zh-v1.5 \cite{bge_embedding}&
  \multicolumn{1}{c|}{0.02B} &
  39.4 &
  \multicolumn{1}{c|}{69.8} &
  36.2 &
  \multicolumn{1}{c|}{53.8} &
  34.3 &
  \multicolumn{1}{c|}{70.7} &
  61.8 &
  56.9 &
  \multicolumn{1}{c|}{60.1} &
  59.7 \\
Conan-embedding-v1 \cite{li2024conanembeddinggeneraltextembedding}&
  \multicolumn{1}{c|}{0.3B} &
  43.2 &
  \multicolumn{1}{c|}{73.1} &
  44.4 &
  \multicolumn{1}{c|}{63.5} &
  37.3 &
  \multicolumn{1}{c|}{75.0} &
  67.2 &
  63.5 &
  \multicolumn{1}{c|}{53.3} &
  64.3 \\
NV-Embed-v2 \cite{lee2024nv}&
  \multicolumn{1}{c|}{7B} &
  52.8 &
  \multicolumn{1}{c|}{78.9} &
  49.5 &
  \multicolumn{1}{c|}{66.8} &
  38.0 &
  \multicolumn{1}{c|}{72.6} &
  77.0 &
  65.5 &
  \multicolumn{1}{c|}{41.2} &
  68.6 \\
BGE-m3 \cite{bge-m3}&
  \multicolumn{1}{c|}{0.5B} &
  50.2 &
  \multicolumn{1}{c|}{78.9} &
  48.0 &
  \multicolumn{1}{c|}{66.6} &
  40.9 &
  \multicolumn{1}{c|}{72.1} &
  73.8 &
  65.1 &
  \multicolumn{1}{c|}{62.3} &
  69.1 \\
BGE-large-zh-v1.5 \cite{bge_embedding}&
  \multicolumn{1}{c|}{0.3B} &
  49.3 &
  \multicolumn{1}{c|}{79.1} &
  48.7 &
  \multicolumn{1}{c|}{65.8} &
  44.4 &
  \multicolumn{1}{c|}{77.2} &
  72.7 &
  67.8 &
  \multicolumn{1}{c|}{63.1} &
  69.7 \\
Qwen3-Embedding \cite{zhang2025qwen3_embed}&
  \multicolumn{1}{c|}{0.6B} &
  52 &
  \multicolumn{1}{c|}{{81.7}} &
  48.5 &
  \multicolumn{1}{c|}{67.6} &
  40.2 &
  \multicolumn{1}{c|}{74.7} &
  72.8 &
  68.3 &
  \multicolumn{1}{c|}{\cellcolor{green!10}\textbf{71.4}} &
  70.9 \\ 
Gte-Qwen2-Instruct \cite{li2023towards}&
  \multicolumn{1}{c|}{7B} &
  54.4 &
  \multicolumn{1}{c|}{78.9} &
  51.5 &
  \multicolumn{1}{c|}{71.9} &
  46.9 &
  \multicolumn{1}{c|}{77.5} &
  76.7 &
  67.9 &
  \multicolumn{1}{c|}{59.5} &
  71.4 \\
MiniCPM-Embedding \cite{hu2024minicpm}&
  \multicolumn{1}{c|}{2B} &
  59.3 &
  \multicolumn{1}{c|}{{83.9}} &
  55.9 &
  \multicolumn{1}{c|}{73.5} &
  44.9 &
  \multicolumn{1}{c|}{{\cellcolor{green!10}\textbf{81.1}}} &
  78.3 &
  70.9 &
  \multicolumn{1}{c|}{68.4} &
  74.3 \\ 
Qwen3-Embedding \cite{zhang2025qwen3_embed}&
  \multicolumn{1}{c|}{4B} &
  60 &
  \multicolumn{1}{c|}{\cellcolor{green!10}\textbf{87.4}} &
  54.1 &
  \multicolumn{1}{c|}{74.2} &
  47.8 &
  \multicolumn{1}{c|}{80.2} &
  81.6 &
  73.4 &
  \multicolumn{1}{c|}{70.1} &
  77.1 \\ 
Qwen3-Embedding \cite{zhang2025qwen3_embed}&
  \multicolumn{1}{c|}{8B} &
  {\cellcolor{green!10}\textbf{61.0}} &
  \multicolumn{1}{c|}{86.8} &
  {\cellcolor{green!10}\textbf{60.7}} &
  \multicolumn{1}{c|}{\cellcolor{green!10}\textbf{78.8}} &
  \cellcolor{green!10}\textbf{50.1} &
  \multicolumn{1}{c|}{{81.0}} &
  {\cellcolor{green!10}\textbf{82.2}} &
  {\cellcolor{green!10}\textbf{75.8}} &
  \multicolumn{1}{c|}{68.6} &
  {\cellcolor{green!10}\textbf{78.2}} \\ \hline
\multicolumn{12}{c}{\textit{\textbf{Visual-based Method}}} \\ \hline
SigLip \cite{zhai2023siglip} &
  \multicolumn{1}{c|}{0.8B} &
  4.0 &
  \multicolumn{1}{c|}{15.2} &
  4.6 &
  \multicolumn{1}{c|}{10.2} &
  9.6 &
  \multicolumn{1}{c|}{39.2} &
  13.4 &
  12.1 &
  \multicolumn{1}{c|}{15.2} &
  13.1 \\
CLIP \cite{radford2021clip} &
  \multicolumn{1}{c|}{0.4B} &
  4.4 &
  \multicolumn{1}{c|}{16.9} &
  2.8 &
  \multicolumn{1}{c|}{11.2} &
  11.7 &
  \multicolumn{1}{c|}{39.0} &
  15.7 &
  12.6 &
  \multicolumn{1}{c|}{17.7} &
  14.7 \\
ColSmolVLM-v0.1 \cite{faysse2024colpali}&
  \multicolumn{1}{c|}{2B} &
  5.1 &
  \multicolumn{1}{c|}{18.5} &
  3.8 &
  \multicolumn{1}{c|}{11.7} &
  11.4 &
  \multicolumn{1}{c|}{44.1} &
  11.0 &
  13.6 &
  \multicolumn{1}{c|}{45.6} &
  15.7 \\
ColPali-v1.2-merged \cite{faysse2024colpali} &
  \multicolumn{1}{c|}{3B} &
  5.5 &
  \multicolumn{1}{c|}{19.1} &
  4.9 &
  \multicolumn{1}{c|}{13.0} &
  15.0 &
  \multicolumn{1}{c|}{44.0} &
  15.5 &
  13.4 &
  \multicolumn{1}{c|}{37.5} &
  17.0 \\
VisRAG-Ret \cite{yu2024visrag} &
  \multicolumn{1}{c|}{3B} &
  29.1 &
  \multicolumn{1}{c|}{56.0} &
  25.0 &
  \multicolumn{1}{c|}{39.8} &
  28.0 &
  \multicolumn{1}{c|}{63.8} &
  44.4 &
  43.5 &
  \multicolumn{1}{c|}{74.8} &
  47.3 \\
Gme-Qwen2-VL-Instruct \cite{zhang2024gme} &
  \multicolumn{1}{c|}{7B} &
  36.8 &
  \multicolumn{1}{c|}{63.7} &
  32.1 &
  \multicolumn{1}{c|}{50.0} &
  33.1 &
  \multicolumn{1}{c|}{71.1} &
  51.1 &
  54.3 &
  \multicolumn{1}{c|}{76.4} &
  55.0 \\
Qwen3-VL-Embedding \cite{qwen3-vl-embedding} &
  \multicolumn{1}{c|}{2B} &
  44.7 &
  \multicolumn{1}{c|}{72.5} &
  32.4 &
  \multicolumn{1}{c|}{54.6} &
  37.7 &
  \multicolumn{1}{c|}{76.3} &
  55.6 &
  64.6 &
  \multicolumn{1}{c|}{{83.4}} &
  62.1 \\
Qwen3-VL-Embedding \cite{qwen3-vl-embedding} &
  \multicolumn{1}{c|}{8B} &
  53.3 &
  \multicolumn{1}{c|}{80.2} &
  44.1 &
  \multicolumn{1}{c|}{63.0} &
  45.5 &
  \multicolumn{1}{c|}{79.6} &
  66.1 &
  71.2 &
  \multicolumn{1}{c|}{{84.0}} &
  70.0 \\
ColQwen2-v1.0-merged \cite{faysse2024colpali} &
  \multicolumn{1}{c|}{2B} &
  66.1 &
  \multicolumn{1}{c|}{89.2} &
  48.7 &
  \multicolumn{1}{c|}{68.9} &
  43.9 &
  \multicolumn{1}{c|}{72.0} &
  76.9 &
  73.8 &
  \multicolumn{1}{c|}{{{82.8}}} &
  76.3 \\
ColQwen2.5-v0.2 \cite{faysse2024colpali} &
  \multicolumn{1}{c|}{3B} &
  {71.3} &
  \multicolumn{1}{c|}{{90.1}} &
  {56.4} &
  \multicolumn{1}{c|}{{72.5}} &
  {{49.7}} &
  \multicolumn{1}{c|}{{82.1}} &
  {80.4} &
  {75.8} &
  \multicolumn{1}{c|}{81.9} &
  {{78.7}} \\ 
Jina-Embedding-v4 \cite{jina-embedding-v4} &
  \multicolumn{1}{c|}{3B} &
  \cellcolor{green!10}\textbf{72.3} &
  \multicolumn{1}{c|}{\cellcolor{green!10}\textbf{90.7}} &
  55.1 &
  \multicolumn{1}{c|}{74.2} &
  49.5 &
  \multicolumn{1}{c|}{\cellcolor{green!10}\textbf{84.4}} &
  80.3 &
  77.3 &
  \multicolumn{1}{c|}{{{82.8}}} &
  79.4 \\
ColNomic-Embed-Multimodal \cite{colnomic} &
  \multicolumn{1}{c|}{3B} &
  70.2 &
  \multicolumn{1}{c|}{90.5} &
  \cellcolor{green!10}\textbf{60.7} &
  \multicolumn{1}{c|}{\cellcolor{green!10}\textbf{76.5}} &
  \cellcolor{green!10}\textbf{51.5} &
  \multicolumn{1}{c|}{83.0} &
  \cellcolor{green!10}\textbf{80.5} &
  \cellcolor{green!10}\textbf{78.2} &
  \multicolumn{1}{c|}{\cellcolor{green!10}\textbf{84.8}} &
  \cellcolor{green!10}\textbf{80.1} \\ \hline
\end{tabular}
}
\vspace{-3mm}
\end{table*}
}

\subsubsection{Retriever Evaluation}

When processing long documents, feeding the entire content into LLMs or LVLMs is highly token-consuming and incurs substantial computational overhead. Additionally, redundant information within the document interferes with the model, degrading its performance, and some models do not even support inputs of such length at all. Thus, using Retrieval-Augmented Generation (RAG) technology to handle long documents may offer a more efficient solution. In an RAG system, the retriever plays a pivotal role. Current retrievers can be categorized into text-based retrievers, which search the relevant OCR results of document pages using a query, and visual-based retrievers, which directly search the relevant document pages using a query in an OCR-free manner. 

{
Therefore, we conducted extensive experiments on LongHisDoc using both types of retrievers, aiming to provide insights for the development of future RAG systems for historical document understanding. We use the top-\(k\) recall rate to assess the retriever's performance. Specifically, for each question, Recall@\(k\) is calculated based on the coverage of all required evidence pages, rather than whether any single evidence page is retrieved. Given the annotated evidence page set \(G\) and the top-\(k\) retrieved page set \(P_k\), Recall@\(k\) is defined as the proportion of annotated evidence pages that appear in the top-\(k\) retrieved pages, as shown in Eq.~\eqref{eq:recall}.
}
\begin{equation}
\mathrm{Recall@}k = \frac{|P_k \cap G|}{|G|}
\label{eq:recall}
\end{equation}

The results in Table~\ref{tab:table4} indicate the following:  
(1) \textbf{Visual-based retrievers demonstrate great potential, showcasing the possibility of achieving OCR-free historical document page retrieval.}  The best-performing visual-based retriever, ColNomic\cite{colnomic} (Top-5 recall: 80.1\%), surpasses the top text-based retriever, Qwen3-Embedding\cite{zhang2025qwen3_embed} (Top-5 recall: 78.2\%). Meanwhile, for visual modality evidence, advanced visual-based retrievers often perform better. 
(2) \textbf{Model training corpus is crucial.} Models trained on Chinese corpus typically outperform those trained primarily on English data. For example, BGE-zh-small significantly outperforms BGE-large-en, despite the latter having over ten times the number of parameters. This suggests that fine-tuning state-of-the-art retrievers with data specific to historical documents may lead to further performance improvements.

\begin{table}[!h]
\caption{\textcolor{black}{End-to-end RAG evaluation on LongHisDoc using Top-5 retrieved results from different text-based and visual-based retrievers with state-of-the-art generation models. Zoom in for better view.}}
\label{tab:combined_rag_eval}
\resizebox{\textwidth}{!}{
\begin{tabular}{llccccccccccccccc}
\hline
\multicolumn{1}{c}{\large \textbf{\multirow{2}{*}{Retriever}}} &
  \multicolumn{1}{c}{\large \textbf{\multirow{2}{*}{Size}}} &
  \multicolumn{1}{c|}{\multirow{2}{*}{\begin{tabular}[c]{@{}c@{}}\textbf{Gen}\\ \textbf{Model}\end{tabular}}} &
  \multicolumn{4}{c|}{Evidence Page} &
  \multicolumn{6}{c|}{Answer Format} &
  \multicolumn{3}{c|}{Evidence Source} &
  \multirow{2}{*}{AVG} \\ \cline{4-16}
\multicolumn{1}{c}{} &
  \multicolumn{1}{c}{} &
  \multicolumn{1}{c|}{} &
  SIN &
  CRO &
  MUL &
  \multicolumn{1}{c|}{UNA} &
  STR &
  INT &
  OPT &
  LIST &
  BOOL &
  \multicolumn{1}{c|}{NONE} &
  TXT &
  LAY &
  \multicolumn{1}{c|}{VIS} &
    \\ \hline
\multicolumn{17}{c}{\textbf{Text-based RAG}} \\ \hline
\multirow{2}{*}{\large No-Retrieval(baseline)} & \multirow{2}{*}{-} & \multicolumn{1}{c|}{DeepSeek-V3.2 \cite{liu2024deepseek}} & 68.7 & 64.7 & 46.3 & \multicolumn{1}{c|}{40.8} & 62.1 & 56.7 & 77.8 & 65.1 & 80.7 & \multicolumn{1}{c|}{40.8} & 69.5 & 64.7 & \multicolumn{1}{c|}{32.2} & 61.4 \\
 &  & \multicolumn{1}{c|}{Gemini-3.1-Pro\cite{Gemini_3_1_pro}} & \cellcolor{green!10}\textbf{73.7} & \cellcolor{green!10}\textbf{73.0} & \cellcolor{green!10}\textbf{63.0} & \multicolumn{1}{c|}{54.1} & \cellcolor{green!10}\textbf{69.3} & \cellcolor{green!10}\textbf{69.2} & \cellcolor{green!10}\textbf{80.6} & \cellcolor{green!10}\textbf{72.6} & \cellcolor{green!10}\textbf{91.2} & \multicolumn{1}{c|}{54.1} & \cellcolor{green!10}\textbf{76.3} & \cellcolor{green!10}\textbf{73.9} & \multicolumn{1}{c|}{\cellcolor{green!10}\textbf{40.5}} & \cellcolor{green!10}\textbf{69.9} \\ \hline
\multirow{2}{*}{\large SBERT \cite{reimers2019sentence}} & \multirow{2}{*}{0.02B} & \multicolumn{1}{c|}{DeepSeek-V3.2 \cite{liu2024deepseek}} & 32.5 \textcolor{gray!85!black}{(-36.2)} & 26.3 \textcolor{gray!85!black}{(-38.4)} & 16.3 \textcolor{gray!85!black}{(-30.0)} & \multicolumn{1}{c|}{65.3 \textcolor{green!70!black}{(+24.5)}} & 27.6 \textcolor{gray!85!black}{(-34.5)} & 14.4 \textcolor{gray!85!black}{(-42.3)} & 47.2 \textcolor{gray!85!black}{(-30.6)} & 31.5 \textcolor{gray!85!black}{(-33.6)} & 42.1 \textcolor{gray!85!black}{(-38.6)} & \multicolumn{1}{c|}{65.3 \textcolor{green!70!black}{(+24.5)}} & 32.0 \textcolor{gray!85!black}{(-37.5)} & 27.3 \textcolor{gray!85!black}{(-37.4)} & \multicolumn{1}{c|}{13.4 \textcolor{gray!85!black}{(-18.8)}} & 31.7 \textcolor{gray!85!black}{(-29.7)} \\
 &  & \multicolumn{1}{c|}{Gemini-3.1-Pro\cite{Gemini_3_1_pro}} & 30.2 \textcolor{gray!85!black}{(-43.5)} & 23.6 \textcolor{gray!85!black}{(-49.4)} & 20.6 \textcolor{gray!85!black}{(-42.4)} & \multicolumn{1}{c|}{\cellcolor{green!10}\textbf{83.7} \textcolor{green!70!black}{(+29.6)}} & 26.4 \textcolor{gray!85!black}{(-42.9)} & 12.7 \textcolor{gray!85!black}{(-56.5)} & 55.6 \textcolor{gray!85!black}{(-25.0)} & 22.7 \textcolor{gray!85!black}{(-49.9)} & 54.4 \textcolor{gray!85!black}{(-36.8)} & \multicolumn{1}{c|}{\cellcolor{green!10}\textbf{83.7} \textcolor{green!70!black}{(+29.6)}} & 30.5 \textcolor{gray!85!black}{(-45.8)} & 25.5 \textcolor{gray!85!black}{(-48.4)} & \multicolumn{1}{c|}{16.2 \textcolor{gray!85!black}{(-24.3)}} & 32.5 \textcolor{gray!85!black}{(-37.4)} \\ \hline
\multirow{2}{*}{\large BGE-large-en-v1.5 \cite{bge_embedding}} & \multirow{2}{*}{0.3B} & \multicolumn{1}{c|}{DeepSeek-V3.2 \cite{liu2024deepseek}} & 40.6 \textcolor{gray!85!black}{(-28.1)} & 32.5 \textcolor{gray!85!black}{(-32.2)} & 20.3 \textcolor{gray!85!black}{(-26.0)} & \multicolumn{1}{c|}{59.2 \textcolor{green!70!black}{(+18.4)}} & 35.7 \textcolor{gray!85!black}{(-26.4)} & 20.3 \textcolor{gray!85!black}{(-36.4)} & 61.1 \textcolor{gray!85!black}{(-16.7)} & 34.4 \textcolor{gray!85!black}{(-30.7)} & 43.9 \textcolor{gray!85!black}{(-36.8)} & \multicolumn{1}{c|}{59.2 \textcolor{green!70!black}{(+18.4)}} & 39.3 \textcolor{gray!85!black}{(-30.2)} & 34.2 \textcolor{gray!85!black}{(-30.5)} & \multicolumn{1}{c|}{18.2 \textcolor{gray!85!black}{(-14.0)}} & 37.4 \textcolor{gray!85!black}{(-24.0)} \\
 &  & \multicolumn{1}{c|}{Gemini-3.1-Pro\cite{Gemini_3_1_pro}} & 38.2 \textcolor{gray!85!black}{(-35.5)} & 28.2 \textcolor{gray!85!black}{(-44.8)} & 18.8 \textcolor{gray!85!black}{(-44.2)} & \multicolumn{1}{c|}{75.5 \textcolor{green!70!black}{(+21.4)}} & 32.9 \textcolor{gray!85!black}{(-36.4)} & 18.6 \textcolor{gray!85!black}{(-50.6)} & 52.8 \textcolor{gray!85!black}{(-27.8)} & 30.6 \textcolor{gray!85!black}{(-42.0)} & 47.4 \textcolor{gray!85!black}{(-43.8)} & \multicolumn{1}{c|}{75.5 \textcolor{green!70!black}{(+21.4)}} & 35.8 \textcolor{gray!85!black}{(-40.5)} & 31.6 \textcolor{gray!85!black}{(-42.3)} & \multicolumn{1}{c|}{19.2 \textcolor{gray!85!black}{(-21.3)}} & 36.6 \textcolor{gray!85!black}{(-33.3)} \\ \hline
\multirow{2}{*}{\large BM25 \cite{10.1561/1500000019}} & \multirow{2}{*}{-} & \multicolumn{1}{c|}{DeepSeek-V3.2 \cite{liu2024deepseek}} & 59.7 \textcolor{gray!85!black}{(-9.0)} & 43.4 \textcolor{gray!85!black}{(-21.3)} & 25.0 \textcolor{gray!85!black}{(-21.3)} & \multicolumn{1}{c|}{50.0 \textcolor{green!70!black}{(+9.2)}} & 50.9 \textcolor{gray!85!black}{(-11.2)} & 38.1 \textcolor{gray!85!black}{(-18.6)} & 72.2 \textcolor{gray!85!black}{(-5.6)} & 40.1 \textcolor{gray!85!black}{(-25.0)} & 68.4 \textcolor{gray!85!black}{(-12.3)} & \multicolumn{1}{c|}{50.0 \textcolor{green!70!black}{(+9.2)}} & 57.9 \textcolor{gray!85!black}{(-11.6)} & 47.3 \textcolor{gray!85!black}{(-17.4)} & \multicolumn{1}{c|}{19.6 \textcolor{gray!85!black}{(-12.6)}} & 49.7 \textcolor{gray!85!black}{(-11.7)} \\
 &  & \multicolumn{1}{c|}{Gemini-3.1-Pro\cite{Gemini_3_1_pro}} & 54.8 \textcolor{gray!85!black}{(-18.9)} & 44.0 \textcolor{gray!85!black}{(-29.0)} & 23.0 \textcolor{gray!85!black}{(-40.0)} & \multicolumn{1}{c|}{66.3 \textcolor{green!70!black}{(+12.2)}} & 47.8 \textcolor{gray!85!black}{(-21.5)} & 33.1 \textcolor{gray!85!black}{(-36.1)} & 66.7 \textcolor{gray!85!black}{(-13.9)} & 40.1 \textcolor{gray!85!black}{(-32.5)} & 63.2 \textcolor{gray!85!black}{(-28.0)} & \multicolumn{1}{c|}{66.3 \textcolor{green!70!black}{(+12.2)}} & 52.6 \textcolor{gray!85!black}{(-23.7)} & 44.9 \textcolor{gray!85!black}{(-29.0)} & \multicolumn{1}{c|}{24.4 \textcolor{gray!85!black}{(-16.1)}} & 48.4 \textcolor{gray!85!black}{(-21.5)} \\ \hline
\multirow{2}{*}{\large BGE-small-zh-v1.5 \cite{bge_embedding}} & \multirow{2}{*}{0.02B} & \multicolumn{1}{c|}{DeepSeek-V3.2 \cite{liu2024deepseek}} & 59.7 \textcolor{gray!85!black}{(-9.0)} & 41.2 \textcolor{gray!85!black}{(-23.5)} & 29.7 \textcolor{gray!85!black}{(-16.6)} & \multicolumn{1}{c|}{33.7 \textcolor{gray!85!black}{(-7.1)}} & 50.9 \textcolor{gray!85!black}{(-11.2)} & 45.4 \textcolor{gray!85!black}{(-11.3)} & 63.9 \textcolor{gray!85!black}{(-13.9)} & 38.9 \textcolor{gray!85!black}{(-26.2)} & 68.4 \textcolor{gray!85!black}{(-12.3)} & \multicolumn{1}{c|}{33.7 \textcolor{gray!85!black}{(-7.1)}} & 57.9 \textcolor{gray!85!black}{(-11.6)} & 48.1 \textcolor{gray!85!black}{(-16.6)} & \multicolumn{1}{c|}{20.6 \textcolor{gray!85!black}{(-11.6)}} & 48.5 \textcolor{gray!85!black}{(-12.9)} \\
 &  & \multicolumn{1}{c|}{Gemini-3.1-Pro\cite{Gemini_3_1_pro}} & 62.4 \textcolor{gray!85!black}{(-11.3)} & 44.2 \textcolor{gray!85!black}{(-28.8)} & 28.8 \textcolor{gray!85!black}{(-34.2)} & \multicolumn{1}{c|}{73.5 \textcolor{green!70!black}{(+19.4)}} & 54.2 \textcolor{gray!85!black}{(-15.1)} & 40.7 \textcolor{gray!85!black}{(-28.5)} & 72.2 \textcolor{gray!85!black}{(-8.4)} & 39.6 \textcolor{gray!85!black}{(-33.0)} & 71.9 \textcolor{gray!85!black}{(-19.3)} & \multicolumn{1}{c|}{73.5 \textcolor{green!70!black}{(+19.4)}} & 59.8 \textcolor{gray!85!black}{(-16.5)} & 50.9 \textcolor{gray!85!black}{(-23.0)} & \multicolumn{1}{c|}{21.1 \textcolor{gray!85!black}{(-19.4)}} & 54.2 \textcolor{gray!85!black}{(-15.7)} \\ \hline
\multirow{2}{*}{\large Conan-embedding-v1 \cite{li2024conanembeddinggeneraltextembedding}} & \multirow{2}{*}{0.3B} & \multicolumn{1}{c|}{DeepSeek-V3.2 \cite{liu2024deepseek}} & 56.9 \textcolor{gray!85!black}{(-11.8)} & 43.7 \textcolor{gray!85!black}{(-21.0)} & 24.4 \textcolor{gray!85!black}{(-21.9)} & \multicolumn{1}{c|}{51.0 \textcolor{green!70!black}{(+10.2)}} & 48.0 \textcolor{gray!85!black}{(-14.1)} & 39.8 \textcolor{gray!85!black}{(-16.9)} & 63.9 \textcolor{gray!85!black}{(-13.9)} & 44.7 \textcolor{gray!85!black}{(-20.4)} & 61.4 \textcolor{gray!85!black}{(-19.3)} & \multicolumn{1}{c|}{51.0 \textcolor{green!70!black}{(+10.2)}} & 53.4 \textcolor{gray!85!black}{(-16.1)} & 47.8 \textcolor{gray!85!black}{(-16.9)} & \multicolumn{1}{c|}{23.1 \textcolor{gray!85!black}{(-9.1)}} & 48.3 \textcolor{gray!85!black}{(-13.1)} \\
 &  & \multicolumn{1}{c|}{Gemini-3.1-Pro\cite{Gemini_3_1_pro}} & 59.5 \textcolor{gray!85!black}{(-14.2)} & 50.0 \textcolor{gray!85!black}{(-23.0)} & 27.9 \textcolor{gray!85!black}{(-35.1)} & \multicolumn{1}{c|}{64.3 \textcolor{green!70!black}{(+10.2)}} & 51.2 \textcolor{gray!85!black}{(-18.1)} & 40.7 \textcolor{gray!85!black}{(-28.5)} & 72.2 \textcolor{gray!85!black}{(-8.4)} & 48.8 \textcolor{gray!85!black}{(-23.8)} & 70.2 \textcolor{gray!85!black}{(-21.0)} & \multicolumn{1}{c|}{64.3 \textcolor{green!70!black}{(+10.2)}} & 55.8 \textcolor{gray!85!black}{(-20.5)} & 54.2 \textcolor{gray!85!black}{(-19.7)} & \multicolumn{1}{c|}{21.3 \textcolor{gray!85!black}{(-19.2)}} & 52.7 \textcolor{gray!85!black}{(-17.2)} \\ \hline
\multirow{2}{*}{\large NV-Embed-v2 \cite{lee2024nv}} & \multirow{2}{*}{7B} & \multicolumn{1}{c|}{DeepSeek-V3.2 \cite{liu2024deepseek}}
& 58.8 \textcolor{gray!85!black}{(-9.9)}
& 49.1 \textcolor{gray!85!black}{(-15.6)}
& 30.1 \textcolor{gray!85!black}{(-16.2)}
& \multicolumn{1}{c|}{51.0 \textcolor{green!70!black}{(+10.2)}}
& 51.9 \textcolor{gray!85!black}{(-10.2)}
& 40.7 \textcolor{gray!85!black}{(-16.0)}
& 63.9 \textcolor{gray!85!black}{(-13.9)}
& 47.6 \textcolor{gray!85!black}{(-17.5)}
& 68.4 \textcolor{gray!85!black}{(-12.3)}
& \multicolumn{1}{c|}{51.0 \textcolor{green!70!black}{(+10.2)}}
& 59.5 \textcolor{gray!85!black}{(-10.0)}
& 49.3 \textcolor{gray!85!black}{(-15.4)}
& \multicolumn{1}{c|}{20.6 \textcolor{gray!85!black}{(-11.6)}}
& 51.3 \textcolor{gray!85!black}{(-10.1)} \\
 &  & \multicolumn{1}{c|}{Gemini-3.1-Pro\cite{Gemini_3_1_pro}} & 62.9 \textcolor{gray!85!black}{(-10.8)} & 53.5 \textcolor{gray!85!black}{(-19.5)} & 28.8 \textcolor{gray!85!black}{(-34.2)} & \multicolumn{1}{c|}{65.3 \textcolor{green!70!black}{(+11.2)}} & 54.0 \textcolor{gray!85!black}{(-15.3)} & 48.7 \textcolor{gray!85!black}{(-20.5)} & 66.7 \textcolor{gray!85!black}{(-13.9)} & 52.0 \textcolor{gray!85!black}{(-20.6)} & 68.4 \textcolor{gray!85!black}{(-22.8)} & \multicolumn{1}{c|}{65.3 \textcolor{green!70!black}{(+11.2)}} & 62.6 \textcolor{gray!85!black}{(-13.7)} & 51.9 \textcolor{gray!85!black}{(-22.0)} & \multicolumn{1}{c|}{25.6 \textcolor{gray!85!black}{(-14.9)}} & 55.5 \textcolor{gray!85!black}{(-14.4)} \\ \hline
\multirow{2}{*}{\large BGE-m3 \cite{bge-m3}} & \multirow{2}{*}{0.5B} & \multicolumn{1}{c|}{DeepSeek-V3.2 \cite{liu2024deepseek}}
& 59.6 \textcolor{gray!85!black}{(-9.1)}
& 48.0 \textcolor{gray!85!black}{(-16.7)}
& 26.2 \textcolor{gray!85!black}{(-20.1)}
& \multicolumn{1}{c|}{50.0 \textcolor{green!70!black}{(+9.2)}}
& 50.7 \textcolor{gray!85!black}{(-11.4)}
& 42.4 \textcolor{gray!85!black}{(-14.3)}
& 63.9 \textcolor{gray!85!black}{(-13.9)}
& 47.6 \textcolor{gray!85!black}{(-17.5)}
& 68.4 \textcolor{gray!85!black}{(-12.3)}
& \multicolumn{1}{c|}{50.0 \textcolor{green!70!black}{(+9.2)}}
& 58.2 \textcolor{gray!85!black}{(-11.3)}
& 48.1 \textcolor{gray!85!black}{(-16.6)}
& \multicolumn{1}{c|}{26.5 \textcolor{gray!85!black}{(-5.7)}}
& 50.7 \textcolor{gray!85!black}{(-10.7)} \\
 &  & \multicolumn{1}{c|}{Gemini-3.1-Pro\cite{Gemini_3_1_pro}} & 60.8 \textcolor{gray!85!black}{(-12.9)} & 48.8 \textcolor{gray!85!black}{(-24.2)} & 31.7 \textcolor{gray!85!black}{(-31.3)} & \multicolumn{1}{c|}{63.3 \textcolor{green!70!black}{(+9.2)}} & 51.9 \textcolor{gray!85!black}{(-17.4)} & 46.6 \textcolor{gray!85!black}{(-22.6)} & 77.8 \textcolor{gray!85!black}{(-2.8)} & 47.6 \textcolor{gray!85!black}{(-25.0)} & 70.2 \textcolor{gray!85!black}{(-21.0)} & \multicolumn{1}{c|}{63.3 \textcolor{green!70!black}{(+9.2)}} & 58.2 \textcolor{gray!85!black}{(-18.1)} & 52.4 \textcolor{gray!85!black}{(-21.5)} & \multicolumn{1}{c|}{28.2 \textcolor{gray!85!black}{(-12.3)}} & 53.8 \textcolor{gray!85!black}{(-16.1)} \\ \hline
\multirow{2}{*}{\large BGE-large-zh-v1.5 \cite{bge_embedding}} & \multirow{2}{*}{0.3B} & \multicolumn{1}{c|}{DeepSeek-V3.2 \cite{liu2024deepseek}} & 60.5 \textcolor{gray!85!black}{(-8.2)} & 45.4 \textcolor{gray!85!black}{(-19.3)} & 26.8 \textcolor{gray!85!black}{(-19.5)} & \multicolumn{1}{c|}{50.0 \textcolor{green!70!black}{(+9.2)}} & 51.0 \textcolor{gray!85!black}{(-11.1)} & 44.9 \textcolor{gray!85!black}{(-11.8)} & 63.9 \textcolor{gray!85!black}{(-13.9)} & 46.8 \textcolor{gray!85!black}{(-18.3)} & 63.2 \textcolor{gray!85!black}{(-17.5)} & \multicolumn{1}{c|}{50.0 \textcolor{green!70!black}{(+9.2)}} & 57.6 \textcolor{gray!85!black}{(-11.9)} & 49.8 \textcolor{gray!85!black}{(-14.9)} & \multicolumn{1}{c|}{23.8 \textcolor{gray!85!black}{(-8.4)}} & 50.8 \textcolor{gray!85!black}{(-10.6)} \\
 &  & \multicolumn{1}{c|}{Gemini-3.1-Pro\cite{Gemini_3_1_pro}} & 61.6 \textcolor{gray!85!black}{(-12.1)} & 47.1 \textcolor{gray!85!black}{(-25.9)} & 32.0 \textcolor{gray!85!black}{(-31.0)} & \multicolumn{1}{c|}{68.4 \textcolor{green!70!black}{(+14.3)}} & 51.5 \textcolor{gray!85!black}{(-17.8)} & 48.3 \textcolor{gray!85!black}{(-20.9)} & 63.9 \textcolor{gray!85!black}{(-16.7)} & 52.6 \textcolor{gray!85!black}{(-20.0)} & 70.2 \textcolor{gray!85!black}{(-21.0)} & \multicolumn{1}{c|}{68.4 \textcolor{green!70!black}{(+14.3)}} & 58.5 \textcolor{gray!85!black}{(-17.8)} & 52.9 \textcolor{gray!85!black}{(-21.0)} & \multicolumn{1}{c|}{27.1 \textcolor{gray!85!black}{(-13.4)}} & 54.4 \textcolor{gray!85!black}{(-15.5)} \\ \hline
\multirow{2}{*}{\large Qwen3-Embedding \cite{zhang2025qwen3_embed}} & \multirow{2}{*}{0.6B} & \multicolumn{1}{c|}{DeepSeek-V3.2 \cite{liu2024deepseek}} & 60.2 \textcolor{gray!85!black}{(-8.5)} & 46.9 \textcolor{gray!85!black}{(-17.8)} & 30.4 \textcolor{gray!85!black}{(-15.9)} & \multicolumn{1}{c|}{44.9 \textcolor{green!70!black}{(+4.1)}} & 51.5 \textcolor{gray!85!black}{(-10.6)} & 42.0 \textcolor{gray!85!black}{(-14.7)} & 63.9 \textcolor{gray!85!black}{(-13.9)} & 51.7 \textcolor{gray!85!black}{(-13.4)} & 66.7 \textcolor{gray!85!black}{(-14.0)} & \multicolumn{1}{c|}{44.9 \textcolor{green!70!black}{(+4.1)}} & 58.1 \textcolor{gray!85!black}{(-11.4)} & 50.9 \textcolor{gray!85!black}{(-13.8)} & \multicolumn{1}{c|}{25.0 \textcolor{gray!85!black}{(-7.2)}} & 51.1 \textcolor{gray!85!black}{(-10.3)} \\
 &  & \multicolumn{1}{c|}{Gemini-3.1-Pro\cite{Gemini_3_1_pro}} & 60.3 \textcolor{gray!85!black}{(-13.4)} & 53.0 \textcolor{gray!85!black}{(-20.0)} & 27.8 \textcolor{gray!85!black}{(-35.2)} & \multicolumn{1}{c|}{62.2 \textcolor{green!70!black}{(+8.1)}} & 51.8 \textcolor{gray!85!black}{(-17.5)} & 47.1 \textcolor{gray!85!black}{(-22.1)} & 63.9 \textcolor{gray!85!black}{(-16.7)} & 50.4 \textcolor{gray!85!black}{(-22.2)} & 70.2 \textcolor{gray!85!black}{(-21.0)} & \multicolumn{1}{c|}{62.2 \textcolor{green!70!black}{(+8.1)}} & 58.6 \textcolor{gray!85!black}{(-17.7)} & 53.1 \textcolor{gray!85!black}{(-20.8)} & \multicolumn{1}{c|}{22.2 \textcolor{gray!85!black}{(-18.3)}} & 53.5 \textcolor{gray!85!black}{(-16.4)} \\ \hline
\multirow{2}{*}{\large Gte-Qwen2-Instruct \cite{li2023towards}} & \multirow{2}{*}{7B} & \multicolumn{1}{c|}{DeepSeek-V3.2 \cite{liu2024deepseek}} & 62.3 \textcolor{gray!85!black}{(-6.4)} & 47.2 \textcolor{gray!85!black}{(-17.5)} & 34.1 \textcolor{gray!85!black}{(-12.2)} & \multicolumn{1}{c|}{30.6 \textcolor{gray!85!black}{(-10.2)}} & 53.6 \textcolor{gray!85!black}{(-8.5)} & 47.1 \textcolor{gray!85!black}{(-9.6)} & 69.4 \textcolor{gray!85!black}{(-8.4)} & 46.3 \textcolor{gray!85!black}{(-18.8)} & 77.2 \textcolor{gray!85!black}{(-3.5)} & \multicolumn{1}{c|}{30.6 \textcolor{gray!85!black}{(-10.2)}} & 60.6 \textcolor{gray!85!black}{(-8.9)} & 52.1 \textcolor{gray!85!black}{(-12.6)} & \multicolumn{1}{c|}{28.0 \textcolor{gray!85!black}{(-4.2)}} & 51.5 \textcolor{gray!85!black}{(-9.9)} \\
 &  & \multicolumn{1}{c|}{Gemini-3.1-Pro\cite{Gemini_3_1_pro}} & 62.7 \textcolor{gray!85!black}{(-11.0)} & 55.5 \textcolor{gray!85!black}{(-17.5)} & 36.1 \textcolor{gray!85!black}{(-26.9)} & \multicolumn{1}{c|}{64.3 \textcolor{green!70!black}{(+10.2)}} & 56.7 \textcolor{gray!85!black}{(-12.6)} & 44.9 \textcolor{gray!85!black}{(-24.3)} & 69.4 \textcolor{gray!85!black}{(-11.2)} & 51.1 \textcolor{gray!85!black}{(-21.5)} & 77.2 \textcolor{gray!85!black}{(-14.0)} & \multicolumn{1}{c|}{64.3 \textcolor{green!70!black}{(+10.2)}} & 64.8 \textcolor{gray!85!black}{(-11.5)} & 54.1 \textcolor{gray!85!black}{(-19.8)} & \multicolumn{1}{c|}{23.4 \textcolor{gray!85!black}{(-17.1)}} & 57.0 \textcolor{gray!85!black}{(-12.9)} \\ \hline
\multirow{2}{*}{\large MiniCPM-Embedding \cite{hu2024minicpm}} & \multirow{2}{*}{2B} & \multicolumn{1}{c|}{DeepSeek-V3.2 \cite{liu2024deepseek}} & 61.3 \textcolor{gray!85!black}{(-7.4)} & 47.7 \textcolor{gray!85!black}{(-17.0)} & 28.3 \textcolor{gray!85!black}{(-18.0)} & \multicolumn{1}{c|}{46.9 \textcolor{green!70!black}{(+6.1)}} & 52.0 \textcolor{gray!85!black}{(-10.1)} & 45.8 \textcolor{gray!85!black}{(-10.9)} & 69.4 \textcolor{gray!85!black}{(-8.4)} & 49.7 \textcolor{gray!85!black}{(-15.4)} & 61.4 \textcolor{gray!85!black}{(-19.3)} & \multicolumn{1}{c|}{46.9 \textcolor{green!70!black}{(+6.1)}} & 58.6 \textcolor{gray!85!black}{(-10.9)} & 51.5 \textcolor{gray!85!black}{(-13.2)} & \multicolumn{1}{c|}{24.8 \textcolor{gray!85!black}{(-7.4)}} & 51.7 \textcolor{gray!85!black}{(-9.7)} \\
 &  & \multicolumn{1}{c|}{Gemini-3.1-Pro\cite{Gemini_3_1_pro}} & 65.0 \textcolor{gray!85!black}{(-8.7)} & 57.2 \textcolor{gray!85!black}{(-15.8)} & 34.2 \textcolor{gray!85!black}{(-28.8)} & \multicolumn{1}{c|}{62.2 \textcolor{green!70!black}{(+8.1)}} & 58.0 \textcolor{gray!85!black}{(-11.3)} & 47.5 \textcolor{gray!85!black}{(-21.7)} & 69.4 \textcolor{gray!85!black}{(-11.2)} & 54.1 \textcolor{gray!85!black}{(-18.5)} & 73.7 \textcolor{gray!85!black}{(-17.5)} & \multicolumn{1}{c|}{62.2 \textcolor{green!70!black}{(+8.1)}} & 64.0 \textcolor{gray!85!black}{(-12.3)} & 57.5 \textcolor{gray!85!black}{(-16.4)} & \multicolumn{1}{c|}{27.1 \textcolor{gray!85!black}{(-13.4)}} & 58.0 \textcolor{gray!85!black}{(-11.9)} \\ \hline
\multirow{2}{*}{\large Qwen3-Embedding \cite{zhang2025qwen3_embed}} & \multirow{2}{*}{4B} & \multicolumn{1}{c|}{DeepSeek-V3.2 \cite{liu2024deepseek}} & 64.1 \textcolor{gray!85!black}{(-4.6)} & 50.2 \textcolor{gray!85!black}{(-14.5)} & 34.9 \textcolor{gray!85!black}{(-11.4)} & \multicolumn{1}{c|}{45.9 \textcolor{green!70!black}{(+5.1)}} & 55.8 \textcolor{gray!85!black}{(-6.3)} & 49.6 \textcolor{gray!85!black}{(-7.1)} & 66.7 \textcolor{gray!85!black}{(-11.1)} & 50.3 \textcolor{gray!85!black}{(-14.8)} & 71.9 \textcolor{gray!85!black}{(-8.8)} & \multicolumn{1}{c|}{45.9 \textcolor{green!70!black}{(+5.1)}} & 63.7 \textcolor{gray!85!black}{(-5.8)} & 53.0 \textcolor{gray!85!black}{(-11.7)} & \multicolumn{1}{c|}{27.4 \textcolor{gray!85!black}{(-4.8)}} & 54.7 \textcolor{gray!85!black}{(-6.7)} \\
 &  & \multicolumn{1}{c|}{Gemini-3.1-Pro\cite{Gemini_3_1_pro}} & 65.7 \textcolor{gray!85!black}{(-8.0)} & 57.2 \textcolor{gray!85!black}{(-15.8)} & 35.9 \textcolor{gray!85!black}{(-27.1)} & \multicolumn{1}{c|}{64.3 \textcolor{green!70!black}{(+10.2)}} & 57.8 \textcolor{gray!85!black}{(-11.5)} & 51.3 \textcolor{gray!85!black}{(-17.9)} & 69.4 \textcolor{gray!85!black}{(-11.2)} & 56.9 \textcolor{gray!85!black}{(-15.7)} & 73.7 \textcolor{gray!85!black}{(-17.5)} & \multicolumn{1}{c|}{64.3 \textcolor{green!70!black}{(+10.2)}} & 64.7 \textcolor{gray!85!black}{(-11.6)} & 58.8 \textcolor{gray!85!black}{(-15.1)} & \multicolumn{1}{c|}{26.1 \textcolor{gray!85!black}{(-14.4)}} & 58.8 \textcolor{gray!85!black}{(-11.1)} \\ \hline
\multirow{2}{*}{\large Qwen3-Embedding \cite{zhang2025qwen3_embed}} & \multirow{2}{*}{8B} & \multicolumn{1}{c|}{DeepSeek-V3.2 \cite{liu2024deepseek}} & 64.0 \textcolor{gray!85!black}{(-4.7)} & 52.8 \textcolor{gray!85!black}{(-11.9)} & 36.2 \textcolor{gray!85!black}{(-10.1)} & \multicolumn{1}{c|}{45.9 \textcolor{green!70!black}{(+5.1)}} & 55.6 \textcolor{gray!85!black}{(-6.5)} & 52.1 \textcolor{gray!85!black}{(-4.6)} & 63.9 \textcolor{gray!85!black}{(-13.9)} & 55.5 \textcolor{gray!85!black}{(-9.6)} & 70.2 \textcolor{gray!85!black}{(-10.5)} & \multicolumn{1}{c|}{45.9 \textcolor{green!70!black}{(+5.1)}} & 63.4 \textcolor{gray!85!black}{(-6.1)} & 55.5 \textcolor{gray!85!black}{(-9.2)} & \multicolumn{1}{c|}{26.6 \textcolor{gray!85!black}{(-5.6)}} & 55.4 \textcolor{gray!85!black}{(-6.0)} \\
 &  & \multicolumn{1}{c|}{Gemini-3.1-Pro\cite{Gemini_3_1_pro}} & 66.5 \textcolor{gray!85!black}{(-7.2)} & 56.5 \textcolor{gray!85!black}{(-16.5)} & 36.2 \textcolor{gray!85!black}{(-26.8)} & \multicolumn{1}{c|}{63.3 \textcolor{green!70!black}{(+9.2)}} & 59.0 \textcolor{gray!85!black}{(-10.3)} & 48.7 \textcolor{gray!85!black}{(-20.5)} & 66.7 \textcolor{gray!85!black}{(-13.9)} & 57.6 \textcolor{gray!85!black}{(-15.0)} & 73.7 \textcolor{gray!85!black}{(-17.5)} & \multicolumn{1}{c|}{63.3 \textcolor{green!70!black}{(+9.2)}} & 66.7 \textcolor{gray!85!black}{(-9.6)} & 56.6 \textcolor{gray!85!black}{(-17.3)} & \multicolumn{1}{c|}{28.9 \textcolor{gray!85!black}{(-11.6)}} & 59.1 \textcolor{gray!85!black}{(-10.8)} \\ \hline
\multicolumn{17}{c}{\textbf{Visual-based RAG}} \\ \hline
\multirow{2}{*}{\large No-Retrieval(baseline)} & \multirow{2}{*}{-} & \multicolumn{1}{c|}{ Qwen3-VL-32B \cite{qwen3-vl}} & 39.2 & 25.6 & 23.2 & \multicolumn{1}{c|}{25.5} & 29.3 & 28.3 & 52.8 & 34.9 & 54.4 & \multicolumn{1}{c|}{25.5} & 35.7 & 29.4 & \multicolumn{1}{c|}{32.1} & 32.8 \\
 &  & \multicolumn{1}{c|}{Gemini-3.1-Pro\cite{Gemini_3_1_pro}} & 51.4 & 38.4 & 35.5 & \multicolumn{1}{c|}{31.6} & 44.1 & 39.2 & 55.6 & 45.6 & 68.4 & \multicolumn{1}{c|}{31.6} & 48.5 & 44.0 & \multicolumn{1}{c|}{38.6} & 44.3 \\ \hline
\multirow{2}{*}{\large SigLip \cite{zhai2023siglip}} & \multirow{2}{*}{0.8B} & \multicolumn{1}{c|}{ Qwen3-VL-32B \cite{qwen3-vl}} & 17.2 \textcolor{gray!85!black}{(-22.0)} & 9.8 \textcolor{gray!85!black}{(-15.8)} & 6.9 \textcolor{gray!85!black}{(-16.3)} & \multicolumn{1}{c|}{15.3 \textcolor{gray!85!black}{(-10.2)}} & 11.8 \textcolor{gray!85!black}{(-17.5)} & 16.7 \textcolor{gray!85!black}{(-11.6)} & 36.1 \textcolor{gray!85!black}{(-16.7)} & 6.5 \textcolor{gray!85!black}{(-28.4)} & 28.1 \textcolor{gray!85!black}{(-26.3)} & \multicolumn{1}{c|}{15.3 \textcolor{gray!85!black}{(-10.2)}} & 18.5 \textcolor{gray!85!black}{(-17.2)} & 8.4 \textcolor{gray!85!black}{(-21.0)} & \multicolumn{1}{c|}{10.8 \textcolor{gray!85!black}{(-21.3)}} & 13.8 \textcolor{gray!85!black}{(-19.0)} \\
 &  & \multicolumn{1}{c|}{Gemini-3.1-Pro\cite{Gemini_3_1_pro}} & 26.2 \textcolor{gray!85!black}{(-25.2)} & 24.2 \textcolor{gray!85!black}{(-14.2)} & 19.9 \textcolor{gray!85!black}{(-15.6)} & \multicolumn{1}{c|}{56.1 \textcolor{green!70!black}{(+24.5)}} & 22.4 \textcolor{gray!85!black}{(-21.7)} & 17.0 \textcolor{gray!85!black}{(-22.2)} & 55.6 \textcolor{green!70!black}{(+0.0)} & 23.1 \textcolor{gray!85!black}{(-22.5)} & 45.6 \textcolor{gray!85!black}{(-22.8)} & \multicolumn{1}{c|}{56.1 \textcolor{green!70!black}{(+24.5)}} & 27.4 \textcolor{gray!85!black}{(-21.1)} & 23.5 \textcolor{gray!85!black}{(-20.5)} & \multicolumn{1}{c|}{15.1 \textcolor{gray!85!black}{(-23.5)}} & 27.6 \textcolor{gray!85!black}{(-16.7)} \\ \hline
\multirow{2}{*}{\large CLIP \cite{radford2021clip}} & \multirow{2}{*}{0.4B} & \multicolumn{1}{c|}{ Qwen3-VL-32B \cite{qwen3-vl}} & 22.2 \textcolor{gray!85!black}{(-17.0)} & 18.6 \textcolor{gray!85!black}{(-7.0)} & 15.1 \textcolor{gray!85!black}{(-8.1)} & \multicolumn{1}{c|}{51.0 \textcolor{green!70!black}{(+25.5)}} & 16.7 \textcolor{gray!85!black}{(-12.6)} & 12.7 \textcolor{gray!85!black}{(-15.6)} & 61.1 \textcolor{green!70!black}{(+8.3)} & 20.0 \textcolor{gray!85!black}{(-14.9)} & 43.9 \textcolor{gray!85!black}{(-10.5)} & \multicolumn{1}{c|}{51.0 \textcolor{green!70!black}{(+25.5)}} & 22.7 \textcolor{gray!85!black}{(-13.0)} & 15.6 \textcolor{gray!85!black}{(-13.8)} & \multicolumn{1}{c|}{24.7 \textcolor{gray!85!black}{(-7.4)}} & 23.1 \textcolor{gray!85!black}{(-9.7)} \\
 &  & \multicolumn{1}{c|}{Gemini-3.1-Pro\cite{Gemini_3_1_pro}} & 30.0 \textcolor{gray!85!black}{(-21.4)} & 23.8 \textcolor{gray!85!black}{(-14.6)} & 20.7 \textcolor{gray!85!black}{(-14.8)} & \multicolumn{1}{c|}{57.1 \textcolor{green!70!black}{(+25.5)}} & 25.6 \textcolor{gray!85!black}{(-18.5)} & 15.3 \textcolor{gray!85!black}{(-23.9)} & 55.6 \textcolor{green!70!black}{(+0.0)} & 25.4 \textcolor{gray!85!black}{(-20.2)} & 49.1 \textcolor{gray!85!black}{(-19.3)} & \multicolumn{1}{c|}{57.1 \textcolor{green!70!black}{(+25.5)}} & 30.6 \textcolor{gray!85!black}{(-17.9)} & 24.0 \textcolor{gray!85!black}{(-20.0)} & \multicolumn{1}{c|}{20.6 \textcolor{gray!85!black}{(-18.0)}} & 29.8 \textcolor{gray!85!black}{(-14.5)} \\ \hline
\multirow{2}{*}{\large ColSmolVLM-v0.1 \cite{faysse2024colpali}} & \multirow{2}{*}{2B} & \multicolumn{1}{c|}{ Qwen3-VL-32B \cite{qwen3-vl}} & 22.7 \textcolor{gray!85!black}{(-16.5)} & 18.6 \textcolor{gray!85!black}{(-7.0)} & 15.3 \textcolor{gray!85!black}{(-7.9)} & \multicolumn{1}{c|}{46.9 \textcolor{green!70!black}{(+21.4)}} & 17.4 \textcolor{gray!85!black}{(-11.9)} & 17.8 \textcolor{gray!85!black}{(-10.5)} & 61.1 \textcolor{green!70!black}{(+8.3)} & 15.4 \textcolor{gray!85!black}{(-19.5)} & 42.1 \textcolor{gray!85!black}{(-12.3)} & \multicolumn{1}{c|}{46.9 \textcolor{green!70!black}{(+21.4)}} & 19.9 \textcolor{gray!85!black}{(-15.8)} & 16.7 \textcolor{gray!85!black}{(-12.7)} & \multicolumn{1}{c|}{37.2 \textcolor{green!70!black}{(+5.1)}} & 23.0 \textcolor{gray!85!black}{(-9.8)} \\
 &  & \multicolumn{1}{c|}{Gemini-3.1-Pro\cite{Gemini_3_1_pro}} & 32.1 \textcolor{gray!85!black}{(-19.3)} & 27.4 \textcolor{gray!85!black}{(-11.0)} & 19.7 \textcolor{gray!85!black}{(-15.8)} & \multicolumn{1}{c|}{52.0 \textcolor{green!70!black}{(+20.4)}} & 26.5 \textcolor{gray!85!black}{(-17.6)} & 24.6 \textcolor{gray!85!black}{(-14.6)} & 61.1 \textcolor{green!70!black}{(+5.5)} & 27.0 \textcolor{gray!85!black}{(-18.6)} & 43.9 \textcolor{gray!85!black}{(-24.5)} & \multicolumn{1}{c|}{52.0 \textcolor{green!70!black}{(+20.4)}} & 30.4 \textcolor{gray!85!black}{(-18.1)} & 24.2 \textcolor{gray!85!black}{(-19.8)} & \multicolumn{1}{c|}{38.2 \textcolor{gray!85!black}{(-0.4)}} & 31.0 \textcolor{gray!85!black}{(-13.3)} \\ \hline
\multirow{2}{*}{\large ColPali-v1.2-merged \cite{faysse2024colpali}} & \multirow{2}{*}{3B} & \multicolumn{1}{c|}{ Qwen3-VL-32B \cite{qwen3-vl}} & 24.4 \textcolor{gray!85!black}{(-14.8)} & 15.0 \textcolor{gray!85!black}{(-10.6)} & 18.2 \textcolor{gray!85!black}{(-5.0)} & \multicolumn{1}{c|}{54.1 \textcolor{green!70!black}{(+28.6)}} & 16.5 \textcolor{gray!85!black}{(-12.8)} & 22.9 \textcolor{gray!85!black}{(-5.4)} & 61.1 \textcolor{green!70!black}{(+8.3)} & 17.2 \textcolor{gray!85!black}{(-17.7)} & 49.1 \textcolor{gray!85!black}{(-5.3)} & \multicolumn{1}{c|}{54.1 \textcolor{green!70!black}{(+28.6)}} & 21.9 \textcolor{gray!85!black}{(-13.8)} & 16.9 \textcolor{gray!85!black}{(-12.5)} & \multicolumn{1}{c|}{34.2 \textcolor{green!70!black}{(+2.1)}} & 24.4 \textcolor{gray!85!black}{(-8.4)} \\
 &  & \multicolumn{1}{c|}{Gemini-3.1-Pro\cite{Gemini_3_1_pro}} & 33.0 \textcolor{gray!85!black}{(-18.4)} & 26.0 \textcolor{gray!85!black}{(-12.4)} & 21.4 \textcolor{gray!85!black}{(-14.1)} & \multicolumn{1}{c|}{56.1 \textcolor{green!70!black}{(+24.5)}} & 26.0 \textcolor{gray!85!black}{(-18.1)} & 24.6 \textcolor{gray!85!black}{(-14.6)} & 61.1 \textcolor{green!70!black}{(+5.5)} & 28.2 \textcolor{gray!85!black}{(-17.4)} & 54.4 \textcolor{gray!85!black}{(-14.0)} & \multicolumn{1}{c|}{56.1 \textcolor{green!70!black}{(+24.5)}} & 32.9 \textcolor{gray!85!black}{(-15.6)} & 22.5 \textcolor{gray!85!black}{(-21.5)} & \multicolumn{1}{c|}{37.4 \textcolor{gray!85!black}{(-1.2)}} & 31.9 \textcolor{gray!85!black}{(-12.4)} \\ \hline
\multirow{2}{*}{\large VisRAG-Ret \cite{yu2024visrag}} & \multirow{2}{*}{3B} & \multicolumn{1}{c|}{ Qwen3-VL-32B \cite{qwen3-vl}} & 21.6 \textcolor{gray!85!black}{(-17.6)} & 16.4 \textcolor{gray!85!black}{(-9.2)} & 12.5 \textcolor{gray!85!black}{(-10.7)} & \multicolumn{1}{c|}{55.1 \textcolor{green!70!black}{(+29.6)}} & 15.6 \textcolor{gray!85!black}{(-13.7)} & 17.0 \textcolor{gray!85!black}{(-11.3)} & 58.3 \textcolor{green!70!black}{(+5.5)} & 14.3 \textcolor{gray!85!black}{(-20.6)} & 40.4 \textcolor{gray!85!black}{(-14.0)} & \multicolumn{1}{c|}{55.1 \textcolor{green!70!black}{(+29.6)}} & 21.0 \textcolor{gray!85!black}{(-14.7)} & 14.7 \textcolor{gray!85!black}{(-14.7)} & \multicolumn{1}{c|}{23.6 \textcolor{gray!85!black}{(-8.5)}} & 22.3 \textcolor{gray!85!black}{(-10.5)} \\
 &  & \multicolumn{1}{c|}{Gemini-3.1-Pro\cite{Gemini_3_1_pro}} & 57.6 \textcolor{green!70!black}{(+6.2)} & 45.2 \textcolor{green!70!black}{(+6.8)} & 26.2 \textcolor{gray!85!black}{(-9.3)} & \multicolumn{1}{c|}{\cellcolor{green!10}\textbf{64.3} \textcolor{green!70!black}{(+32.7)}} & 47.4 \textcolor{green!70!black}{(+3.3)} & 46.7 \textcolor{green!70!black}{(+7.5)} & 66.7 \textcolor{green!70!black}{(+11.1)} & 43.5 \textcolor{gray!85!black}{(-2.1)} & 71.9 \textcolor{green!70!black}{(+3.5)} & \multicolumn{1}{c|}{\cellcolor{green!10}\textbf{64.3} \textcolor{green!70!black}{(+32.7)}} & 52.7 \textcolor{green!70!black}{(+4.2)} & 46.1 \textcolor{green!70!black}{(+2.1)} & \multicolumn{1}{c|}{43.2 \textcolor{green!70!black}{(+4.6)}} & 50.5 \textcolor{green!70!black}{(+6.2)} \\ \hline
\multirow{2}{*}{\large Gme-Qwen2-VL-Instruct \cite{zhang2024gme}} & \multirow{2}{*}{7B} & \multicolumn{1}{c|}{ Qwen3-VL-32B \cite{qwen3-vl}} & 44.2 \textcolor{green!70!black}{(+5.0)} & 28.1 \textcolor{green!70!black}{(+2.5)} & 27.2 \textcolor{green!70!black}{(+4.0)} & \multicolumn{1}{c|}{32.7 \textcolor{green!70!black}{(+7.2)}} & 34.5 \textcolor{green!70!black}{(+5.2)} & 33.9 \textcolor{green!70!black}{(+5.6)} & 69.4 \textcolor{green!70!black}{(+16.6)} & 34.9 \textcolor{green!70!black}{(+0.0)} & 61.4 \textcolor{green!70!black}{(+7.0)} & \multicolumn{1}{c|}{32.7 \textcolor{green!70!black}{(+7.2)}} & 36.8 \textcolor{green!70!black}{(+1.1)} & 37.2 \textcolor{green!70!black}{(+7.8)} & \multicolumn{1}{c|}{42.3 \textcolor{green!70!black}{(+10.2)}} & 37.1 \textcolor{green!70!black}{(+4.3)} \\
 &  & \multicolumn{1}{c|}{Gemini-3.1-Pro\cite{Gemini_3_1_pro}} & 53.5 \textcolor{green!70!black}{(+2.1)} & 38.6 \textcolor{green!70!black}{(+0.2)} & 31.4 \textcolor{gray!85!black}{(-4.1)} & \multicolumn{1}{c|}{46.9 \textcolor{green!70!black}{(+15.3)}} & 44.4 \textcolor{green!70!black}{(+0.3)} & 40.7 \textcolor{green!70!black}{(+1.5)} & 69.4 \textcolor{green!70!black}{(+13.8)} & 41.8 \textcolor{gray!85!black}{(-3.8)} & 70.2 \textcolor{green!70!black}{(+1.8)} & \multicolumn{1}{c|}{46.9 \textcolor{green!70!black}{(+15.3)}} & 48.0 \textcolor{gray!85!black}{(-0.5)} & 43.0 \textcolor{gray!85!black}{(-1.0)} & \multicolumn{1}{c|}{48.7 \textcolor{green!70!black}{(+10.1)}} & 46.2 \textcolor{green!70!black}{(+1.9)} \\ \hline
\multirow{2}{*}{\large Qwen3-VL-Embedding \cite{qwen3-vl-embedding}} & \multirow{2}{*}{2B} & \multicolumn{1}{c|}{ Qwen3-VL-32B \cite{qwen3-vl}} & 47.7 \textcolor{green!70!black}{(+8.5)} & 31.2 \textcolor{green!70!black}{(+5.6)} & 25.7 \textcolor{green!70!black}{(+2.5)} & \multicolumn{1}{c|}{38.8 \textcolor{green!70!black}{(+13.3)}} & 37.2 \textcolor{green!70!black}{(+7.9)} & 34.2 \textcolor{green!70!black}{(+5.9)} & 69.4 \textcolor{green!70!black}{(+16.6)} & 36.7 \textcolor{green!70!black}{(+1.8)} & 70.2 \textcolor{green!70!black}{(+15.8)} & \multicolumn{1}{c|}{38.8 \textcolor{green!70!black}{(+13.3)}} & 38.0 \textcolor{green!70!black}{(+2.3)} & 41.4 \textcolor{green!70!black}{(+12.0)} & \multicolumn{1}{c|}{44.6 \textcolor{green!70!black}{(+12.5)}} & 39.9 \textcolor{green!70!black}{(+7.1)} \\
 &  & \multicolumn{1}{c|}{Gemini-3.1-Pro\cite{Gemini_3_1_pro}} & 59.1 \textcolor{green!70!black}{(+7.7)} & 40.6 \textcolor{green!70!black}{(+2.2)} & 32.1 \textcolor{gray!85!black}{(-3.4)} & \multicolumn{1}{c|}{52.0 \textcolor{green!70!black}{(+20.4)}} & 48.3 \textcolor{green!70!black}{(+4.2)} & 46.7 \textcolor{green!70!black}{(+7.5)} & 72.2 \textcolor{green!70!black}{(+16.6)} & 45.2 \textcolor{gray!85!black}{(-0.4)} & 71.9 \textcolor{green!70!black}{(+3.5)} & \multicolumn{1}{c|}{52.0 \textcolor{green!70!black}{(+20.4)}} & 48.6 \textcolor{green!70!black}{(+0.1)} & 51.3 \textcolor{green!70!black}{(+7.3)} & \multicolumn{1}{c|}{52.8 \textcolor{green!70!black}{(+14.2)}} & 50.3 \textcolor{green!70!black}{(+6.0)} \\ \hline
\multirow{2}{*}{\large Qwen3-VL-Embedding \cite{qwen3-vl-embedding}} & \multirow{2}{*}{8B} & \multicolumn{1}{c|}{ Qwen3-VL-32B \cite{qwen3-vl}} & 52.2 \textcolor{green!70!black}{(+13.0)} & 32.1 \textcolor{green!70!black}{(+6.5)} & 29.4 \textcolor{green!70!black}{(+6.2)} & \multicolumn{1}{c|}{34.7 \textcolor{green!70!black}{(+9.2)}} & 41.2 \textcolor{green!70!black}{(+11.9)} & 39.2 \textcolor{green!70!black}{(+10.9)} & 66.7 \textcolor{green!70!black}{(+13.9)} & 39.2 \textcolor{green!70!black}{(+4.3)} & 71.9 \textcolor{green!70!black}{(+17.5)} & \multicolumn{1}{c|}{34.7 \textcolor{green!70!black}{(+9.2)}} & 43.4 \textcolor{green!70!black}{(+7.7)} & 43.8 \textcolor{green!70!black}{(+14.4)} & \multicolumn{1}{c|}{43.8 \textcolor{green!70!black}{(+11.7)}} & 42.7 \textcolor{green!70!black}{(+9.9)} \\
 &  & \multicolumn{1}{c|}{Gemini-3.1-Pro\cite{Gemini_3_1_pro}} & 63.2 \textcolor{green!70!black}{(+11.8)} & 46.3 \textcolor{green!70!black}{(+7.9)} & 37.3 \textcolor{green!70!black}{(+1.8)} & \multicolumn{1}{c|}{44.9 \textcolor{green!70!black}{(+13.3)}} & 53.0 \textcolor{green!70!black}{(+8.9)} & 50.8 \textcolor{green!70!black}{(+11.6)} & 80.6 \textcolor{green!70!black}{(+25.0)} & 50.3 \textcolor{green!70!black}{(+4.7)} & 73.7 \textcolor{green!70!black}{(+5.3)} & \multicolumn{1}{c|}{44.9 \textcolor{green!70!black}{(+13.3)}} & 54.5 \textcolor{green!70!black}{(+6.0)} & 53.3 \textcolor{green!70!black}{(+9.3)} & \multicolumn{1}{c|}{\cellcolor{green!10}\textbf{60.5} \textcolor{green!70!black}{(+21.9)}} & 53.7 \textcolor{green!70!black}{(+9.4)} \\ \hline
\multirow{2}{*}{\large ColQwen2-v1.0-merged \cite{faysse2024colpali}} & \multirow{2}{*}{2B} & \multicolumn{1}{c|}{ Qwen3-VL-32B \cite{qwen3-vl}} & 53.9 \textcolor{green!70!black}{(+14.7)} & 29.3 \textcolor{green!70!black}{(+3.7)} & 25.8 \textcolor{green!70!black}{(+2.6)} & \multicolumn{1}{c|}{38.8 \textcolor{green!70!black}{(+13.3)}} & 42.8 \textcolor{green!70!black}{(+13.5)} & 36.4 \textcolor{green!70!black}{(+8.1)} & 66.7 \textcolor{green!70!black}{(+13.9)} & 36.2 \textcolor{green!70!black}{(+1.3)} & 64.9 \textcolor{green!70!black}{(+10.5)} & \multicolumn{1}{c|}{38.8 \textcolor{green!70!black}{(+13.3)}} & 44.5 \textcolor{green!70!black}{(+8.8)} & 40.9 \textcolor{green!70!black}{(+11.5)} & \multicolumn{1}{c|}{46.7 \textcolor{green!70!black}{(+14.6)}} & 42.9 \textcolor{green!70!black}{(+10.1)} \\
 &  & \multicolumn{1}{c|}{Gemini-3.1-Pro\cite{Gemini_3_1_pro}} & 65.5 \textcolor{green!70!black}{(+14.1)} & 43.8 \textcolor{green!70!black}{(+5.4)} & 32.2 \textcolor{gray!85!black}{(-3.3)} & \multicolumn{1}{c|}{42.9 \textcolor{green!70!black}{(+11.3)}} & 54.1 \textcolor{green!70!black}{(+10.0)} & 50.9 \textcolor{green!70!black}{(+11.7)} & 75.0 \textcolor{green!70!black}{(+19.4)} & 45.4 \textcolor{gray!85!black}{(-0.2)} & 75.4 \textcolor{green!70!black}{(+7.0)} & \multicolumn{1}{c|}{42.9 \textcolor{green!70!black}{(+11.3)}} & 58.4 \textcolor{green!70!black}{(+9.9)} & 50.9 \textcolor{green!70!black}{(+6.9)} & \multicolumn{1}{c|}{50.4 \textcolor{green!70!black}{(+11.8)}} & 53.4 \textcolor{green!70!black}{(+9.1)} \\ \hline
\multirow{2}{*}{\large ColQwen2.5-v0.2 \cite{faysse2024colpali}} & \multirow{2}{*}{3B} & \multicolumn{1}{c|}{ Qwen3-VL-32B \cite{qwen3-vl}} & 55.7 \textcolor{green!70!black}{(+16.5)} & 33.5 \textcolor{green!70!black}{(+7.9)} & 33.6 \textcolor{green!70!black}{(+10.4)} & \multicolumn{1}{c|}{30.6 \textcolor{green!70!black}{(+5.1)}} & 42.4 \textcolor{green!70!black}{(+13.1)} & 48.7 \textcolor{green!70!black}{(+20.4)} & 75.0 \textcolor{green!70!black}{(+22.2)} & 43.6 \textcolor{green!70!black}{(+8.7)} & 75.4 \textcolor{green!70!black}{(+21.0)} & \multicolumn{1}{c|}{30.6 \textcolor{green!70!black}{(+5.1)}} & 47.8 \textcolor{green!70!black}{(+12.1)} & 44.7 \textcolor{green!70!black}{(+15.3)} & \multicolumn{1}{c|}{49.5 \textcolor{green!70!black}{(+17.4)}} & 45.2 \textcolor{green!70!black}{(+12.4)} \\
 &  & \multicolumn{1}{c|}{Gemini-3.1-Pro\cite{Gemini_3_1_pro}} & 64.5 \textcolor{green!70!black}{(+13.1)} & 47.3 \textcolor{green!70!black}{(+8.9)} & 38.8 \textcolor{green!70!black}{(+3.3)} & \multicolumn{1}{c|}{43.9 \textcolor{green!70!black}{(+12.3)}} & 54.9 \textcolor{green!70!black}{(+10.8)} & 49.6 \textcolor{green!70!black}{(+10.4)} & 77.8 \textcolor{green!70!black}{(+22.2)} & \cellcolor{green!10}\textbf{50.5} \textcolor{green!70!black}{(+4.9)} & 79.0 \textcolor{green!70!black}{(+10.6)} & \multicolumn{1}{c|}{43.9 \textcolor{green!70!black}{(+12.3)}} & 59.5 \textcolor{green!70!black}{(+11.0)} & 53.1 \textcolor{green!70!black}{(+9.1)} & \multicolumn{1}{c|}{50.3 \textcolor{green!70!black}{(+11.7)}} & 54.8 \textcolor{green!70!black}{(+10.5)} \\ \hline
\multirow{2}{*}{\large Jina-Embedding-v4 \cite{jina-embedding-v4}} & \multirow{2}{*}{3B} & \multicolumn{1}{c|}{ Qwen3-VL-32B \cite{qwen3-vl}} & 55.6 \textcolor{green!70!black}{(+16.4)} & 33.5 \textcolor{green!70!black}{(+7.9)} & 30.6 \textcolor{green!70!black}{(+7.4)} & \multicolumn{1}{c|}{32.7 \textcolor{green!70!black}{(+7.2)}} & 43.7 \textcolor{green!70!black}{(+14.4)} & 43.3 \textcolor{green!70!black}{(+15.0)} & 66.7 \textcolor{green!70!black}{(+13.9)} & 41.7 \textcolor{green!70!black}{(+6.8)} & 73.7 \textcolor{green!70!black}{(+19.3)} & \multicolumn{1}{c|}{32.7 \textcolor{green!70!black}{(+7.2)}} & 46.5 \textcolor{green!70!black}{(+10.8)} & 46.4 \textcolor{green!70!black}{(+17.0)} & \multicolumn{1}{c|}{43.4 \textcolor{green!70!black}{(+11.3)}} & 44.8 \textcolor{green!70!black}{(+12.0)} \\
 &  & \multicolumn{1}{c|}{Gemini-3.1-Pro\cite{Gemini_3_1_pro}} & 63.9 \textcolor{green!70!black}{(+12.5)} & 47.2 \textcolor{green!70!black}{(+8.8)} & 36.7 \textcolor{green!70!black}{(+1.2)} & \multicolumn{1}{c|}{48.0 \textcolor{green!70!black}{(+16.4)}} & 53.6 \textcolor{green!70!black}{(+9.5)} & 50.0 \textcolor{green!70!black}{(+10.8)} & 75.0 \textcolor{green!70!black}{(+19.4)} & 48.8 \textcolor{green!70!black}{(+3.2)} & \cellcolor{green!10}\textbf{84.2} \textcolor{green!70!black}{(+15.8)} & \multicolumn{1}{c|}{48.0 \textcolor{green!70!black}{(+16.4)}} & 57.8 \textcolor{green!70!black}{(+9.3)} & 53.9 \textcolor{green!70!black}{(+9.9)} & \multicolumn{1}{c|}{47.2 \textcolor{green!70!black}{(+8.6)}} & 54.5 \textcolor{green!70!black}{(+10.2)} \\ \hline
\multirow{2}{*}{\large ColNomic-Embed-Multimodal \cite{colnomic}} & \multirow{2}{*}{3B} & \multicolumn{1}{c|}{ Qwen3-VL-32B \cite{qwen3-vl}} & 55.8 \textcolor{green!70!black}{(+16.6)} & 34.2 \textcolor{green!70!black}{(+8.6)} & 33.4 \textcolor{green!70!black}{(+10.2)} & \multicolumn{1}{c|}{31.6 \textcolor{green!70!black}{(+6.1)}} & 44.0 \textcolor{green!70!black}{(+14.7)} & 46.7 \textcolor{green!70!black}{(+18.4)} & 69.4 \textcolor{green!70!black}{(+16.6)} & 42.0 \textcolor{green!70!black}{(+7.1)} & 73.7 \textcolor{green!70!black}{(+19.3)} & \multicolumn{1}{c|}{31.6 \textcolor{green!70!black}{(+6.1)}} & 47.4 \textcolor{green!70!black}{(+11.7)} & 45.9 \textcolor{green!70!black}{(+16.5)} & \multicolumn{1}{c|}{49.0 \textcolor{green!70!black}{(+16.9)}} & 45.5 \textcolor{green!70!black}{(+12.7)} \\
 &  & \multicolumn{1}{c|}{Gemini-3.1-Pro\cite{Gemini_3_1_pro}} & \cellcolor{green!10}\textbf{66.8} \textcolor{green!70!black}{(+15.4)} & \cellcolor{green!10}\textbf{51.8} \textcolor{green!70!black}{(+13.4)} & \cellcolor{green!10}\textbf{39.9} \textcolor{green!70!black}{(+4.4)} & \multicolumn{1}{c|}{42.9 \textcolor{green!70!black}{(+11.3)}} & \cellcolor{green!10}\textbf{57.0} \textcolor{green!70!black}{(+12.9)} & \cellcolor{green!10}\textbf{56.7} \textcolor{green!70!black}{(+17.5)} & \cellcolor{green!10}\textbf{83.3} \textcolor{green!70!black}{(+27.7)} & {50.3} \textcolor{green!70!black}{(+4.7)} & 80.7 \textcolor{green!70!black}{(+12.3)} & \multicolumn{1}{c|}{42.9 \textcolor{green!70!black}{(+11.3)}} & \cellcolor{green!10}\textbf{60.4} \textcolor{green!70!black}{(+11.9)} & \cellcolor{green!10}\textbf{57.3} \textcolor{green!70!black}{(+13.3)} & \multicolumn{1}{c|}{54.0 \textcolor{green!70!black}{(+15.4)}} & \cellcolor{green!10}\textbf{57.0} \textcolor{green!70!black}{(+12.7)} \\ \hline
\end{tabular}
}
\end{table}

\subsubsection{{End-to-End RAG Evaluation}}
{
Furthermore, we fed the Top-5 retrieved results (ranked by similarity score) from various retrieval methods into state-of-the-art closed- and open-source large models to evaluate the impact of an end-to-end RAG system on long-context ancient Chinese document understanding. As shown in Table~\ref{tab:combined_rag_eval}, the end-to-end question-answering performance positively correlates with the retrieval performance of the retriever.
}

{In the visual-based RAG setting, Gemini 3.1 Pro achieved an overall performance of 50.5 when paired with Top-5 results from the VisRAG-Ret retriever (Top-5 recall: 47.3\%), already surpassing the full-document baseline. Similarly, Qwen3-VL-32B outperformed its baseline when paired with the Gme-Qwen2-VL-Instruct retriever (Top-5 recall: 55.0\%), achieving a final score of 37.1. Moreover, as retrieval performance improves, the final end-to-end performance is expected to further increase. These results demonstrate that in long ancient document scenarios with visual modality input, a high-performance visual retriever can significantly enhance the model's comprehension of ancient texts; meanwhile, smaller models exhibit a higher dependency on retriever performance.
}

{Conversely, in the text-based RAG setting, none of the tested text retrievers yielded performance improvements—even though the best-performing Qwen3-Embedding achieved a Top-5 recall as high as 78.2\%. This indicates that existing state-of-the-art models already possess formidable long-text processing capabilities for ancient Chinese, allowing them to effectively filter out redundant text and accurately pinpoint evidence information. This phenomenon imposes a higher standard on text-modality retrievers: if the retrieved results fail to precisely contain the key evidence text, performing chunking and retrieval is even less effective than directly feeding the entire document into the model.}\\

\section{Conclusion}

In this paper, we introduce \textbf{LongHisDoc}, the first benchmark dedicated to Chinese long historical document understanding. It comprises 101 historical documents and 1,012 human-annotated question-answer pairs, covering four question types and three evidence sources. Through comprehensive evaluations of LLMs and LVLMs, we reveal that long historical document understanding remains challenging, particularly for cross-page reasoning, multi-hop evidence integration, and visual comprehension. Our results demonstrate that OCR-based text inputs generally outperform direct image inputs, although OCR-induced noise remains a limiting factor. Furthermore, evaluations of multimodal retrievers and end-to-end RAG systems highlight the potential of visual-based retrieval while emphasizing the importance of accurate evidence selection for effective document understanding. We hope LongHisDoc will serve as a valuable benchmark for advancing long-context reasoning, multimodal understanding, and retrieval-augmented generation for historical documents.



\section*{Acknowledgments}
We sincerely thank Professor Zhaopeng Wang and his team at Sichuan University for their invaluable support and assistance during the data annotation process.

\bibliography{references}

\clearpage

\appendix

\noindent\textbf{\LARGE Appendix}

\section{Datasheet for LongHisDoc}
\label{append-a}
\subsection{Motivation}

\textbf{1. For what purpose was the dataset created? Was there a specific task in mind? Was there a specific gap that needed to be filled? Please provide a description.}

A1: 
The purpose of the LongHisDoc benchmark is to quantify the understanding capabilities of existing Large Language Models (LLMs) and Large Vision-Language Models(LVLMs) on Chinese long historical document, providing a reference for the development of this field. Existing benchmarks for historical document \cite{xu2019casia, ma2020joint, NEURIPS2023_f7b424d2, shi2025large} primarily focus on traditional machine learning tasks such as text detection and recognition, layout analysis, and reading order prediction, while lacking benchmarks that evaluate models' understanding capabilities of historical documents. And existing benchmarks for document understanding \cite{tito2023hierarchical, tanaka2023slidevqa, van2023dude, ma2024mmlongbench, deng2024longdocurl} primarily focus on modern general documents, leaving historical documents unexplored. To fill this gap, we propose LongHisDoc, a pioneering benchmark specifically designed to evaluate the capabilities of LLMs and LVLMs in long-context historical document understanding tasks. This benchmark includes 101 historical documents across 10 categories, with 1,012 expert-annotated question-answer pairs covering four types, and the evidence for the questions is drawn from three modalities. Overall, LongHisDoc serve as a valuable resource to advance research in long-context historical document understanding, foster deeper engagement with cultural heritage and drive the development of AI systems for interpreting the rich tapestry of human history.

\subsection{Composition}

\textbf{1. What do the instances that comprise the dataset represent (e.g., documents, photos, people, countries)? Are there multiple types of instances (e.g., movies, users, and ratings; people and interactions between them; nodes and edges)? Please provide a description.}

A1: LongHisDoc consists of 101 historical documents, their corresponding Optical Character Recognition (OCR) results, and 1,012 QA pairs based on these documents. Each QA pair is represented in a dictionary format, containing the following keys: ID, question, answer, evidence pages, QA pair type, document type, and book name. OCR results and QA pairs are stored in a JSON file. For the historical documents, we provide the original PDF files and screenshots of each page.

\textbf{2. How many instances are there in total (of each type, if appropriate)?}

A2: The QA pairs in LongHisDoc are categorized as follows: 546 single-page QA pairs (evidence located on a single page), 196 cross-page QA pairs (evidence spread across two consecutive pages), 172 multi-hop QA pairs (evidence distributed across two or more non-contiguous pages), and 98 unanswerable QA pairs (no sufficient supporting evidence available in the document). 

\textbf{3. Does the dataset contain all possible instances or is it a sample (not necessarily random) of instances from a larger set? If the dataset is a sample, then what is the larger set? Is the sample representative of the larger set (e.g., geographic coverage)? If so, please describe how this represent ativeness was validated/verified. If it is not representative of the larger set, please describe why not (e.g., to cover a more diverse range of instances because instances were withheld or unavailable).}



{
A3: LongHisDoc does not contain all possible instances. Instead, it is a curated and quality-controlled subset constructed from the larger HisDoc-1B collection. HisDoc-1B \cite{shi2025large} serves as the source set and contains 40,281 historical documents. We first applied category-keyword matching based on category definitions and keywords derived from digital humanities resources, including Daizhige\cite{daizhige}, Shidianguji\footnote{https://www.shidianguji.com/} and Guoxue Master websites \footnote{https://www.guoxuedashi.net/}, resulting in 8,181 matched documents. We then reduced these matched documents to 200 candidates through human filtering based on document completeness, scanning quality, category coverage, layout structure, and suitability for long-context QA annotation. Finally, 101 documents were retained in LongHisDoc.
}\\
{
LongHisDoc is not a random sample of HisDoc-1B and is not intended to be a statistically representative subset of the full HisDoc-1B distribution. It is designed as a balanced, quality-controlled benchmark for long historical-document understanding. To improve coverage and mitigate category-keyword bias and book-title bias, we adopted multi-source keyword construction, keyword-level redundancy constraints, and stratified category coverage. Specifically, category keywords were derived from two independent digital humanities platforms and further verified manually after LLM-assisted extraction. We limited the number of selected documents associated with each keyword to at most three, in order to avoid over-sampling popular keywords or high-frequency book titles. The final benchmark maintains a relatively balanced distribution across 10 historical-document categories.
}\\
{
We further validated the reliability of keyword matching through manual sampling. Two domain experts randomly sampled 300 documents from the 8,181 matched documents, with 30 documents from each category, and examined whether the assigned categories were consistent with the actual document contents. The manual audit yielded a classification accuracy of 96.3\%. Therefore, LongHisDoc should be understood as a task-oriented benchmark that is manually verified and quality-controlled, rather than a complete or random sample covering the full distribution of HisDoc-1B.
}

\textbf{4. What data does each instance consist of?}

A4: The LongHisDoc benchmark consists of 1,012 QA pairs based on historical documents. Each QA pair is represented in a dictionary format, containing the following keys: ID, question, answer, evidence pages, QA pair type, document type, and book name.

\textbf{5. Is there a label or target associated with each instance?}

A5: The LongHisDoc benchmark consists of 101 historical documents, their corresponding Optical Character Recognition (OCR) results, and 1,012 QA pairs based on these documents. In addition to the QA pairs, we include supplementary information such as the document title, document category, and QA type.

\textbf{6. Is any information missing from individual instances? If so, please provide a description, explaining why this information is missing (e.g., because it was unavailable). This does not include intentionally removed information, but might include, e.g., redacted text.}

A6: No, each of our data instances contains comprehensive information.

\textbf{7. Are relationships between individual instances made explicit (e.g., users’ movie ratings, social network links)? If so, please describe how these relationships are made explicit.}

A7: No, our data instances are independent of each other and have no interrelations. 

\textbf{8. Are there recommended data splits (e.g., training, development/validation, testing)? If so, please provide a description of these splits, explaining the rationale behind them.}

A8: No, our data is utilized solely for the purpose of testing and not for training.

\textbf{9. Are there any errors, sources of noise, or redundancies in the dataset? If so, please provide a
description.}

A9: In LongHisDoc, we have employed experts in the field of historical documents and have conducted extensive manual calibration to minimize errors and noise as much as possible, though we cannot guarantee that every data sample is flawless.

\textbf{10. Is the dataset self-contained, or does it link to or otherwise rely on external resources (e.g., websites, tweets, other datasets)? }

A10: Yes, the LongHisDoc benchmark is self-contained.

\textbf{11. Does the dataset contain data that might be considered confidential (e.g., data that is protected by legal privilege or by doctorpatient confidentiality, data that includes the content of individuals non-public communications)? If so, please provide a description.}

A11: No, we construct the LongHisDoc benchmark from publicly accessible sources.

\textbf{12. Does the dataset contain data that, if viewed directly, might be offensive, insulting, threatening, or might otherwise cause anxiety? If so, please describe why.}

A12: No, the LongHisDocs benchmark does not contain any content that is offensive, insulting, threatening, or might otherwise cause anxiety.

\textbf{13. Does the dataset relate to people?}

A13: No, the LongHisDoc benchmark has nothing to do with people.

\subsection{Collection Process}

\textbf{1. How was the data associated with each instance acquired? Was the data directly observable (e.g., raw text, movie ratings), reported by subjects (e.g., survey responses), or indirectly inferred/derived from other data (e.g., part-of-speech tags, model-based guesses for age or language)? If data was reported by subjects or indirectly inferred/derived from other data, was the data validated/verified? If so, please describe how.}

A1: The data construction and specific sources are thoroughly described in Section~3 and Appendix~\ref{appendix-b}. We construct our dataset based on HisDoc-1B \cite{shi2025large}, categorizing the historical documents into ten expert-defined classes. All data construction procedures have been carried out with the permission of the authors.

\textbf{2. What mechanisms or procedures were used to collect the data (e.g., hardware apparatus or sensor, manual human curation, software program, software API)? How were these mechanisms or procedures validated?}

A2: Our data collection is carried out through three mechanisms: manual human curation, keyword matching program, and the API of LLMs and LVLMs.

\textbf{3. Over what timeframe was the data collected? }

A3: Our data collection spans a period of four months.

\textbf{4. Did you collect the data from the individuals in question directly, or obtain it via third parties or other sources (e.g., websites)?}

A4: We primarily collect data from open-source datasets, and all data construction procedures have been conducted with the permission of the dataset authors.

\subsection{Preprocessing/cleaning/labeling}

\textbf{1. Was any preprocessing/cleaning/labeling of the data done (e.g., discretization or bucketing, tokenization, part-of-speech tagging, SIFT feature extraction, removal of instances, processing of missing values)? If so, please provide a description. If not, you may skip the remaining questions in this section.}

A1: Yes, the detail is discussed in Section~3.

\textbf{2. Was the “raw” data saved in addition to the preprocessed/cleaned/labeled data (e.g., to support unanticipated future uses)? If so, please provide a link or other access point to the “raw” data.}

A2: Yes, we provide the original PDF files of the historical documents along with their titles to facilitate traceability.

\subsection{Uses}

\textbf{1. Has the dataset been used for any tasks already? If so, please provide a description.}

A1: No, the LongHisDoc benchmark has not appeared in previous works.

\textbf{2. What (other) tasks could the dataset be used for?}

A2: The LongHisDoc benchmark can be used to evaluate LLMs, LVLMs, and RAG systems.

\textbf{3. Are there tasks for which the dataset should not be used? If so, please provide a description.}

A3: To prevent data leakage and ensure the fairness of the evaluation \cite{liu2024datasets,chang2024evalsurvey}, the LongHisDoc benchmark cannot be used for training LLMs/LVLMs.

\subsection{Distribution}

\textbf{1. How will the dataset will be distributed (e.g., tarball on website, API, GitHub)? }

A1: Our dataset is publicly available at \url{https://huggingface.co/datasets/qweq12433454/LongHisDoc}.

\textbf{2. When will the dataset be distributed?}

A2: Our dataset has been already distributed.

\textbf{3. Will the dataset be distributed under a copyright or other intellectual property (IP) license, and/or under applicable terms of use (ToU)? }

A3: For details, please refer to Appendix~\ref{appendix-G}.

\subsection{Maintenance}

\textbf{1. Who will be supporting/hosting/maintaining the dataset?}

A1: The authors of this paper will maintain the dataset in collaboration with expert annotators.

\textbf{2. How can the owner/curator/manager of the dataset be contacted (e.g., email address)?}

A2: The manager can be contacted through the Hugging Face or Github.

\textbf{3. Will the dataset be updated (e.g., to correct labeling errors, add new instances, delete instances)? If so, please describe how often, by whom, and how updates will be communicated to dataset consumers (e.g., mailing list, GitHub)?}

A3: We will continuously maintain our benchmark and once any errors are found or more reasonable evaluation methods are identified, we will promptly update the LongHisDoc benchmark.

\textbf{4. Will older versions of the dataset continue to be supported/hosted/maintained? If so, please describe how. If not, please describe how its obsolescence will be communicated to dataset consumers.}

A4: We will maintain all versions of the LongHisDoc benchmark on Hugging Face.

\textbf{5. If others want to extend/augment/build on/contribute to the dataset, is there a mechanism for them to do so? If so, please provide a description. Will these contributions be validated/verified? If so, please describe how. If not, why not? Is there a process for communicating/distributing these contributions to dataset consumers? If so, please provide a description.}

A5: If other researchers or individuals are interested in extending, augmenting, building on, or contributing to the dataset, they should contact us via email, Hugging Face, or Github, clearly articulating their intentions and requesting our consent prior to any further actions.

\newpage

\section{Dataset}
\label{appendix-b}

\subsection{Historical Document Type}
\label{appendix-b1}
The following presents the definitions of the 10 historical document categories in LongHisDoc, along with their corresponding keywords：
\begin{itemize}
\item \begin{CJK*}{UTF8}{gbsn}史藏\end{CJK*} (History)

    \textbf{Definition:} History includes books that record the history of China.
    
    \textbf{Key words:} {\begin{CJK*}{UTF8}{gbsn}['春秋左传', '资治通鉴', '史记', '二十四史', '三国志', '汉书', '五代史', '战国策', '文史通义','竹书纪年']\end{CJK*}}

\item \begin{CJK*}{UTF8}{gbsn}诗藏\end{CJK*} (Poetry)

    \textbf{Definition:} Poetry encompass ancient poems of various forms and from different dynasties.
    
    \textbf{Key words:} {\begin{CJK*}{UTF8}{gbsn}['唐诗', '宋诗', '清诗', '明诗', '元诗', '六十种曲', '元曲', '对联', '声律', '楚辞', '文赋', '词话', '词集', '诗话']\end{CJK*}}

\item \begin{CJK*}{UTF8}{gbsn}佛藏\end{CJK*} (Buddhism)

    \textbf{Definition:} Buddhism refers to religious classics belonging to, or associated with, Buddhism and its traditions. 
    
    \textbf{Key words:} {\begin{CJK*}{UTF8}{gbsn}['大乘', '乾隆藏', '大藏经', '嘉兴藏', '续藏经', '金刚经','法华经','佛说', '波罗蜜','菩提']\end{CJK*}}

\item \begin{CJK*}{UTF8}{gbsn}儒藏\end{CJK*} (Confucianism)

    \textbf{Definition:} Confucianism includes books that represent Confucian thought, exemplified by Confucius and Mencius. 
    
    \textbf{Key words:} {\begin{CJK*}{UTF8}{gbsn}['孔子','孟子', '论语', '四书五经', '中庸', '尚书', '小学', '幼学', '孝经', '弟子规']\end{CJK*}}

\item \begin{CJK*}{UTF8}{gbsn}道藏\end{CJK*} (Taoism)

    \textbf{Definition:} Taoism refers to a series of ancient Chinese works that belong to Taoist thought and its traditions.
    
    \textbf{Key words:} {\begin{CJK*}{UTF8}{gbsn}['道德经','庄子', '老子',"列子", '太上', "太平",'上清', '道家', '道法','冲虚']\end{CJK*}}

\item \begin{CJK*}{UTF8}{gbsn}集藏\end{CJK*} (Anthology)

    \textbf{Definition:} Anthology is a collection of works such as novels, prose, and biographies.
    
    \textbf{Key words:} {\begin{CJK*}{UTF8}{gbsn}['文集', '演义', '红楼梦', '西游记', '水浒', '选集', '宝卷', '文评', '笑林', '英雄传']\end{CJK*}}

\item \begin{CJK*}{UTF8}{gbsn}易藏\end{CJK*} ({Yizang})

    \textbf{Definition:} {Yizang} contains books that record the methods used for divination in ancient China. 
    
    \textbf{Key words:} {\begin{CJK*}{UTF8}{gbsn}['易经', '易学', '易传', '易理', '周易', '占卜', '八卦', '术数', '六壬', '奇门', '遁甲', '卜法']\end{CJK*}}

\item \begin{CJK*}{UTF8}{gbsn}艺藏\end{CJK*} (Literature)

    \textbf{Definition:} Literature includes works related to Chinese classical literary activities, such as zither, chess, calligraphy, and painting. 
    
    \textbf{Key words:} {\begin{CJK*}{UTF8}{gbsn}['琴', '棋', '画','书法', '太极', '武术', '拳法', '剑法', '丸经', '印说', '乐府', '茶谱', '酒谱','乐书']\end{CJK*}}

\item \begin{CJK*}{UTF8}{gbsn}子藏\end{CJK*} (Philosophy)

    \textbf{Definition:} Philosophy is a collection of writings by renowned scholars of ancient China, representing the crystallization of Chinese civilization's wisdom.
    
    \textbf{Key words:} {\begin{CJK*}{UTF8}{gbsn}['诸子百家', '兵法', '齐民要术', '法家','墨家','法家', '农家', '管子', '韩非子', '算法', '算术', '物理']\end{CJK*}}
\item \begin{CJK*}{UTF8}{gbsn}医藏\end{CJK*} (Medicine)

    \textbf{Definition:} Medicine contains works on traditional Chinese medicine. 
    
    \textbf{Key words:} {\begin{CJK*}{UTF8}{gbsn}['中医', '伤寒', '本草', '医方', '医学', '儿科', '眼科', '黄帝内经', '神农', '扁鹊']\end{CJK*}}

\end{itemize}

\subsection{{Validation of keyword-matching accuracy}}
{To verify the reliability of the automated keyword-matching results, we conducted a manual sampling inspection. At this stage, keyword matching retrieved 8,181 candidate documents from HisDoc-1B. We then invited two domain experts to randomly sample 300 documents from the matched results, with 30 documents per category, and assess whether each document was assigned to the correct category based on its actual content. For samples on which the two experts disagreed, a final consensus was reached through discussion. The manual review shows a classification accuracy of 96.3\%, indicating that the matching strategy based on authoritative metadata and category keywords is highly reliable for historical-document categorization.}

\subsection{{Details of Historical Documents in LongHisDoc}}

{
As shown in Table~\ref{tab:ancient_books_all}, we present detailed information on all historical documents in LongHisDoc, including the document title, publication dynasty, and author's dynasty.
According to our statistics, the publication dynasties of the historical documents included in LongHisDoc span from the Song dynasty (960–1279) to the Republican period (1912-1949), covering approximately 900 years and six publication dynasties, with Ming and Qing editions accounting for the majority. The time span of the authors’ dynasties is even broader, ranging from the Spring and Autumn period (approximately 770 BCE) to the Republican period (1912-1949), covering approximately 2,700 years, and encompassing 21 distinct author dynasties.
}

\begin{CJK*}{UTF8}{gkai}

\begin{longtable}{p{0.52\linewidth} >{\centering\arraybackslash}p{0.24\linewidth} >{\centering\arraybackslash}p{0.24\linewidth}}
    \caption{{Details of Historical Documents in LongHisDoc. Zoom in for better view.}} \label{tab:ancient_books_all} \\
    \toprule
    \textbf{Historical Document Title} & \textbf{Publication Dynasty} & \textbf{Author's Dynasty} \\
    \midrule
    \endfirsthead

    \multicolumn{3}{c}{\small\itshape（Continued）} \\
    \toprule
    \textbf{Historical Document Title} & \textbf{Publication Dynasty} & \textbf{Author's Dynasty} \\
    \midrule
    \endhead

    \midrule
    \multicolumn{3}{r}{\small\itshape Continued on next page\dots} \\
    \endfoot

    \bottomrule
    \endlastfoot

    
    \multicolumn{3}{c}{\textbf{Buddhism}} \\
    \midrule
    《二经同卷2卷.佛说波斯匿王太后崩尘岔身经1卷.西晋.释法炬译.须摩提女经1卷.吴.释支谦译.美国加利福尼亚大学伯克利分校藏宋刊本>古籍/》 & 宋 (Song) & 西晋/三国·吴 (Western Jin / Three Kingdoms - Wu) \\
    《佛母出生三法藏般若波罗蜜多经.存卷第20.卷第25.宋.释施护等译.美国加利福尼亚大学伯克利分校藏元至元南山普宁寺刊本》 & 元 (Yuan) & 宋 (Song) \\
    《佛说佛名经二》 & 明 (Ming) & 北魏/元魏 (Northern Wei) \\
    《佛说仁王护国般若波罗蜜经2卷.后秦.释鸠摩罗什译.美国加利福尼亚大学伯克利分校藏明万历20年（1592）刊本》 & 明 (Ming) & 后秦 (Later Qin) \\
    《佛说大自在天子因地经.1卷.佛说如意摩尼陀罗经.1卷.宋释施护译.赵城广胜寺金版大藏经之一.金刊本》 & 金 (Jin) & 宋 (Song) \\
    《佛说观无量寿佛经疏2卷.刘宋.释畺良耶舍译.隋.释智顗疏.美国加利福尼亚大学伯克利分校藏明刊本》 & 明 (Ming) & 刘宋/隋 (Liu Song / Sui) \\
    《大般若波罗蜜多经.存卷第299.唐.释玄奘译.美国加利福尼亚大学伯克利分校藏元至元南山普宁寺刊本》 & 元 (Yuan) & 唐 (Tang) \\
    《大般若波罗蜜多经四十》 & 唐 (Tang) & 唐 (Tang) \\
    《金刚般若波罗蜜经注解.鸠摩罗什译;宗泐,如同注.[1600]1册》 & 明 (Ming) & 后秦/明 (Later Qin / Ming) \\
    
    \midrule
    \multicolumn{3}{c}{\textbf{Confucianism}} \\
    \midrule
    《论语集解著者何晏（魏）[数量]2册[书志事项]刊本（序刊）》 & 清 (Qing) & 三国·魏 (Three Kingdoms - Wei) \\
    《五经衷要春秋衷要\_李式谷辑南海叶梦龙风满楼》 & 清 (Qing) & 清 (Qing) \\
    《四库全书珍本初集经部五经总义类之十三经注解正字27》 & 清 (Qing) & 清 (Qing) \\
    《四书集注著者朱熹（宋）元延祐赵凤仪刊本/四书集注13》 & 元 (Yuan) & 宋 (Song) \\
    《扬子太玄经选者扬雄（汉）／校订者王道焜（明）／校订者朱钦明（明）明天启06年刊本/扬子太玄经1》 & 明 (Ming) & 汉 (Han) \\
    《论语孟子或问著者朱熹（宋）正保04年刊本/论语孟子或问5》 & 清 (Song) & 宋 (Song) \\
    《论语经正录》(清)王肇晋撰 & 清 (Qing) & 清 (Qing) \\
    《通志堂经解140种1860卷清康熙12年》 & 清 (Qing) & 清 (Qing) \\
    
    \midrule
    \multicolumn{3}{c}{\textbf{Medicine}} \\
    \midrule
    《新刊黄帝内经灵枢卷12》 & 元 (Yuan) & 先秦/汉 (Pre-Qin / Han) \\
    《本草纲目选者李时珍（明）／校订者稻生若水[数量]45册[书志事项]刊本,正德04年》 & 明 (Ming) & 明 (Ming) \\
    《医学六要选者张三锡（明）／校订者王肯堂（明）[数量]39册[书志事项]刊本,江户[旧藏者]昌平坂学问所》 & 清 (Qing) & 明 (Ming) \\
    《古今医统正脉全书编者吴勉学（明）[数量]75册[书志事项]刊本（序刊）,明万历,明万历29年[旧藏者]医学馆》 & 明 (Ming) & 明 (Ming) \\
    《医心方\_卷第十七》 & 清 (Qing) & 日本平安时代 (Heian Period) \\
    《伤寒论刘氏传（1-4卷）/伤寒论刘氏传卷一》 & 清 (Qing) & 东汉 (Eastern Han) \\
    《伤寒论疏义（1-10卷）/伤寒论疏义卷二》 & 清 (Qing) & 清 (Qing) \\
    《保婴全书（1-20卷）/保婴全书卷三》 & 明 (Ming) & 明 (Ming) \\
    《傅氏眼科审视瑶函六\_傅仁宇撰》 & 明 (Ming) & 明 (Ming) \\
    《扁鹊脉书难经一\_熊庆笏撰》 & 清 (Qing) & 清 (Qing) \\
    《新刊黄帝内经灵枢卷04》 & 元 (Yuan) & 先秦/汉 (Pre-Qin / Han) \\
    《神农本经会通九\_滕弘辑》 & 明 (Ming) & 明 (Ming) \\
    《秘传眼科龙木医书总论》(明)葆光道人撰 & 明 (Ming) & 明 (Ming) \\
    
    \midrule
    \multicolumn{3}{c}{\textbf{History}} \\
    \midrule
    《音注全文春秋括例始末左传句读直解著者林尧叟（宋）》 & 清 (Qing) & 宋 (Song) \\
    《明史稿选者王鸿绪（清）》 & 清 (Qing) & 清 (Qing) \\
    《晏子春秋著者晏婴（齐）》 & 明 (Ming) & 春秋 (Spring and Autumn) \\
    《史记选者司马迁（汉）／其他明钟惺删定[数量]15册[书志事项]刊本,宽政12年》 & 清 (Qing) & 西汉 (Western Han) \\
    《弇州史料选者王世贞（明）／编者董复表（明）[数量]30册[书志事项]刊本,明[旧藏者]昌平坂学问所1》 & 明 (Ming) & 明 (Ming) \\
    《辽史选者脱脱（元）[数量]12册[书志事项]刊本,明万历,明万历34年》 & 元 (Yuan) & 元 (Yuan) \\
    《续资治通鉴著者毕沅（清）[数量]64册[书志事项]刊本（序刊）,清嘉庆,清嘉庆06年[旧藏者]昌平坂学问所》 & 清 (Qing) & 清 (Qing) \\
    《后汉书选者范晔（宋）／注释者李贤（唐）／著者司马彪（晋）／注释者刘昭（梁）[数量]20册[书志事项]刊本,明崇祯,明崇祯16年,汲古阁》 & 明 (Ming) & 宋 (Song) \\
    《哈佛监本21史\_宋书100卷.04.梁.沈约撰》 & 明 (Ming) & 梁 (Liang) \\
    《明实录明武宗实录六》 & 明 (Ming) & 明 (Ming) \\
    《太平经国之书十一卷,首一卷 郑伯谦撰.夏小正戴氏传四卷 傅崧卿注.通志堂,清康熙19[1680].2册》 & 清 (Qing) & 宋 (Song) \\
    
    \midrule
    \multicolumn{3}{c}{\textbf{Philosophy}} \\
    \midrule
    《墨子[16卷].灵岩山馆.乾隆》 & 清 (Qing) & 战国 (Warring States) \\
    《韩非子[数量]8册[书志事项]刊本,明[旧藏者]昌平坂学问所》 & 明 (Ming) & 战国 (Warring States) \\
    《四库全书珍本初集子部艺术类之六艺之一录131》 & 清 (Qing) & 清 (Qing) \\
    《太学新增合璧联珠声律万卷菁华二十四\_李昭辑》 & 宋 (Song) & 宋 (Song) \\
    《帝王经世图谱\_广雅书局》 & 清 (Qing) & 宋 (Song) \\
    《新刊详明算法朝鲜刊本/新刊详明算法1》 & 元 (Yuan) & 元 (Yuan) \\
    《洴澼百金方》(清)惠麓酒民编 & 清 (Qing) & 清 (Qing) \\
    《浮山此藏轩物理小识选者方以智（明）》 & 明 (Ming) & 明 (Ming) \\
    《纂图互注荀子20卷.唐.杨倞注.美国加利福尼亚大学伯克利分校藏元刻本》 & 元 (Yuan) & 战国 (Warring States) \\
    《诸葛孔明异传兵法注解评林著者章婴（明）万治04年刊本/诸葛孔明异传兵法注解评林3》 & 明 (Ming) & 明 (Ming) \\
    
    \midrule
    \multicolumn{3}{c}{\textbf{Yizang}} \\
    \midrule
    《池纪解编》 & 明 (Ming) & 明 (Ming) \\
    《八卦余生四\_邓梦文撰》 & 明 (Ming) & 明 (Ming) \\
    《周易古今文全书明万历20年刊本/周易古今文全书1》 & 明 (Ming) & 明 (Ming) \\
    《四库全书珍本初集子部术数类之观物篇解06》 & 民国 (Republic of China) & 宋 (Song) \\
    《易经详说三十八\_冉觐祖撰》 & 清 (Qing) & 清 (Qing) \\
    《笺元遁甲句解烟波钓叟歌.1卷.赵普撰》 & 明 (Ming) & 宋 (Song) \\
    《明道易经》 & 清 (Qing) & 清 (Qing) \\
    《易学 一卷 王湜撰》 & 宋 (Song) & 宋 (Song) \\
    
    \midrule
    \multicolumn{3}{c}{\textbf{Literature}} \\
    \midrule
    《丛书集成初编0708灵棋经》 & 民国 (Republic of China) & 汉/魏晋 (Han / Wei-Jin) \\
    《丛书集成初编1681投壶仪节丸经》 & 民国 (Republic of China) & 宋/元 (Song / Yuan) \\
    《太古正音琴经》(明)张大命辑 & 明 (Ming) & 明 (Ming) \\
    《孙氏书画钞》(明)孙凤撰 & 明 (Ming) & 明 (Ming) \\
    《岭南玉社丛书.第1集.上下册.广州岭南玉社编.民国14年.1925年>古籍善本影印\textbackslash{}金石书法\textbackslash{}》 & 民国 (Republic of China) & 民国 (Republic of China) \\
    《乐府原三\_徐献忠撰》 & 明 (Ming) & 明 (Ming) \\
    《琴谱新声》(清)曹尚絅等辑 & 清 (Qing) & 清 (Qing) \\
    《粤东金石略.9卷.附九曜石考.2卷.清翁方纲编.石洲草堂.乾隆36年刊>古籍善本影印\textbackslash{}金石书法\textbackslash{}》 & 清 (Qing) & 清 (Qing) \\
    《翰苑印林.四卷.明.吴日章编.明崇祯七年刊钤印本>古籍善本影印\textbackslash{}金石书法\textbackslash{}》 & 明 (Ming) & 明 (Ming) \\
    《茶书编者喻政（明）明万历40年刊本/茶书1》 & 明 (Ming) & 明 (Ming) \\
    《大还阁琴谱6卷万峯阁指法閟笺1卷溪山琴况1卷.清.徐谼撰.清.夏溥辑.美国加利福尼亚大学伯克利分校藏清康熙12年（1673）大还阁刊本》 & 清 (Qing) & 清 (Qing) \\
    
    \midrule
    \multicolumn{3}{c}{\textbf{Poetry}} \\
    \midrule
    《元诗选编者顾奎光（清）[数量]4册[书志事项]刊本（序刊）,清乾隆,清乾隆16年[旧藏者]林家（大学头）》 & 清 (Qing) & 清 (Qing) \\
    《宋诗纪事编者厉额（清）／编者马日琯（清）[数量]32册[书志事項]刊本（序刊）,清乾隆,清乾隆11年[旧藏者]林家（大学头）》 & 清 (Qing) & 清 (Qing) \\
    《元曲选十三\_臧懋循辑》 & 明 (Ming) & 明 (Ming) \\
    《六十种曲》 & 明 (Ming) & 明 (Ming) \\
    《包公案词话八种五》 & 明 (Ming) & 明 (Ming) \\
    《忠雅堂文集忠雅堂词集\_蒋士铨辑》 & 清 (Qing) & 清 (Qing) \\
    《楚辞二十卷》 & 明 (Ming) & 战国/汉 (Warring States / Han) \\
    《津逮秘书紫薇诗话一卷吕本中撰石林诗话一卷叶梦得撰毛氏汲古阁明崇祯》 & 明 (Ming) & 宋 (Song) \\
    《皇明诗选13卷明崇祯6v》 & 明 (Ming) & 明 (Ming) \\
    《皇清诗选三十卷首一卷清康熙44年》 & 清 (Qing) & 清 (Qing) \\
    《唐诗类苑 二百卷 张之象纂辑;赵应元等编次;王彻补订.曹仁孙,明万历辛丑》 & 明 (Ming) & 明 (Ming) \\
    
    \midrule
    \multicolumn{3}{c}{\textbf{Taoism}} \\
    \midrule
    《太上老子道德经4卷. 明》 & 明 (Ming) & 春秋 (Spring and Autumn Period) \\
    《锲五代萨真人得道咒枣记著者邓志谟（明）[数量]2册[书志事项]刊本,明,萃庆堂[旧藏者]红叶山文库》 & 明 (Ming) & 明 (Ming) \\
    《太古遗音一\_杨抡辑》 & 明 (Ming) & 明 (Ming) \\
    《老子通著者沈一贯（明）明万历27年刊本/老子通1》 & 明 (Ming) & 明 (Ming) \\
    《庄子翼》 & 明 (Ming) & 明 (Ming) \\
    《性命双修万神圭旨.性命圭旨.4卷.尹真人口述.清康熙9年棣鄂堂刊本》 & 清 (Qing) & 明 (Ming) \\
    《老子元翼》 & 明 (Ming) & 明 (Ming) \\
    
    \midrule
    \multicolumn{3}{c}{\textbf{Anthology}} \\
    \midrule
    《新刊大宋演义中兴英烈传编者熊大木（明）／编者李春芳（明）[数量]10册[书志事项]刊本,明嘉靖,明嘉靖31年,杨氏清白堂[旧藏者]红叶山文库》 & 明 (Ming) & 明 (Ming) \\
    《新刻钟伯敬先生批评封神演义著者许仲琳（明）[数量]20册[书志事项]刊本,明,舒文渊[旧藏者]红叶山文库》 & 明 (Ming) & 明 (Ming) \\
    《七十二朝人物演义[数量]16册[书志事项]刊本,明[旧藏者]红叶山文库》 & 明 (Ming) & 明 (Ming) \\
    《东莱吕太史别集16卷.015卷.宋.吕祖谦撰.美国加利福尼亚大学伯克利分校藏宋嘉泰4年（1204）吕乔年刻元明递修本》 & 明 (Ming) & 宋 (Song) \\
    《京本增补校正全像忠义水浒志传评林明刊本/京本增补校正全像忠义水浒志传评林5》 & 明 (Ming) & 元 (Yuan) \\
    《儿女英雄传》(清)文康撰 & 清 (Qing) & 清 (Qing) \\
    《家堂宝卷\_黄福钦》 & 清 (Qing) & 清 (Qing) \\
    《拟山园选集三十三\_王铎撰》 & 清 (Qing) & 清 (Qing) \\
    《望溪先生文集》(清)方苞撰 & 清 (Qing) & 清 (Qing) \\
    《李卓吾先生批评西游记》(明)吴承恩撰 & 明 (Ming) & 明 (Ming) \\
    《胡祭酒集一\_胡俨撰》 & 明 (Ming) & 明 (Ming) \\
    《茶香室丛钞》(清)俞樾撰 & 清 (Qing) & 清 (Qing) \\
    《豫章先生文集著者黄庭坚（宋）宋刊本》 & 宋 (Song) & 宋 (Song) \\

\end{longtable}

\end{CJK*}

\subsection{More Statistics of LongHisDoc}
\label{appendix-b2}
\begin{figure*}[htbp]
\centering
\subfigure{%
    \resizebox*{6.7cm}{!}{\includegraphics{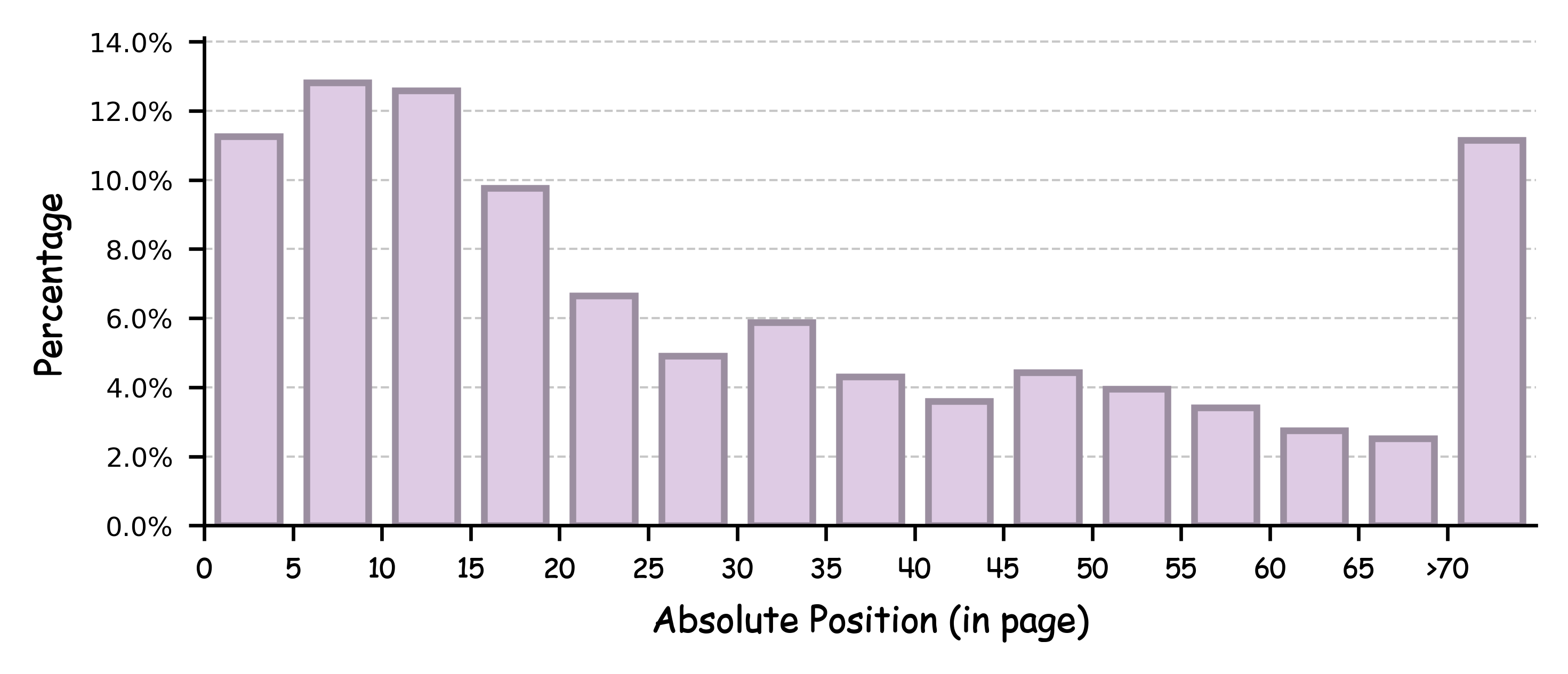}}} 
    \label{fig:panel-3-a}
\subfigure{%
    \resizebox*{6.7cm}{!}{\includegraphics{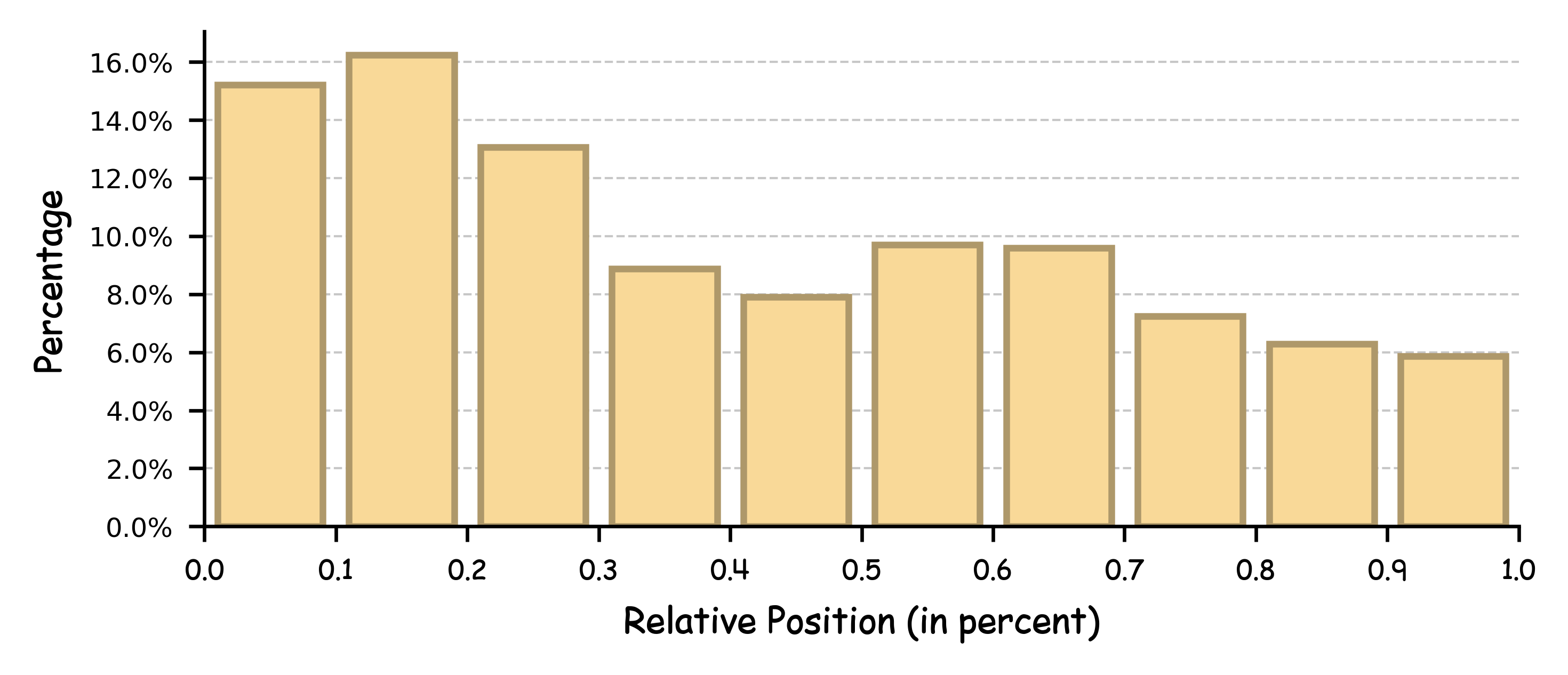}}} \label{fig:panel-3-b}
\subfigure{%
    \resizebox*{6.7cm}{!}{\includegraphics{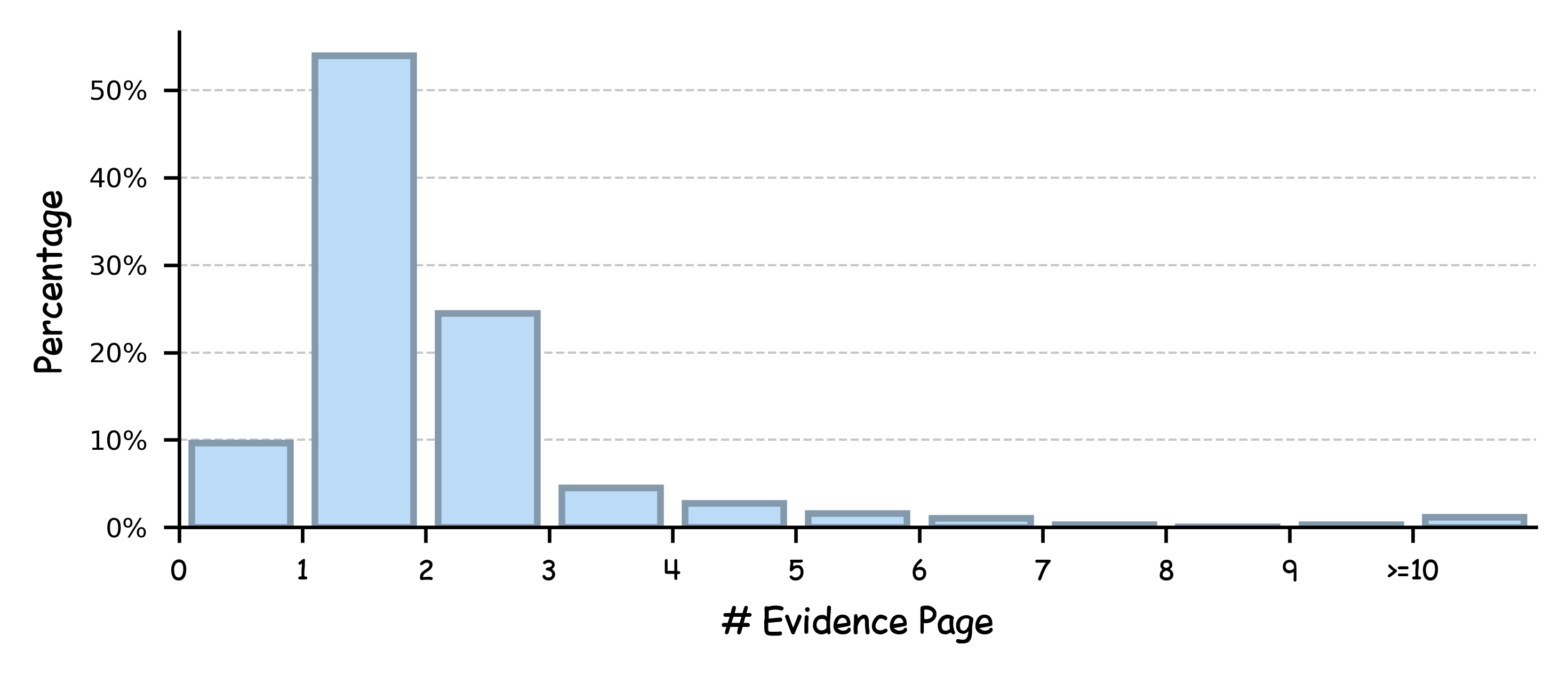}}
    \label{fig.2.c}} 
\subfigure{%
    \resizebox*{6.7cm}{!}{\includegraphics{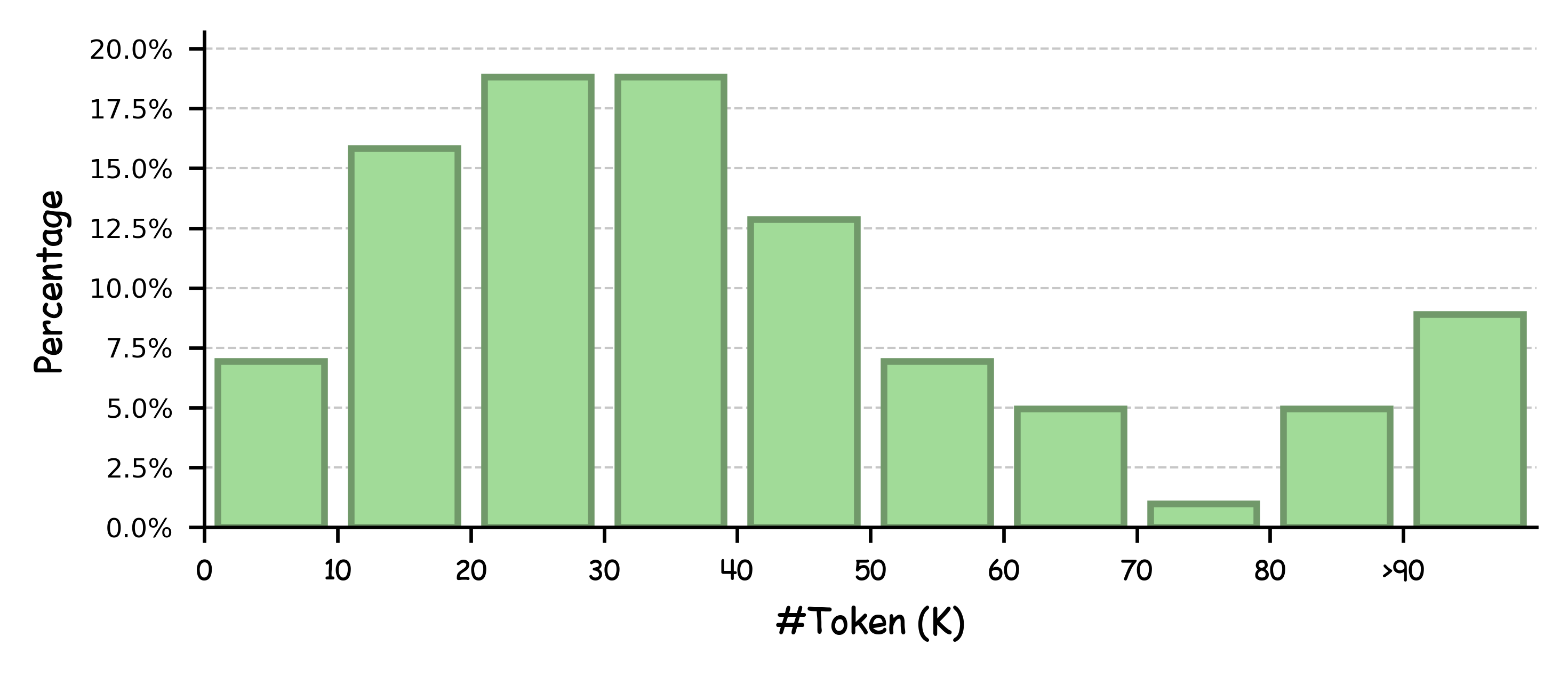}}
    \label{fig.2.d}} 
\caption{More distribution characteristics of LongHisDoc. \textbf{Top Left}: Absolute position of answer evidences (the page index). \textbf{Top Right}: Relative position (the page index / document page number). \textbf{Bottom Left}: Evidence page number of each question. \textbf{Bottom Right}: Document textual token number. }
\label{fig:more_character_dist}
\end {figure*}

Figure \ref{fig:more_character_dist} illustrate our dataset absolute and relative position of answer evidences, evidence pages length, and document textual token number.

\section{Evaluation}
\label{appendix-C}
\subsection{Evaluated Model}
Details of all evaluated LLMs and LVLMs are shown in Table \ref{table:model_details}. And details of all evaluated retrievers are presented in Table \ref{table:retriever_details}

\begin{table*}[h]
\caption{Details of all evaluated LLMs and LVLMs. Zoom in for better view.}
\label{table:model_details}
\resizebox{\textwidth}{!}{
\begin{tabular}{lccccl}
\hline
\multicolumn{1}{c}{Model} &
  Size &
  \multicolumn{1}{l}{Context Window} &
  Access &
  \multicolumn{1}{c}{API Version} &
  Website \\ \hline
LLaVA-V1.6-Mistral \cite{li2024llava} &
  7B &
  32K &
  Weights &
  - &
  https://huggingface.co/liuhaotian/llava-v1.6-mistral-7b \\
Llama-3.2-Vision-Instruct \cite{meta2024llama3.2} &
  11B &
  128K &
  Weights &
  - &
  https://huggingface.co/meta-llama/Llama-3.2-11B-Vision-Instruct \\
Qwen2.5-Instruct \cite{yang2024qwen2.5} & 7B   & 128K & Weights & -                      & https://huggingface.co/Qwen/Qwen2.5-7B-Instruct     \\
Qwen2.5-Instruct \cite{yang2024qwen2.5} & 14B  & 128K & Weights & -                      & https://huggingface.co/Qwen/Qwen2.5-14B-Instruct    \\
Qwen2.5-Instruct \cite{yang2024qwen2.5} & 32B  & 128K & Weights & -                      & https://huggingface.co/Qwen/Qwen2.5-32B-Instruct    \\
mPlug-DocOwl2 \cite{hu2024mplug}   & 7B   & 8K   & Weights & -                      & https://huggingface.co/mPLUG/DocOwl2                \\
MiniCPM-V-2.6 \cite{yao2024minicpm}  & 8B   & 32K  & Weights & -                      & https://huggingface.co/openbmb/MiniCPM-V-2\_6       \\
MiniCPM-o-2.6 \cite{yao2024minicpm}   & 8B   & 32K  & Weights & -                      & https://huggingface.co/openbmb/MiniCPM-o-2\_6       \\
GLM-4V \cite{glm2024chatglm}  & 9B   & 8K   & Weights & -                      & https://huggingface.co/THUDM/glm-4v-9b              \\
InternLM3 \cite{cai2024internlm2} &
  8B &
  1M &
  \multicolumn{1}{l}{Weights} &
  - &
  https://huggingface.co/internlm/internlm3-8b-instruct \\
InternVL3   \cite{zhu2025internvl3}     & 8B   & 128K & Weights & -                      & https://huggingface.co/OpenGVLab/InternVL3-8B       \\
InternVL3   \cite{zhu2025internvl3}     & 38B  & 128K & Weights & -                      & https://huggingface.co/OpenGVLab/InternVL3-38B      \\
Qwen2.5-VL \cite{Qwen2.5-VL}     & 7B   & 128k & Weights & -                      & https://huggingface.co/Qwen/Qwen2.5-VL-7B-Instruct  \\
Qwen2.5-VL \cite{Qwen2.5-VL}     & 32B  & 128k & Weights & -                      & https://huggingface.co/Qwen/Qwen2.5-VL-32B-Instruct \\
Qwen3 \cite{yang2025qwen3}     & 0.6B  & 32k & Weights & -                      & https://huggingface.co/Qwen/Qwen3-0.6B \\
Qwen3 \cite{yang2025qwen3}     & 4B  & 128k & Weights & -                      & https://huggingface.co/Qwen/Qwen3-4B \\

Qwen3 \cite{yang2025qwen3}     & 8B  & 128k & Weights & -                      & https://huggingface.co/Qwen/Qwen3-8B \\

Qwen3 \cite{yang2025qwen3}     & 14B  & 128k & Weights & -                      & https://huggingface.co/Qwen/Qwen3-14B \\

Qwen3 \cite{yang2025qwen3}     & 32B  & 128k & Weights & -                      & https://huggingface.co/Qwen/Qwen3-32B \\

Qwen3-VL \cite{qwen3-vl}     & 8B  & 256k & Weights & -                      & https://huggingface.co/Qwen/Qwen3-VL-8B-Instruct \\

Qwen3-VL \cite{qwen3-vl}     & 32B  & 256k & Weights & -                      & https://huggingface.co/Qwen/Qwen3-VL-32B-Instruct \\

DeepSeek-V3.1  \cite{liu2024deepseek}  & 671B & 128K  & API     & DeepSeek-V3-2025-12-15 & https://www.deepseek.com/                           \\
DeepSeek-R1  \cite{guo2025deepseek-r1}    & 671B & 64K  & API     & DeepSeek-R1-2025-05-28 & https://www.deepseek.com/                           \\
Qwen-Max \cite{qwen2025qwen25max}   & -    & 128K & API     & Qwen-Max-2025-01-25    & https://bailian.console.aliyun.com/                 \\
Qwen3.5-Plus  \cite{qwen35blog}    & -    & 1M & API     & Qwen3.5-Plus-2026-02-15 & https://bailian.console.aliyun.com/                 \\
GPT-4o  \cite{gpt-4o}   & -    & 128K & API     & GPT-4o-2024-11-20      & https://openai.com/                                 \\
GPT-5.2  \cite{GPT5}   & -    & 400K & API     & GPT-5.2-2026-02-10      & https://openai.com/                                 \\
Claude-4.6-Opus \cite{claude_opus_4_6} &
  - &
  200K &
  API &
  Claude-Opus-4-6 &
  https://www.anthropic.com/claude/sonnet \\

Gemini-2.5-Pro \cite{comanici2025gemini} &
  - &
  1M &
  API &
  Gemini-2.5-Pro-2025-6-17 &
  https://cloud.google.com/vertex-ai/generative-ai/docs/models/gemini/2-5-pro \\
Gemini-3.1-Flash \cite{Gemini_3_1_pro} &
  - &
  1M &
  API &
  Gemini-3.1-Pro-preview &
  https://ai.google.dev/gemini-api/docs/models/gemini-3.1-pro-preview \\
  \hline
\end{tabular}
}
\end{table*}

\begin{table*}[h]
\caption{Details of all evaluated retrievers. Zoom in for better view.}
\label{table:retriever_details}
\resizebox{\textwidth}{!}{
\begin{tabular}{lccccl}
\hline
\multicolumn{1}{l}{Retriever} &
  Size &
  \multicolumn{1}{l}{Context Window} &
  Access &
  \multicolumn{1}{c}{Base Model} &
  Website \\ \hline
SBERT \cite{reimers2019sentence}&
  0.02B &
  512 &
  Weights &
  - &
  https://huggingface.co/sentence-transformers/all-MiniLM-L6-v2 \\
BGE-large-en-v1.5 \cite{bge_embedding}  & 0.3B  & 512  & Weights & -                      & https://huggingface.co/BAAI/bge-large-en-v1.5            \\
BM25 \cite{10.1561/1500000019}    & -     & -    & Program & -                      & https://github.com/dorianbrown/rank\_bm25                \\
BGE-small-zh-v1.5  \cite{bge_embedding} & 0.02B & 512  & Weights & -                      & https://huggingface.co/BAAI/bge-small-zh-v1.5            \\
Conan-embedding-v1  \cite{li2024conanembeddinggeneraltextembedding} & 0.3B  & 512  & Weights & -                      & https://huggingface.co/TencentBAC/Conan-embedding-v1     \\
NV-Embed-v2  \cite{lee2024nv}      & 7B    & 32K  & Weights & Mistral-7B-v0.1 \cite{Jiang2023Mistral7}        & https://huggingface.co/nvidia/NV-Embed-v2                \\
BGE-m3 \cite{bge-m3}           & 0.5B  & 8192 & Weights & -                      & https://huggingface.co/BAAI/bge-m3                       \\
BGE-large-zh-v1.5 \cite{bge_embedding} & 0.3B  & 512  & Weights & -                      & https://huggingface.co/BAAI/bge-large-zh-v1.5            \\
Gte-Qwen2-Instruct \cite{li2023towards} &
  7B &
  32K &
  Weights &
  Qwen2-7B \cite{Yang2024Qwen2} &
  https://huggingface.co/Alibaba-NLP/gte-Qwen2-7B-instruct \\
MiniCPM-Embedding \cite{hu2024minicpm} &
  2B &
  32K &
  \multicolumn{1}{l}{Weights} &
  MiniCPM-2B-sft-bf16 \cite{hu2024minicpm}&
  https://huggingface.co/openbmb/MiniCPM-Embedding \\
Qwen3-Embedding \cite{zhang2025qwen3_embed} &
  0.6B &
  32K &
  \multicolumn{1}{l}{Weights} &
  Qwen3-0.6B \cite{yang2025qwen3}&
  https://huggingface.co/Qwen/Qwen3-Embedding-0.6B \\

Qwen3-Embedding \cite{zhang2025qwen3_embed} &
  4B &
  128K &
  \multicolumn{1}{l}{Weights} &
  Qwen3-4B \cite{yang2025qwen3}&
  https://huggingface.co/Qwen/Qwen3-Embedding-4B \\

Qwen3-Embedding \cite{zhang2025qwen3_embed} &
  8B &
  128K &
  \multicolumn{1}{l}{Weights} &
  Qwen3-8B \cite{yang2025qwen3}&
  https://huggingface.co/Qwen/Qwen3-Embedding-8B \\

SigLip \cite{zhai2023siglip}   & 0.8B  & 64   & Weights & -                      & https://huggingface.co/google/siglip-so400m-patch14-384  \\
CLIP  \cite{radford2021clip} & 0.4B  & 77   & Weights & -                      & https://huggingface.co/openai/clip-vit-large-patch14-336 \\
ColSmolVLM-v0.1 \cite{faysse2024colpali}   & 2B    & 16K  & Weights & SmolVLM-Instruct \cite{marafioti2025smolvlm}      & https://huggingface.co/vidore/colsmolvlm-v0.1            \\
ColPali-v1.2-merged \cite{faysse2024colpali} &
  3B &
  8192 &
  Weights &
  Paligemma-3B-pt-448 \cite{beyer2024paligemma} &
  https://huggingface.co/vidore/colpali-v1.2-merged \\
VisRAG-Ret \cite{yu2024visrag} & 3B    & 4096 & Weights & MiniCPM-3B-V2.0 \cite{yao2024minicpm}       & https://huggingface.co/openbmb/VisRAG-Ret                \\
Gme-Qwen2-VL-Instruct \cite{zhang2024gme} &
  7B &
  32K &
  Weights &
  Qwen2-VL-7B-Instruct &
  https://huggingface.co/Alibaba-NLP/gme-Qwen2-VL-2B-Instruct \\
ColQwen2 \cite{faysse2024colpali} & 2B    & 32K  & Weights & Qwen2-VL-2B-Instruct \cite{wang2024qwen2}  & https://huggingface.co/vidore/colqwen2-v1.0-merged       \\
ColQwen2.5 \cite{faysse2024colpali} & 3B    & 128K & Weights & Qwen2.5-VL-3B-Instruct \cite{Qwen2.5-VL} & https://huggingface.co/vidore/colqwen2.5-v0.2            \\
Qwen3-VL-Embedding \cite{qwen3-vl-embedding} & 2B    & 256K & Weights & Qwen3-VL-2B-Instruct \cite{qwen3-vl} & https://huggingface.co/Qwen/Qwen3-VL-Embedding-2B            \\
Qwen3-VL-Embedding \cite{qwen3-vl-embedding} & 8B    & 256K & Weights & Qwen3-VL-2B-Instruct \cite{qwen3-vl} & https://huggingface.co/Qwen/Qwen3-VL-Embedding-8B            \\
Jina-Embeddings-V4 \cite{jina-embedding-v4} & 3B    & 128K & Weights & Qwen2.5-VL-3B-Instruct \cite{Qwen2.5-VL} & https://huggingface.co/jinaai/jina-embeddings-v4            \\ 
ColNomic-Embed-Multimodal \cite{colnomic} & 3B    & 128K & Weights & Qwen2.5-VL-3B-Instruct \cite{Qwen2.5-VL} & https://huggingface.co/nomic-ai/colnomic-embed-multimodal-3b            \\ \hline
\end{tabular}
}
\end{table*}

\subsection{Prompt for Response Generation}
\begin{figure*}[]
  \centering
  \includegraphics[width=0.9\linewidth]{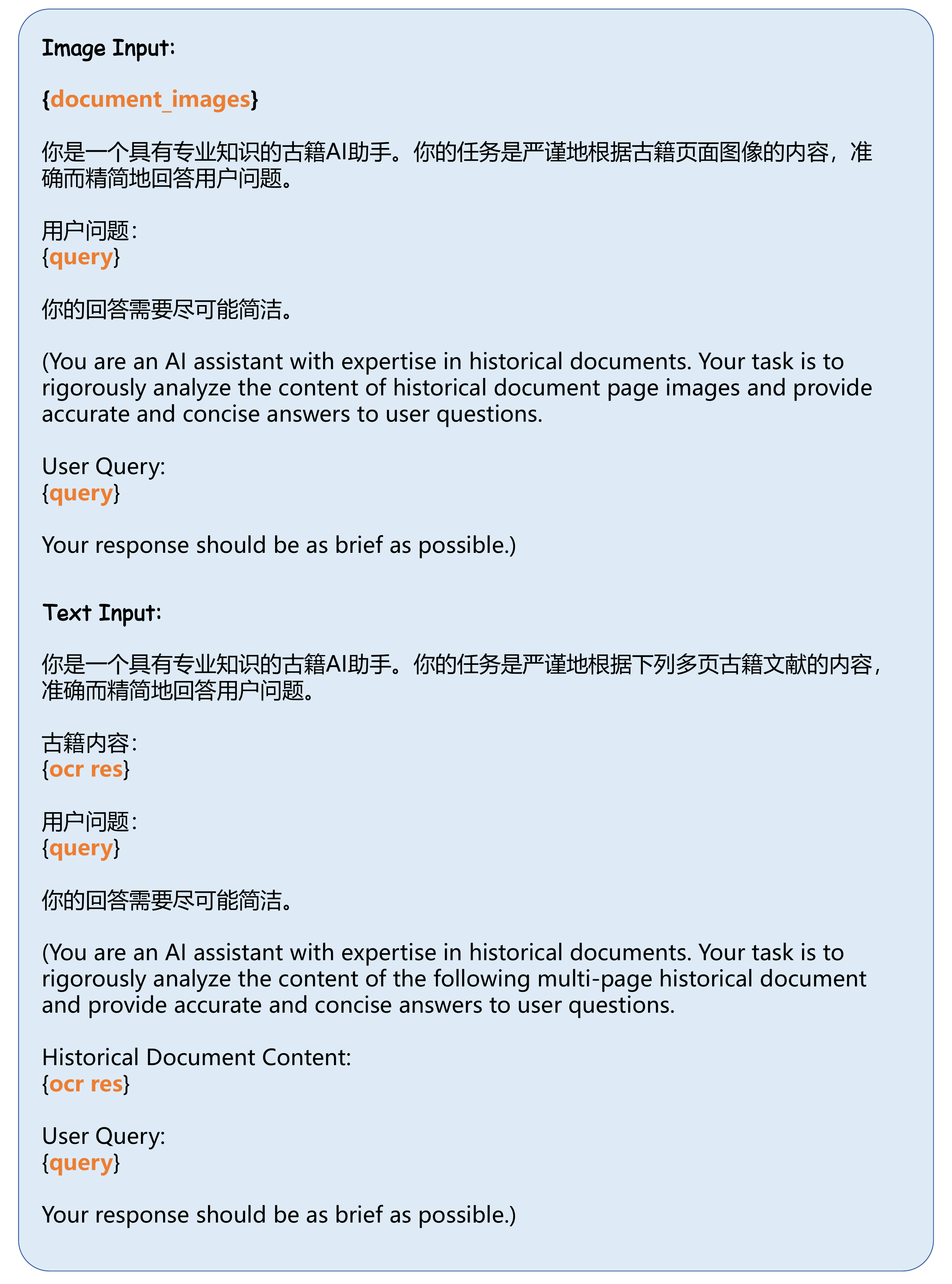}
  \caption{Prompt for response generation.}
  \label{fig:generate_prompt}
\end{figure*}

The prompts we designed for image-input (based on the page screenshots) and text-input (based on the OCR results) are shown in the Figure \ref{fig:generate_prompt}.

\subsection{Prompt for Answer Extraction}
\begin{figure*}[]
  \centering
  \includegraphics[width=1.0\linewidth]{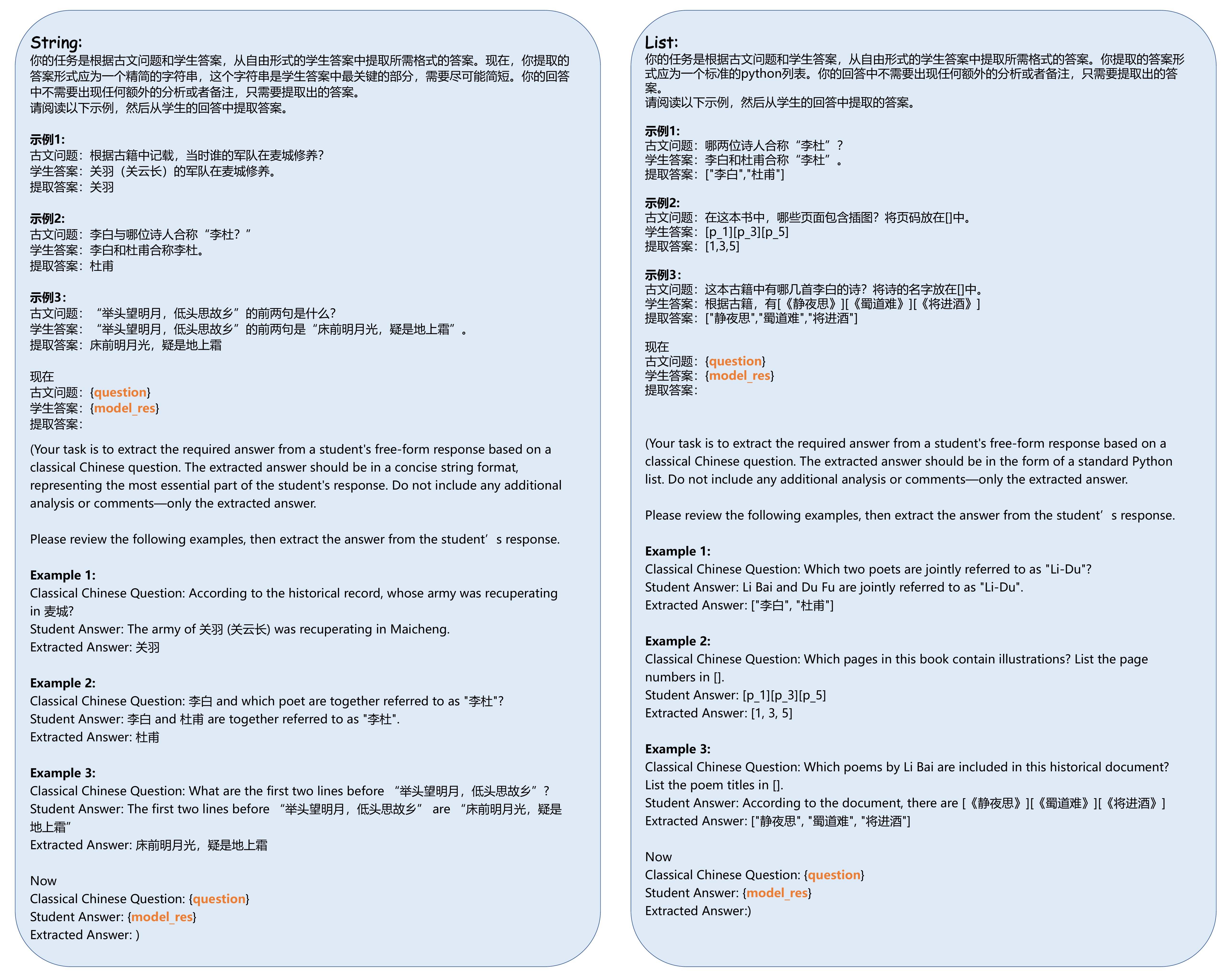}
  \caption{Prompt for answer extraction (Format: String and List). Zoom in for better view.}
  \label{fig:extract_prompt_1}
\end{figure*}

\begin{figure*}[]
  \centering
  \includegraphics[width=1.0\linewidth]{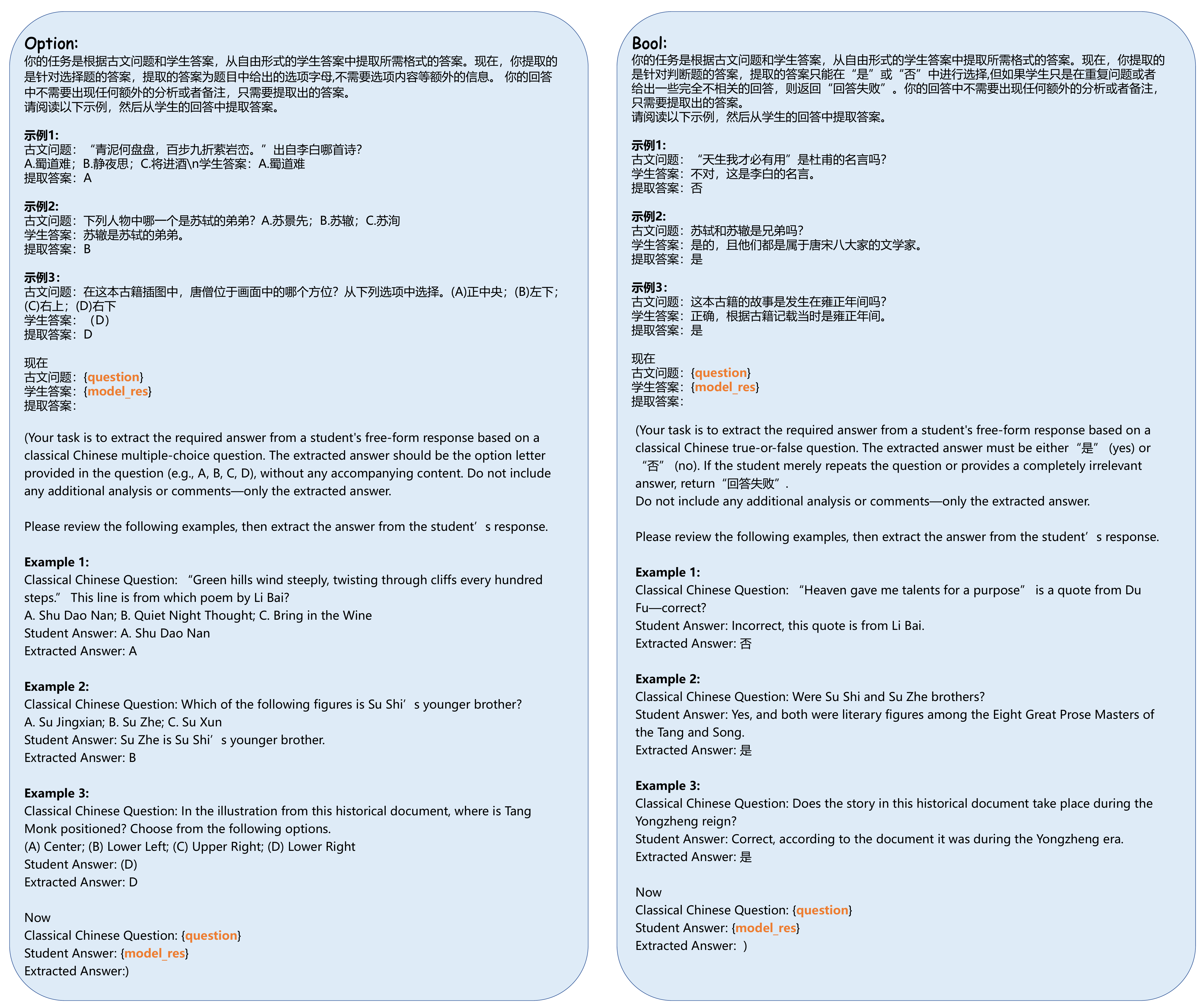}
  \caption{Prompt for answer extraction (Format: Option and Bool). Zoom in for better view.}
  \label{fig:extract_prompt_2}
\end{figure*}

\begin{figure*}[]
  \centering
  \includegraphics[width=1.0\linewidth]{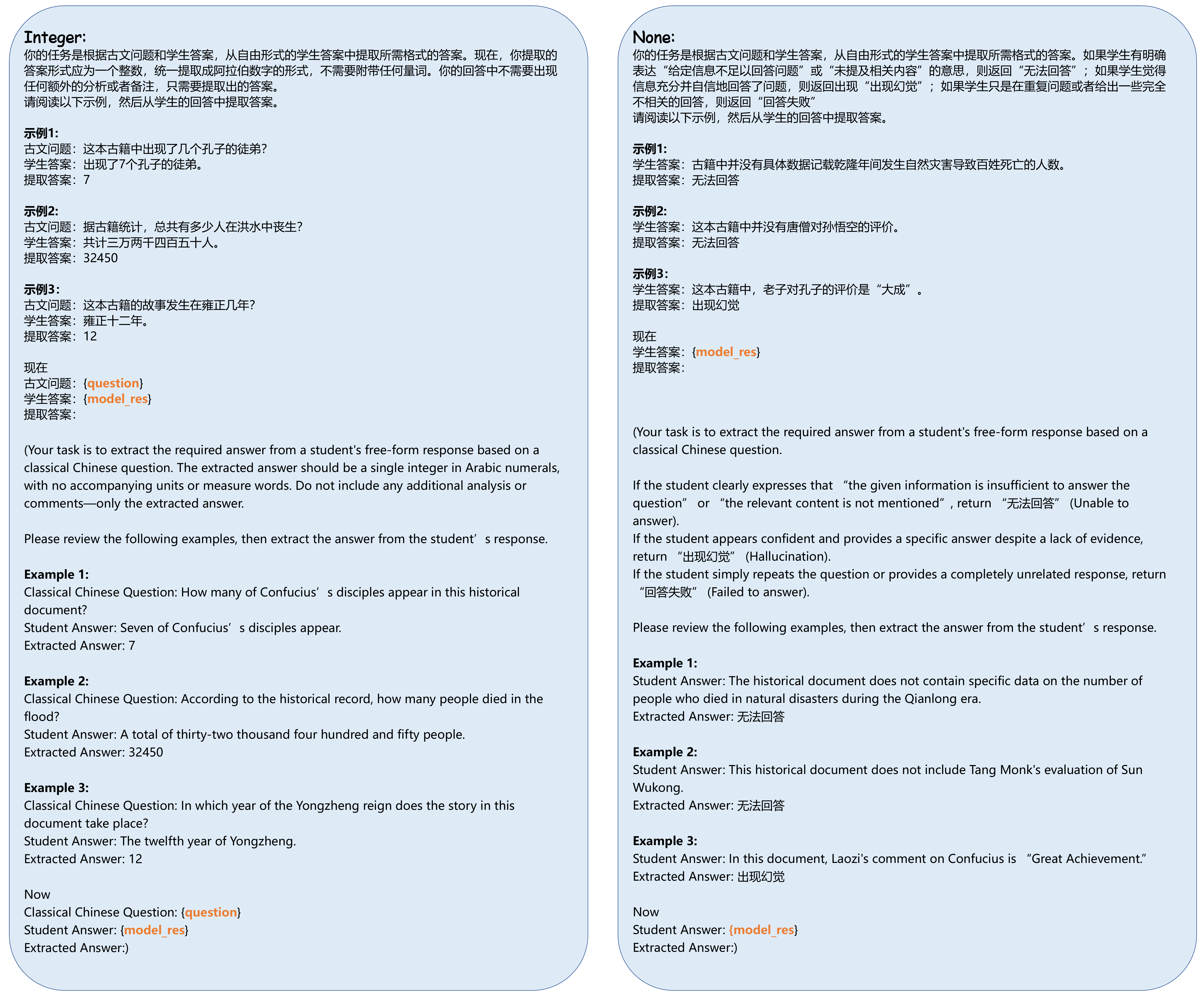}
  \caption{Prompt for answer extraction (Format: Integer and None). Zoom in for better view.}
  \label{fig:extract_prompt_3}
\end{figure*}

We design specific answer extraction (GPT-4o \cite{gpt-4o}) prompts for each of the six answer formats, as shown in Figure \ref{fig:extract_prompt_1} 
 \ref{fig:extract_prompt_2}  \ref{fig:extract_prompt_3}.

\begin{algorithm}[]
\caption{Compute Score Between Prediction and GT Lists}
\label{algorithm 1}
\raggedright 
\textbf{Input:} \\
\textit{pred\_list}: the list-format answer predicted by the model; \\ \textit{gt\_list}: the ground truth list. \\
\textbf{Output:} \\
\textit{final\_score}: generalized accuracy score (ranging from 0 to 1).
\begin{algorithmic}[1]
\State $\text{pred\_list} \gets \text{Sort}(\text{pred\_list})$
\State $\text{gt\_list} \gets \text{Sort}(\text{gt\_list})$
\State $\text{scores\_list} \gets [\ ]$
\ForAll{$\text{gt} \in \text{gt\_list}$}
    \State $\text{max\_score} \gets 0$
    \ForAll{$\text{pred} \in \text{pred\_list}$}
        \If{$\text{TypeOf(gt)} = \text{Int}$}
            \State $\text{score\_val} \gets \textbf{ExactMatch}(\text{pred}, \text{gt})$ 
        \Else
            \State $\text{score\_val} \gets \textbf{ANLS}(\text{pred}, \text{gt})$
        \EndIf
        \If{$\text{score\_val} > \text{max\_score}$}
            \State $\text{max\_score} \gets \text{score\_val}$
        \EndIf
    \EndFor
    \State Append $\text{max\_score}$ to $\text{scores\_list}$
\EndFor
\State $\text{avg\_score} \gets \frac{\sum \text{scores\_list}}{|\text{gt\_list}|}$
\State $\text{length\_penalty} \gets \min(1, \frac{|\text{gt\_list}|}{|\text{pred\_list}|})^{0.5}$
\State $\text{final\_score} \gets \text{avg\_score} \times \text{length\_penalty}$
\end{algorithmic}
\end{algorithm}

\subsection{Rules for Score Calculation}
Following MMLongBench-Doc \cite{ma2024mmlongbench} and LongDocURL \cite{deng2024longdocurl}, we evaluate the model's responses by scoring the extracted answers against the reference answers. The scoring process is rule-based, with different strategies applied according to the format of the reference answers. We detail its rules as follow:

\textbf{String:} We compute the Average Normalized Levenshtein Similarity (ANLS) using a predefined threshold (\(\tau = 0.5\)).

\textbf{Integer:} We perform strict exact matching and assign a score of either 0 or 1 to each answer. For the reference answers, we provide both Chinese numerals and Arabic numerals, and the higher of the two matching scores is taken as the final score.

\textbf{Option:} We perform strict exact matching with the answer options: a successful match receives a score of 1, otherwise 0.

\textbf{Bool:} We perform exact matching between the extracted judgment result and the ground truth (GT). A match is awarded a score of 1; otherwise, 0.

\textbf{List:} Following the approach of LongDocURL \cite{deng2024longdocurl}, we adopt the same scoring mechanism for list-format answers. Compared to MMLongBench \cite{ma2024mmlongbench}, our method is more lenient, as reflected in the following two aspects:
(1) When the number of predicted items differs from that of the reference answer, we apply a length-based penalty instead of assigning a score of zero directly;
(2) For each element in the reference list, the score is determined by the highest similarity obtained from comparisons with all elements in the predicted list.
A detailed description of the procedure is shown in Algorithm~\ref{algorithm 1}.

\textbf{None:} If the model explicitly indicates that the question is not answerable, it receives a score of 1; otherwise, 0.

\subsection{Selection Rules of Images Input}
\begin{figure*}[h]
  \centering
  \includegraphics[width=1.0\linewidth]{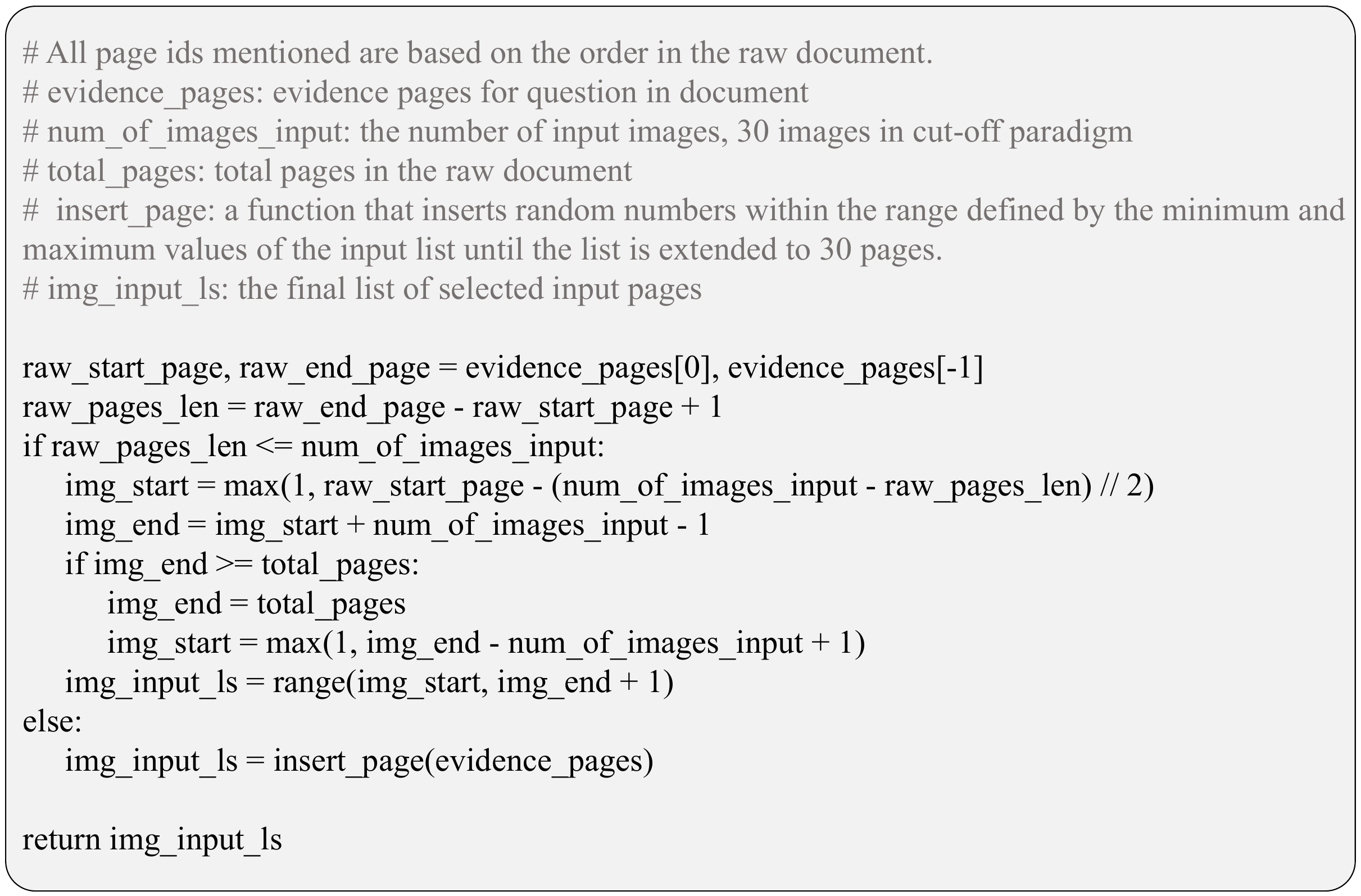}
  \caption{Selection rules of images input.}
  \label{fig:select rule}
\end{figure*}

We follow the cut-off strategy used in LongDocURL \cite{deng2024longdocurl}, selecting 30 page images that include the evidence page(s) as input. The pseudocode of our selection method is shown in Figure \ref{fig:select rule}.


\subsection{Impact of LLM-based Answer Extractor}
To verify the reliability of using GPT-4o as an answer extractor, we invited human experts to manually extract answers from the raw outputs of four representative models. We then calculated the scores based on these human-extracted answers and compared them with the scores obtained using GPT-4o-based extraction. Furthermore, we repeated the GPT-4o-based answer extraction process five times for each model and calculated the variance of the final scores. This is conducted to demonstrate the stability of using GPT-4o as an automated extractor. The results are summarized in Table \ref{tab:LLM_extract}.

\subsection{{Subgroup-level Uncertainty Analysis}}
\label{Apendix C.8}
{We conduct a subgroup-level uncertainty analysis to assess the reliability of the fine-grained evaluation results on LongHisDoc. Since the benchmark contains 101 documents and 1,012 QA pairs, some subgroups are relatively small, such as visual-evidence QA, Boolean answers, and enum/option-format answers. To avoid over-interpreting subgroup-level point estimates, we report 95\% bootstrap confidence intervals for each subgroup across four representative evaluated models.
For each subgroup, we perform \(N=1{,}000\) bootstrap resampling iterations over QA instances. In each iteration, QA instances are sampled with replacement from the corresponding subgroup, and the average score is recomputed. The lower and upper bounds of the 95\% confidence interval are given by the 2.5th and 97.5th percentiles of the resulting bootstrap distribution.
Table~\ref{tab:subgroup_uncertainty_multi_model} summarizes the subgroup size, average score, and 95\% bootstrap confidence interval across evidence modality, task type, answer type, and document category. The results show that larger subgroups generally have narrower confidence intervals, while smaller subgroups, such as visual-evidence QA, Boolean answers, and enum answers, tend to have wider intervals. We therefore interpret subgroup-level results as diagnostic evidence and treat comparisons on small subgroups with caution.}


\subsection{More Experimental Results}
We present more evaluation results of models on LongHisDoc in Table \ref{tab:table7}.



\begin{table}[h]
\centering
\begin{tabular}{l|c c c}
\hline
Model &LLM Extract(\%)  & Human Extract(\%) & LLM Extract Var(\%)  \\
\hline
GPT-5.2 & 41.0 & 41.3(+0.3) & 0.04\\
Gemini-3.1-Pro & 62.9 & 63.1(+0.2) & 0.09\\
Claude-4.6-Opus & 46.0 & 46.1(+0.1) & 0.13\\
Qwen3.5-Plus & 54.1 & 53.9(-0.2) & 0.14\\

\hline
\end{tabular}
\caption{\label{tab:LLM_extract}Reliability Validation of the Automated Answer Extractor.}
\end{table}

\section{Error Analysis}
\label{appendix-D}
\subsection{Impact of OCR Noise}
In the text-input experimental setting, we use the Optical Character Recognition (OCR) annotations provided by HisDoc-1B as model input. HisDoc-1B adopts a semi-automatic annotation pipeline, which includes using YOLOv7 \cite{wang2023yolov7} as the text detector, employing a Vision Transformer (ViT) \cite{dosovitskiy2020image} for character recognition, applying heuristic rules for reading order prediction, and utilizing a Transformer \cite{vaswani2017attention} architecture model trained on classical Chinese corpora \cite{daizhige} to generate punctuation.

\begin{figure*}[h]
  \centering
  \includegraphics[width=1.0\linewidth]{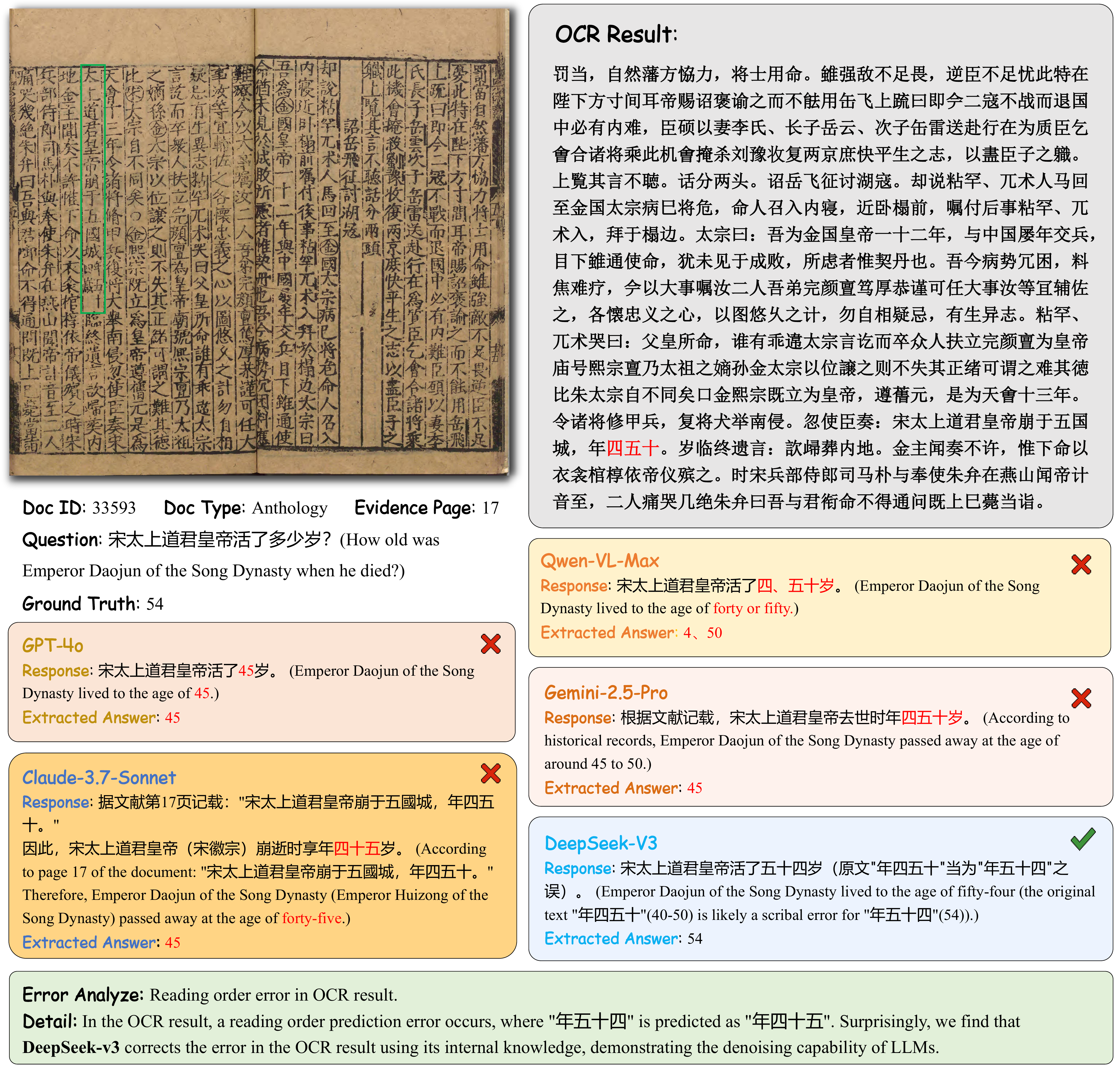}
  \caption{Text-input case 1. Evidence source: "Layout". Answer format: "List". Key information in the image is highlighted with green boxes, while incorrect information is marked in red. Zoom in for better view.}
  \label{fig:case_ocr_2}
\end{figure*}

Although the OCR technology adopted by HisDoc-1B is relatively advanced for historical documents, the generated OCR results inevitably contain a certain degree of noise. We consider this noise to be one of the key factors limiting model performance. Specifically, OCR noise can be categorized into two types: (1) Errors introduced by the OCR pipeline in text detection, recognition, reading order, and punctuation, which lead to noisy inputs that interfere with the model's understanding. (2) Inability of the OCR engine to interpret visual elements in historical documents, such as illustrations, seals, font styles, and physical damage. These visual elements also disrupt the OCR results of text on the same page. Figure \ref{fig:case_ocr_2}\textasciitilde\ref{fig:case_ocr_4} present the responses of different models, highlighting the impact of OCR noise on model performance. We also provide brief error analyses for each case.

\subsection{Challenges for LVLMs}
The experimental results in Section 4 indicate that historical document understanding remains a significant challenge for current LVLMs. Figures \ref{fig:case_img_2} \textasciitilde \ref{fig:case_img_4} present several specific examples to further illustrate this point, along with error analyses for each case.

\section{Retriever Evaluation}
\label{appendix-E}
In Section 4.6, we conduct extensive experiments on both text-based and visual-based retrievers to evaluate their effectiveness in the context of historical documents. To further investigate the performance of the retrievers and more clearly illustrate the differences between them, we carry out experiments under additional top-k settings. The results are shown in Figure \ref{fig:retriever}, with recall rate as the evaluation metric.

\section{Limitation}
\label{appendix-F}
The main limitation of our work lies in the scale of the benchmark. Due to the complexity of historical documents, annotators must possess extensive knowledge of classical Chinese, and the annotation process is both time-consuming and labor-intensive. On average, it takes over an hour for an expert-level annotator to annotate a \textbf{single} historical document. Therefore, as the first benchmark for long-context historical document understanding, we prioritize annotation quality over quantity at this stage. To transparently characterize the potential impact of small-sample subsets, we additionally provide a subgroup-level uncertainty analysis in Appendix~\ref{Apendix C.8}

\section{Statement of Responsibility}
\label{appendix-G}
The LongHisDoc benchmark is released under the CC-BY-NC-ND-4.0 license. The data source of LongHisDoc is HisDoc-1B \cite{shi2025large} and the entire data-construction process is conducted under the supervision of the HisDoc-1B authors, in full compliance with their data-use protocols and with their explicit permission. We have reviewed the ethical guidelines and ensured that the content of the paper and the benchmark are in compliance with the guidelines. Meanwhile, all data annotators are fully informed of the purpose and intended use of our dataset.

\section{Broader Impact}
\label{appendix-H}
We believe that the LongHisDoc benchmark will serve as a valuable resource to advance research in long-context historical document understanding, foster deeper engagement with cultural heritage and drive the development of AI systems for interpreting the rich tapestry of human history. Therefore, our work carries a positive social impact.

\begin{table*}[t]
\centering
\small
\setlength{\tabcolsep}{3.5pt}

\caption{Subgroup uncertainty estimates (95\% bootstrap CI, $N=1,000$) across evaluated frontier models on \textsc{LongHisDoc}.}
\label{tab:subgroup_uncertainty_multi_model}

\resizebox{\textwidth}{!}{%
\begin{tabular}{l c cc cc cc cc}
\toprule
& & \multicolumn{2}{c}{\textbf{Gemini-3.1-Pro+IMG}} & \multicolumn{2}{c}{\textbf{Qwen3-VL-32B+IMG}} & \multicolumn{2}{c}{\textbf{Gemini-3.1-Pro+OCR }} & \multicolumn{2}{c}{\textbf{DeepSeek-v3.2+OCR}} \\
\cmidrule(lr){3-4} \cmidrule(lr){5-6} \cmidrule(lr){7-8} \cmidrule(lr){9-10}
\textbf{Subgroup Category} & \textbf{Size ($n$)} & \textbf{Score} & \textbf{95\%CI} & \textbf{Score} & \textbf{95\%CI} & \textbf{Score} & \textbf{95\%CI} & \textbf{Score} & \textbf{95\%CI} \\
\midrule
\multicolumn{10}{l}{\textit{Task Type}} \\
\quad Single Page & 546 & 66.9\% & [63.2\%, 70.5\%] & 53.9\% & [49.9\%, 57.6\%] & 73.7\% & [70.2\%, 77.2\%] & 68.7\% & [65.0\%, 72.2\%] \\
\quad Cross Pages & 196 & 61.5\% & [55.5\%, 67.7\%] & 36.2\% & [30.0\%, 42.4\%] & 73.0\% & [67.3\%, 78.3\%] & 64.7\% & [58.4\%, 70.6\%] \\
\quad Multi-Hop & 172 & 49.3\% & [42.0\%, 56.1\%] & 44.3\% & [37.4\%, 51.4\%] & 63.0\% & [56.0\%, 69.6\%] & 46.3\% & [38.9\%, 53.3\%] \\
\quad Unanswerable & 98 & 67.4\% & [58.2\%, 76.5\%] & 40.8\% & [30.6\%, 51.0\%] & 54.1\% & [43.9\%, 64.3\%] & 40.8\% & [30.6\%, 51.0\%] \\
\midrule
\multicolumn{10}{l}{\textit{Answer Type}} \\
\quad Str & 569 & 60.8\% & [57.2\%, 64.6\%] & 42.5\% & [39.0\%, 46.5\%] & 69.3\% & [65.9\%, 72.7\%] & 62.1\% & [58.5\%, 66.0\%] \\
\quad Int & 120 & 56.7\% & [47.5\%, 65.8\%] & 55.0\% & [45.8\%, 64.2\%] & 69.2\% & [60.8\%, 76.7\%] & 56.7\% & [48.3\%, 65.8\%] \\
\quad Enum & 36 & 75.0\% & [61.1\%, 88.9\%] & 61.1\% & [44.4\%, 77.8\%] & 80.6\% & [66.7\%, 91.7\%] & 77.8\% & [63.9\%, 91.7\%] \\
\quad List & 132 & 61.6\% & [54.6\%, 67.6\%] & 50.2\% & [43.8\%, 56.9\%] & 72.6\% & [66.0\%, 77.9\%] & 65.1\% & [58.6\%, 71.4\%] \\
\quad Bool & 57 & 84.2\% & [73.7\%, 93.0\%] & 79.0\% & [66.7\%, 89.5\%] & 91.2\% & [82.5\%, 98.2\%] & 80.7\% & [70.2\%, 89.5\%] \\
\quad None & 98 & 67.4\% & [58.2\%, 76.5\%] & 40.8\% & [30.6\%, 51.0\%] & 54.1\% & [43.9\%, 64.3\%] & 40.8\% & [30.6\%, 51.0\%] \\
\midrule
\multicolumn{10}{l}{\textit{Evidence Modality}} \\
\quad Text & 455 & 66.0\% & [61.9\%, 70.1\%] & 47.4\% & [43.1\%, 51.4\%] & 76.3\% & [72.6\%, 79.5\%] & 69.5\% & [65.4\%, 73.4\%] \\
\quad Layout & 362 & 60.2\% & [55.7\%, 64.8\%] & 49.9\% & [45.1\%, 54.7\%] & 73.9\% & [69.4\%, 78.0\%] & 64.7\% & [59.9\%, 69.5\%] \\
\quad Visual & 97 & 53.6\% & [43.6\%, 63.1\%] & 46.6\% & [37.4\%, 56.3\%] & 40.5\% & [31.6\%, 50.1\%] & 32.2\% & [23.3\%, 41.7\%] \\
\midrule
\multicolumn{10}{l}{\textit{Document Category}} \\
\quad History & 111 & 72.6\% & [64.3\%, 79.7\%] & 49.7\% & [42.1\%, 58.4\%] & 83.3\% & [77.1\%, 88.9\%] & 64.1\% & [56.2\%, 72.3\%] \\
\quad Yizang & 67 & 42.2\% & [31.0\%, 53.1\%] & 44.3\% & [33.4\%, 55.8\%] & 52.4\% & [41.7\%, 63.6\%] & 45.6\% & [33.8\%, 57.6\%] \\
\quad Literature & 113 & 59.5\% & [51.4\%, 67.9\%] & 46.7\% & [38.5\%, 54.8\%] & 59.0\% & [50.8\%, 67.7\%] & 52.2\% & [43.7\%, 60.5\%] \\
\quad Anthology & 129 & 62.5\% & [54.7\%, 70.1\%] & 46.2\% & [38.6\%, 53.6\%] & 67.7\% & [60.5\%, 74.9\%] & 68.8\% & [61.6\%, 75.8\%] \\
\quad Taoism & 82 & 61.8\% & [51.5\%, 71.3\%] & 49.2\% & [39.4\%, 59.7\%] & 68.8\% & [58.4\%, 77.8\%] & 56.9\% & [46.1\%, 66.2\%] \\
\quad Buddhism & 98 & 66.3\% & [58.0\%, 74.8\%] & 41.2\% & [32.2\%, 50.3\%] & 70.4\% & [62.4\%, 78.8\%] & 68.3\% & [59.9\%, 76.9\%] \\
\quad Medicine & 134 & 71.7\% & [64.7\%, 78.6\%] & 54.4\% & [46.4\%, 63.1\%] & 76.2\% & [69.3\%, 82.9\%] & 64.4\% & [56.9\%, 72.1\%] \\
\quad Confucianism & 68 & 57.9\% & [47.3\%, 68.9\%] & 51.8\% & [40.4\%, 62.7\%] & 71.2\% & [61.4\%, 81.3\%] & 61.1\% & [50.1\%, 71.7\%] \\
\quad Poetry & 114 & 60.8\% & [51.5\%, 69.9\%] & 47.5\% & [39.0\%, 56.0\%] & 75.8\% & [68.4\%, 82.2\%] & 65.8\% & [57.8\%, 73.3\%] \\
\quad Philosophy & 96 & 61.7\% & [52.2\%, 70.0\%] & 42.6\% & [33.9\%, 52.2\%] & 65.5\% & [55.9\%, 74.3\%] & 57.8\% & [48.3\%, 67.0\%] \\
\bottomrule

\end{tabular}
}
\end{table*}

{\footnotesize
\begin{table*}[h]
\caption{More model evaluation results.}
\label{tab:table7}
\vspace{2pt}
\resizebox{\textwidth}{!}{ 
\begin{tabular}{lccccccccccccccc}

\hline

\multicolumn{1}{c}{\multirow{2}{*}{\textbf{Models}}} &
  \multicolumn{1}{c|}{\multirow{2}{*}{\textbf{Size}}} &
  \multicolumn{4}{c|}{\textbf{Evidence \ Page}} &
  \multicolumn{6}{c|}{\textbf{Answer \ Format}} &
  \multicolumn{3}{c|}{\textbf{Evidence \ Source}} &
  \multicolumn{1}{c}{\multirow{2}{*}{\textbf{AVG}}} \\ \cline{3-15}
\multicolumn{1}{c}{} &
  \multicolumn{1}{c|}{} &
  SIN  &
  CRO  &
  MUL  &
  \multicolumn{1}{c|}{UNA} &
  STR\   &
  INT  &
  OPT  &
  LIST &
  BOOL &
  \multicolumn{1}{c|}{NONE} &
  TXT &
  LAY &
  \multicolumn{1}{c|}{VIS} &
  \multicolumn{1}{c}{} \\ \hline
\multicolumn{16}{c}{\textit{OCR Results + Large Language Models (LLMs)}} \\ \hline
Qwen3 \cite{yang2025qwen3}&
  \multicolumn{1}{c|}{0.6B} &
  21.2 &
  13.8 &
  18.5 &
  \multicolumn{1}{c|}{26.5} &
  15.7 &
  17.7 &
  47.2 &
  14.4 &
  49.1 &
  \multicolumn{1}{c|}{26.5} &
  20.3 &
  18.5 &
  \multicolumn{1}{c|}{15.8} &
  \multicolumn{1}{c}{19.8} \\
  
Qwen3 (Think) \cite{yang2025qwen3}&
  \multicolumn{1}{c|}{0.6B} &
  22.8 &
  13.3 &
  20.7 &
  \multicolumn{1}{c|}{22.5} &
  17.4 &
  16.8 &
  50.0 &
  13.7 &
  54.4 &
  \multicolumn{1}{c|}{22.5} &
  21.8 &
  19.1 &
  \multicolumn{1}{c|}{18.5} &
  \multicolumn{1}{c}{20.6} \\

Qwen3 \cite{yang2025qwen3}&
  \multicolumn{1}{c|}{1.7B} &
  36.4 &
  25.1 &
  23.2 &
  \multicolumn{1}{c|}{11.2} &
  28.9 &
  29.4 &
  44.4 &
  26.6 &
  64.9 &
  \multicolumn{1}{c|}{11.2} &
  36.3 &
  27.8 &
  \multicolumn{1}{c|}{22.7} &
  \multicolumn{1}{c}{29.5} \\
  
Qwen3 (Think) \cite{yang2025qwen3}&
  \multicolumn{1}{c|}{1.7B} &
  39.1 &
  26.3 &
  26.2 &
  \multicolumn{1}{c|}{28.6} &
  30.1 &
  29.4 &
  61.1 &
  32.2 &
  68.4 &
  \multicolumn{1}{c|}{28.6} &
  39.5 &
  30.0 &
  \multicolumn{1}{c|}{22.9} &
  \multicolumn{1}{c}{33.4} \\

Qwen2.5-VL-Instruct \cite{Qwen2.5-VL}&
  \multicolumn{1}{c|}{7B} &
  45.3 &
  31.5 &
  21.3 &
  \multicolumn{1}{c|}{33.7} &
  38.3 &
  26.3 &
  52.8 &
  30.0 &
  64.9 &
  \multicolumn{1}{c|}{33.7} &
  43.5 &
  34.9 &
  \multicolumn{1}{c|}{22.2} &
  \multicolumn{1}{c}{37.4} \\
Qwen3 \cite{yang2025qwen3}&
  \multicolumn{1}{c|}{4B} &
  50.9 &
  40.2 &
  28.4 &
  \multicolumn{1}{c|}{30.6} &
  43.1 &
  42.0 &
  66.7 &
  32.4 &
  75.4 &
  \multicolumn{1}{c|}{30.6} &
  50.8 &
  41.0 &
  \multicolumn{1}{c|}{26.7} &
  \multicolumn{1}{c}{43.0} \\
  
Qwen3 (Think) \cite{yang2025qwen3}&
  \multicolumn{1}{c|}{4B} &
  50.8 &
  40.8 &
  33.3 &
  \multicolumn{1}{c|}{33.7} &
  43.4 &
  38.7 &
  72.2 &
  37.0 &
  80.7 &
  \multicolumn{1}{c|}{33.7} &
  53.3 &
  41.3 &
  \multicolumn{1}{c|}{23.5} &
  \multicolumn{1}{c}{44.2} \\

Qwen2.5-Instruct \cite{yang2024qwen2.5}&
  \multicolumn{1}{c|}{7B} &
  47.5 &
  36.1 &
  31.1 &
  \multicolumn{1}{c|}{42.9} &
  41.8 &
  37.3 &
  47.2 &
  32.3 &
  71.9 &
  \multicolumn{1}{c|}{42.9} &
  48.5 &
  39.7 &
  \multicolumn{1}{c|}{19.6} &
  \multicolumn{1}{c}{44.4} \\

Qwen3 \cite{yang2025qwen3}&
  \multicolumn{1}{c|}{8B} &
  52.3 &
  41.8 &
  37.3 &
  \multicolumn{1}{c|}{36.7} &
  44.5 &
  40.3 &
  77.8 &
  43.3 &
  79.0 &
  \multicolumn{1}{c|}{36.7} &
  53.7 &
  44.1 &
  \multicolumn{1}{c|}{28.6} &
  \multicolumn{1}{c}{46.2} \\
  
Qwen3 (Think) \cite{yang2025qwen3}&
  \multicolumn{1}{c|}{8B} &
  52.6 &
  47.5 &
  39.5 &
  \multicolumn{1}{c|}{39.8} &
  46.0 &
  42.9 &
  75.0 &
  44.8 &
  86.0 &
  \multicolumn{1}{c|}{39.8} &
  56.6 &
  46.0 &
  \multicolumn{1}{c|}{24.9} &
  \multicolumn{1}{c}{48.2} \\
Qwen2.5-Instruct \cite{yang2024qwen2.5}&
  \multicolumn{1}{c|}{14B} &
  54.3 &
  49.0 &
  37.7 &
  \multicolumn{1}{c|}{50.0} &
  48.3 &
  39.8 &
  75.0 &
  45.7 &
  82.5 &
  \multicolumn{1}{c|}{50.0} &
  57.3 &
  47.8 &
  \multicolumn{1}{c|}{24.4} &
  \multicolumn{1}{c}{50.0} \\

Qwen2.5-Instruct \cite{yang2024qwen2.5}&
  \multicolumn{1}{c|}{32B} &
  59.8 &
  51.3 &
  41.5 &
  \multicolumn{1}{c|}{56.1} &
  52.6 &
  50.9 &
  69.4 &
  52.4 &
  77.2 &
  \multicolumn{1}{c|}{56.1} &
  63.3 &
  51.8 &
  \multicolumn{1}{c|}{24.0} &
  \multicolumn{1}{c}{54.7} \\
  \hline

\end{tabular}
}
\end{table*}
}

\begin{figure*}[h]
  \centering
  \includegraphics[width=1.0\linewidth]{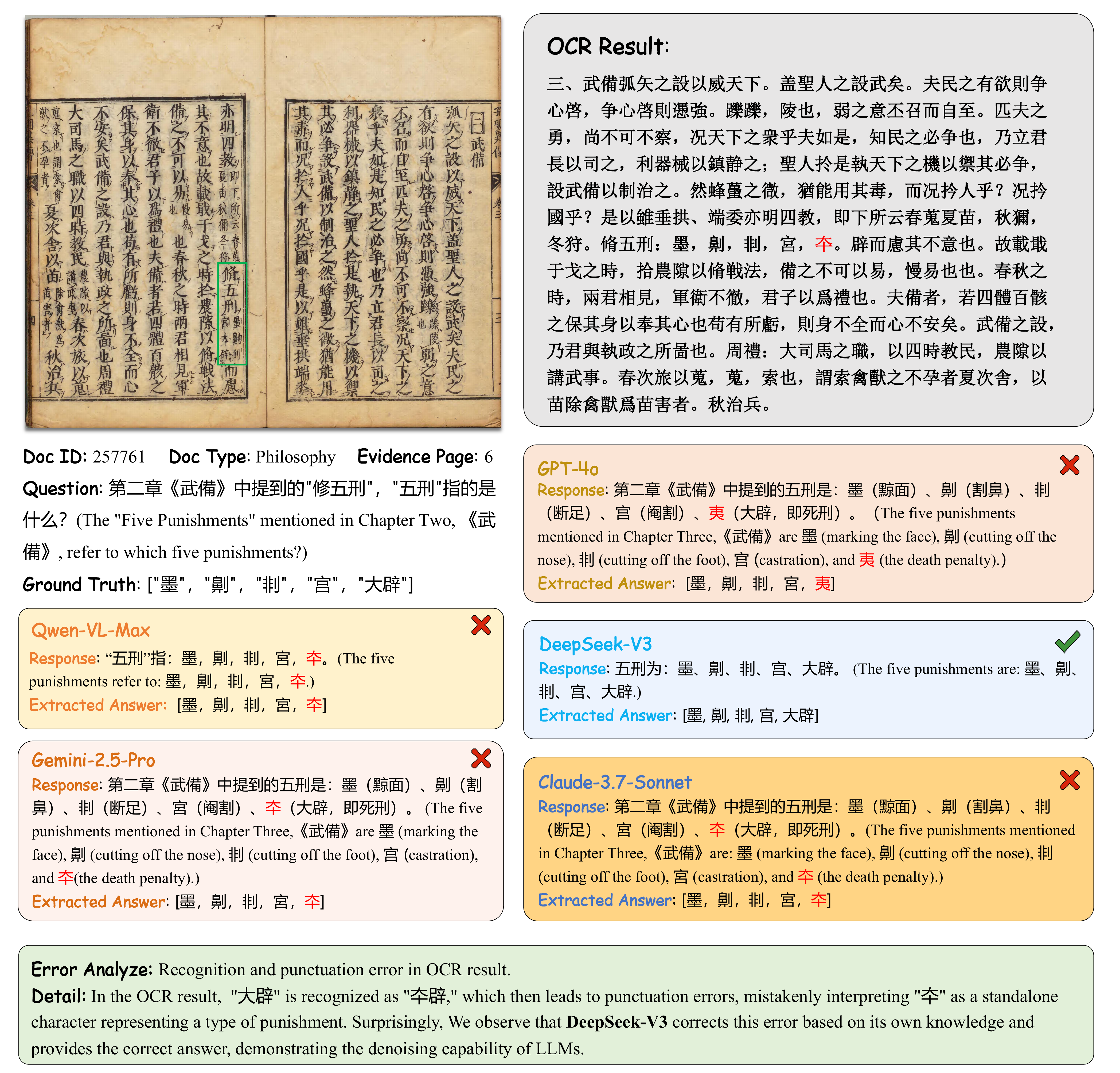}
  \caption{Text-input case 2. Evidence source: "Visual". Answer format: "String". Key information in the image is highlighted with green boxes, while incorrect information is marked in red. Zoom in for better view.}
  \label{fig:case_ocr_3}
\end{figure*}

\begin{figure*}[h]
  \centering
  \includegraphics[width=1.0\linewidth]{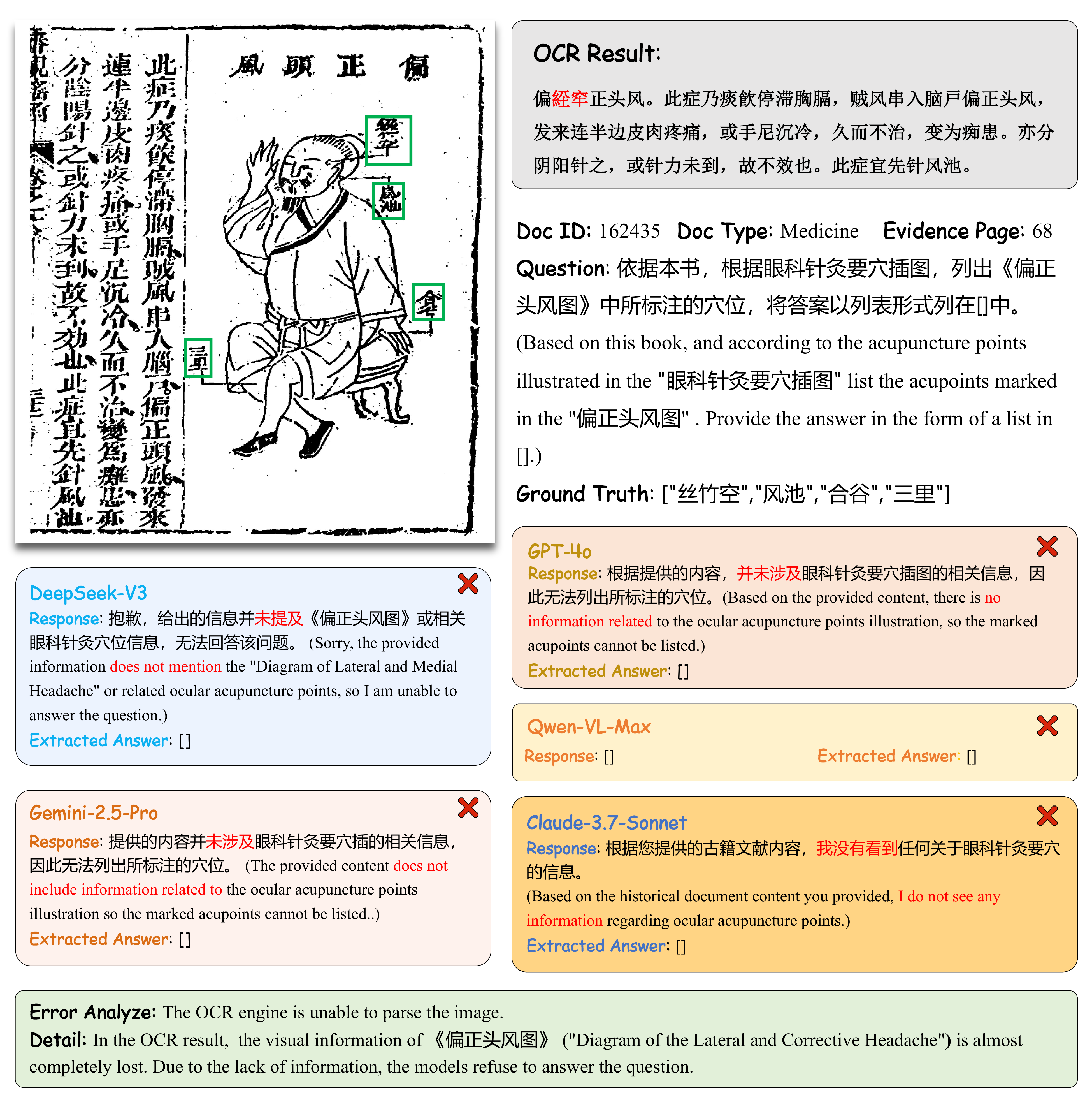}
  \caption{Text-input case 3. Evidence source: "Visual". Answer format: "List". Key information in the image is highlighted with green boxes, while incorrect information is marked in red. Zoom in for better view.}
  \label{fig:case_ocr_4}
\end{figure*}



\begin{figure*}[h]
  \centering
  \includegraphics[width=1.0\linewidth]{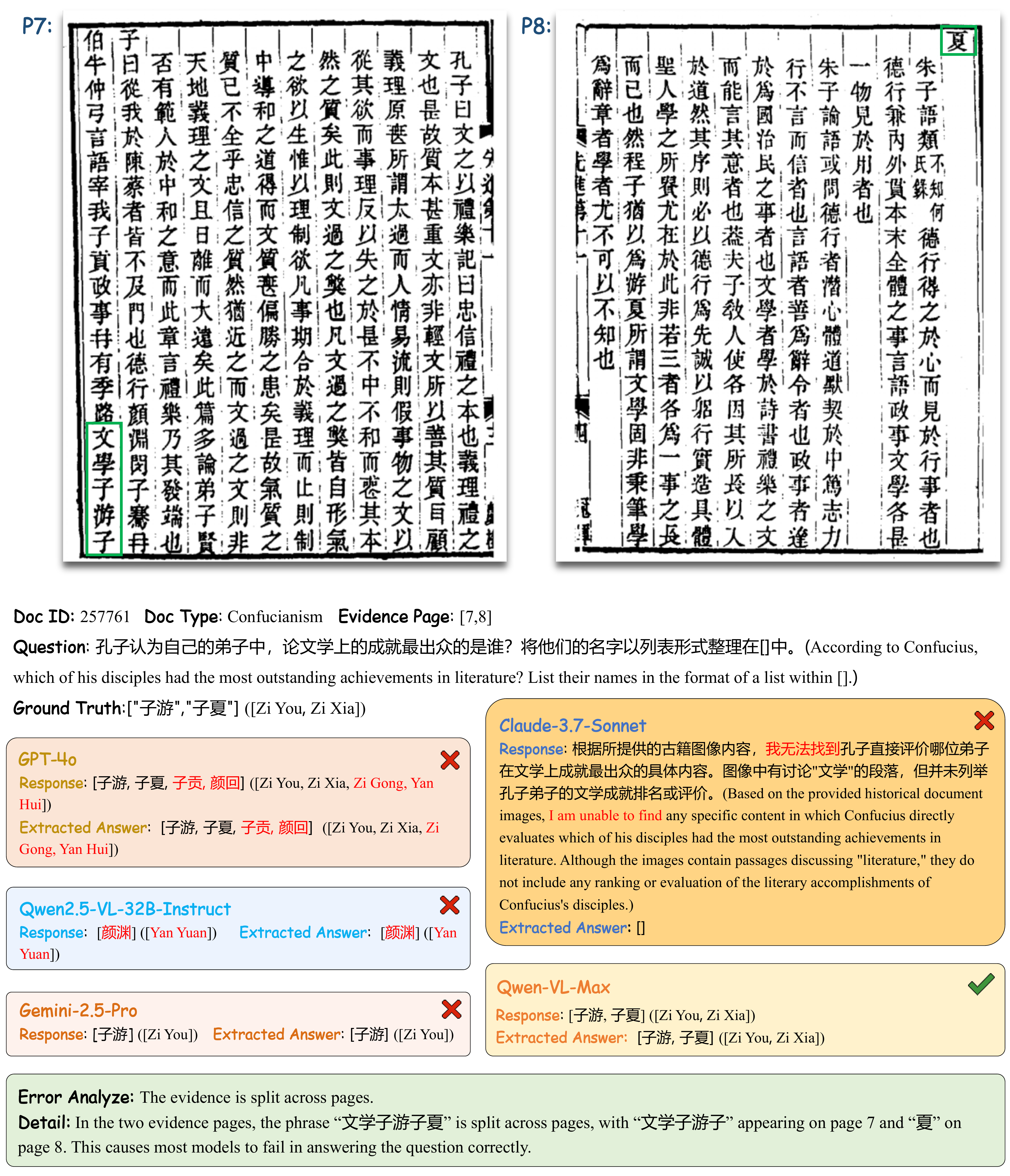}
  \caption{Image-input case 1. Evidence source: "Text". Answer format: "List". Key information in the images is highlighted with green boxes, while incorrect information is marked in red. Zoom in for better view.}
  \label{fig:case_img_2}
\end{figure*}

\begin{figure*}[h]
  \centering
  \includegraphics[width=1.0\linewidth]{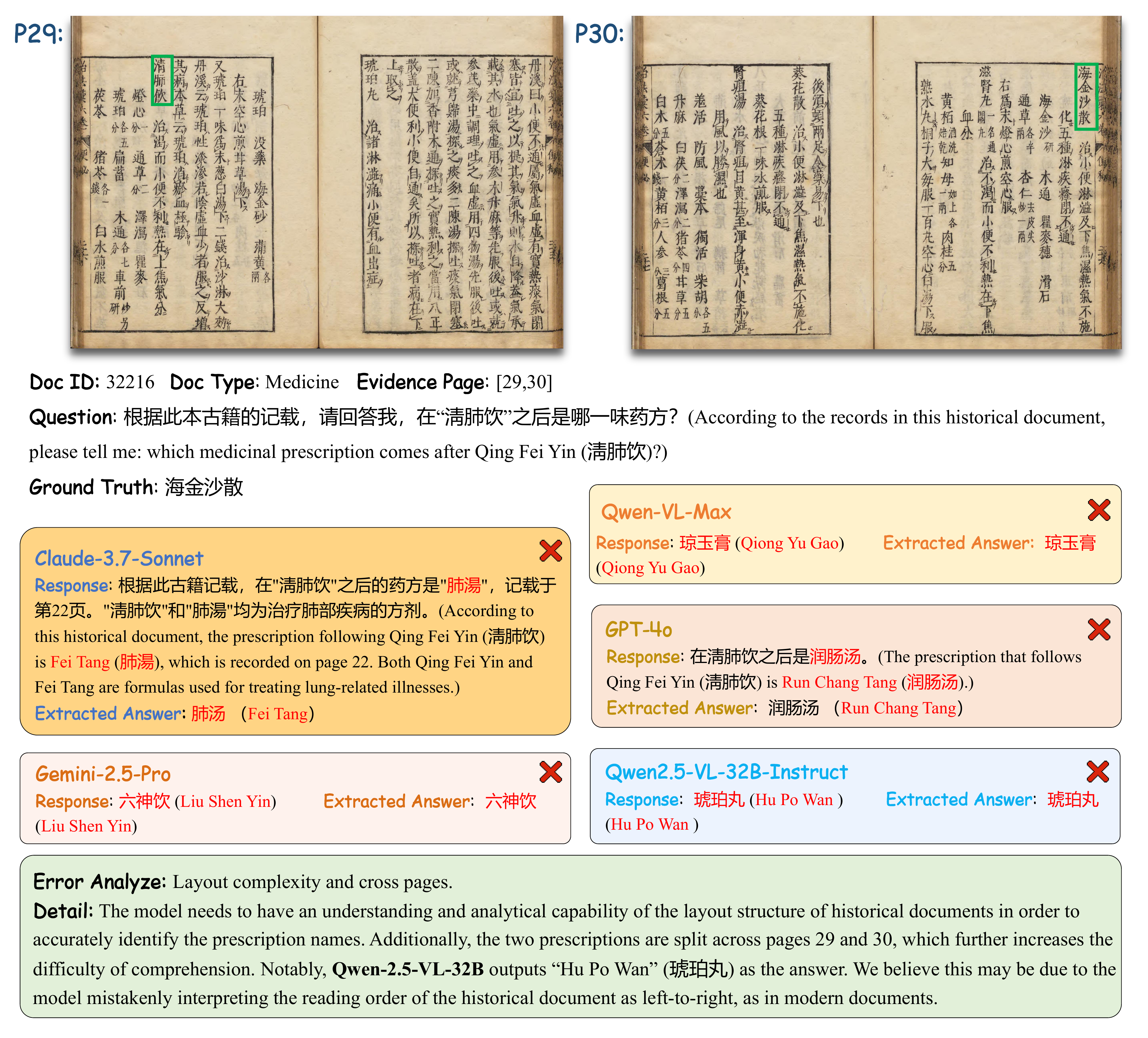}
  \caption{Image-input case 2. Evidence source: "Layout". Answer format: "String". Key information in the images is highlighted with green boxes, while incorrect information is marked in red. Zoom in for better view.}
  \label{fig:case_img_3}
\end{figure*}

\begin{figure*}[h]
  \centering
  \includegraphics[width=1.0\linewidth]{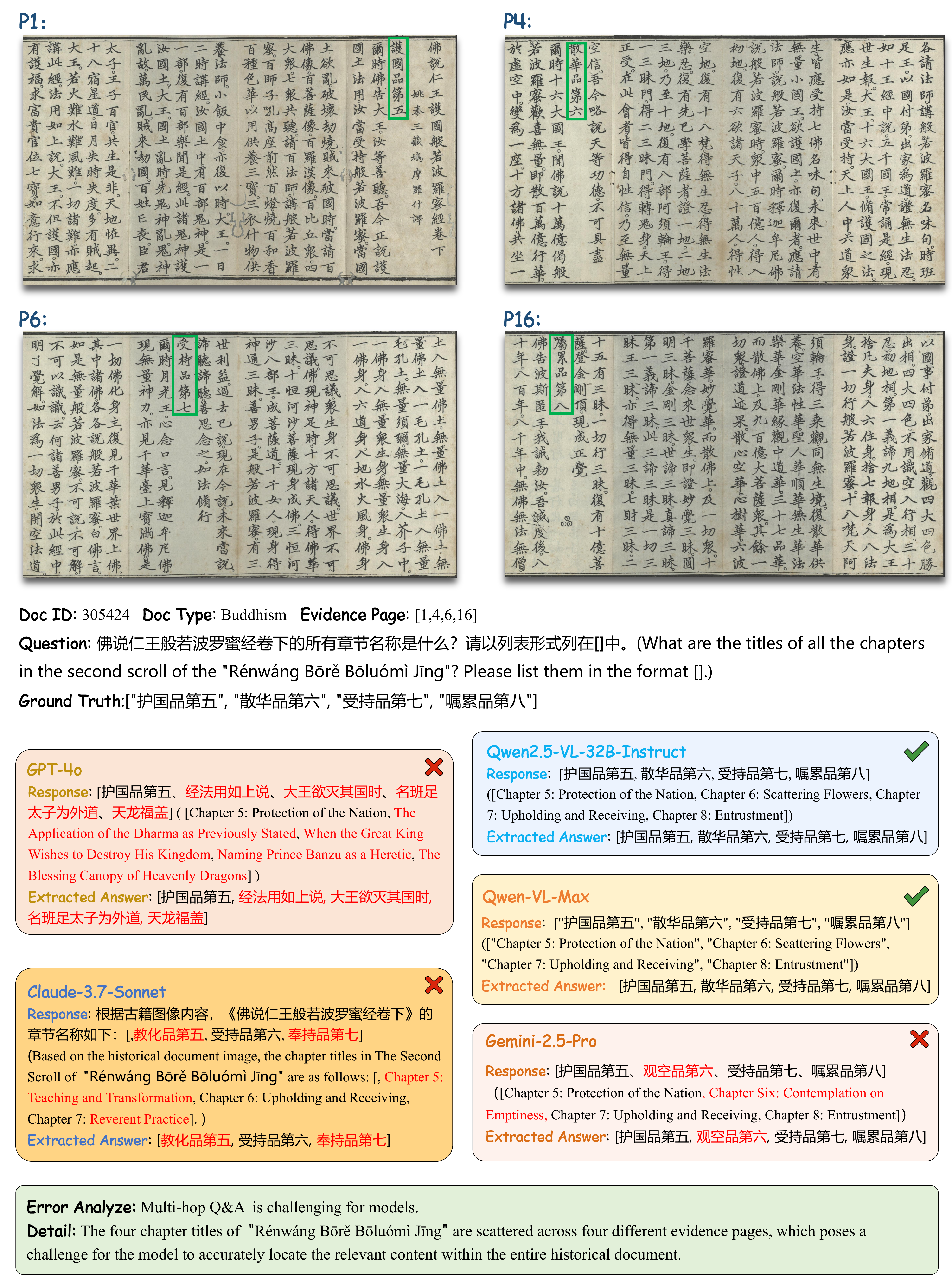}
  \caption{Image-input case 3. Evidence source: "Layout". Answer format: "List". Key information in the images is highlighted with green boxes, while incorrect information is marked in red. Zoom in for better view.}
  \label{fig:case_img_4}
\end{figure*}


\begin{figure*}[h]
  \centering
  \includegraphics[width=1.0\linewidth]{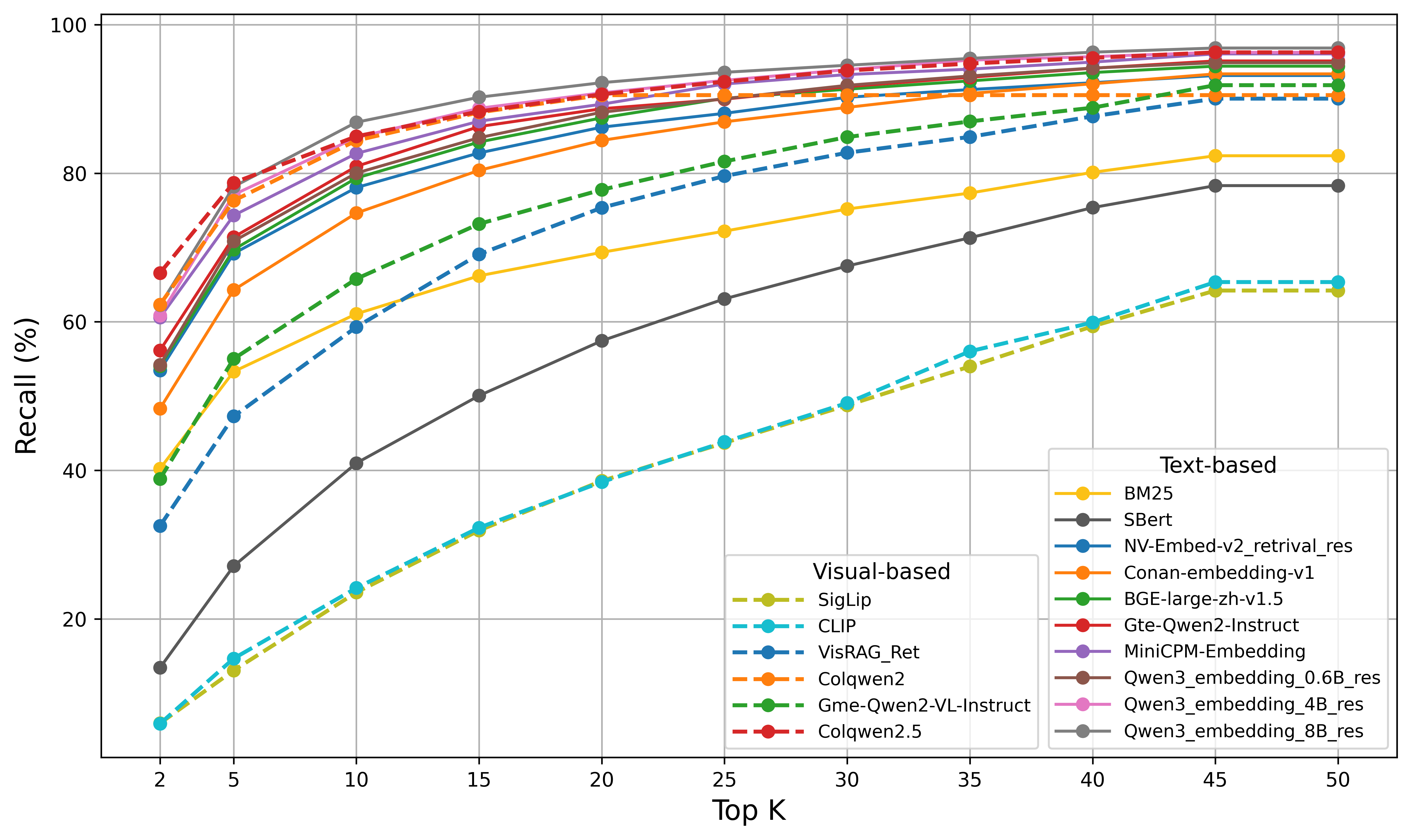}
  \caption{Top 2\textasciitilde50 recall rate of retrievers on LongHisDoc.}
  \label{fig:retriever}
\end{figure*}

\clearpage

\end{document}